\documentclass[journal,lettersize]{IEEEtran}
\ifCLASSOPTIONcompsoc
  \usepackage[nocompress]{cite}
\else
  \usepackage{cite}
\fi
\usepackage{amsmath,graphicx,epstopdf,amssymb,amsthm}
\usepackage{dsfont}
\usepackage{url}
\usepackage{color}
\usepackage{epstopdf}
\usepackage{verbatim}
\usepackage{cuted}
\usepackage{lipsum,graphicx,subcaption}
\usepackage[utf8]{inputenc}
\usepackage[english]{babel}
\usepackage{caption}
\usepackage{enumitem}
\usepackage[linesnumbered,ruled,vlined]{algorithm2e}
\DeclareCaptionLabelFormat{lc}{\MakeLowercase{#1}~#2}
\usepackage{balance}
\usepackage{float}
\usepackage{mathtools}

\DeclarePairedDelimiter\floor{\lfloor}{\rfloor}

\newtheorem{assumption}{Assumption}
\newtheorem{theorem}{Theorem}
\newtheorem{lemma}{Lemma}
\newtheorem{definition}{Definition}
\newtheorem{proposition}{Proposition}

\newtheorem{remark}{Remark}

\allowdisplaybreaks
\usepackage[protrusion=true,expansion=true]{microtype}

\usepackage[font=small]{caption}
\usepackage{enumitem}

\usepackage{tabularx,booktabs}
\newcolumntype{Y}{>{\centering\arraybackslash}X}
\usepackage{multirow}
\allowdisplaybreaks
 \usepackage{bbm}

\newlength{\Oldarrayrulewidth}

\def\BibTeX{{\rm B\kern-.05em{\sc i\kern-.025em b}\kern-.08em
    T\kern-.1667em\lower.7ex\hbox{E}\kern-.125emX}}

\makeatletter
\newcommand{\vast}{\bBigg@{3.5}}
\newcommand{\Vast}{\bBigg@{5}}
\makeatother
\usepackage{changepage}

\usepackage{amsmath,bm}
\newcommand\semiHuge{\fontsize{23.7}{31.38}\selectfont}
\newenvironment{skproof}{\noindent\textit{Sketch of  Proof:}}{\hfill$\blacksquare$}

\begin{document}
\title{\semiHuge UAV-assisted Online Machine Learning over Multi-Tiered Networks: A Hierarchical Nested Personalized Federated Learning Approach}
\author{Su Wang,~\IEEEmembership{Student~Member,~IEEE}, Seyyedali~Hosseinalipour,~\IEEEmembership{Member,~IEEE}, Maria~Gorlatova, \IEEEmembership{Member,~IEEE}, Christopher G. Brinton,~\IEEEmembership{Senior~Member,~IEEE}, and Mung Chiang,~\IEEEmembership{Fellow,~IEEE}
\thanks{S. Wang, C. Brinton, and M. Chiang are with Purdue University, IN, USA e-mail: \{wang2506,cgb,chiang\}@purdue.edu.}
\thanks{S. Hosseinalipour is with University at Buffalo (SUNY), NY, USA email: alipour@buffalo.edu}
\thanks{M. Gorlatova is with Duke University, Durham, NC, USA e-mail: maria.gorlatova@duke.edu.}
\thanks{This work was supported in part by the National Science Foundation (NSF) under grants CNS-2146171 and CNS-1908051, the Office of Naval Research (ONR) under grants N000142112472 and N000142212305, and by an IBM Faculty Award. }}
\maketitle

\begin{abstract}
We investigate training machine learning (ML) models across a set of geo-distributed, resource-constrained clusters of devices through unmanned aerial vehicles (UAV) swarms. The presence of time-varying data heterogeneity and computational resource inadequacy among device clusters motivate four key parts of our methodology: (i) \textit{stratified UAV swarms} of leader, worker, and coordinator UAVs, (ii) \textit{hierarchical nested personalized federated learning} ({\tt HN-PFL}), a distributed ML framework for personalized model training across the worker-leader-core network hierarchy, (iii) \textit{cooperative UAV resource pooling} to address computational inadequacy of devices by conducting model training among the UAV swarms, and (iv) \textit{model/concept drift} to model time-varying data distributions. 
In doing so, we consider both \textit{micro} (i.e., UAV-level) and \textit{macro} (i.e., swarm-level) system design.
At the micro-level, we propose network-aware {\tt HN-PFL}, where we distributively orchestrate UAVs inside swarms to optimize energy consumption and ML model performance with performance guarantees. 
At the macro-level, we focus on swarm trajectory and learning duration design, which we formulate as a sequential decision making problem tackled via deep reinforcement learning. 
Our simulations demonstrate the improvements achieved by our methodology in terms of ML performance, network resource savings, and swarm trajectory efficiency.
\end{abstract}

\vspace{-3mm}
\section{Introduction}
\label{sec:intro}
\noindent 
{\color{black} Traditionally, machine learning (ML) has been managed centrally~\cite{abdulrahman2020survey,wahab2021federated} with the training performed at one location using all of the data. However, Internet of Things (IoT) use cases, e.g., autonomous driving, smart manufacturing, and object tracking, have highly distributed datasets that are challenging or sometimes impossible to centralize~\cite{pham2022energy}. This has motivated the development of distributed ML, in particular federated learning (FL), techniques~\cite{hosseinalipour2020federated} to distribute model training across devices themselves. 
Though IoT devices are large in number, in many situations they are computationally limited (e.g., low cost wireless sensors)~\cite{jiang2022model} leading to situations where data are highly distributed and cannot be locally processed.}

\begin{figure}[t]
    \centering
    \includegraphics[width=0.43\textwidth]{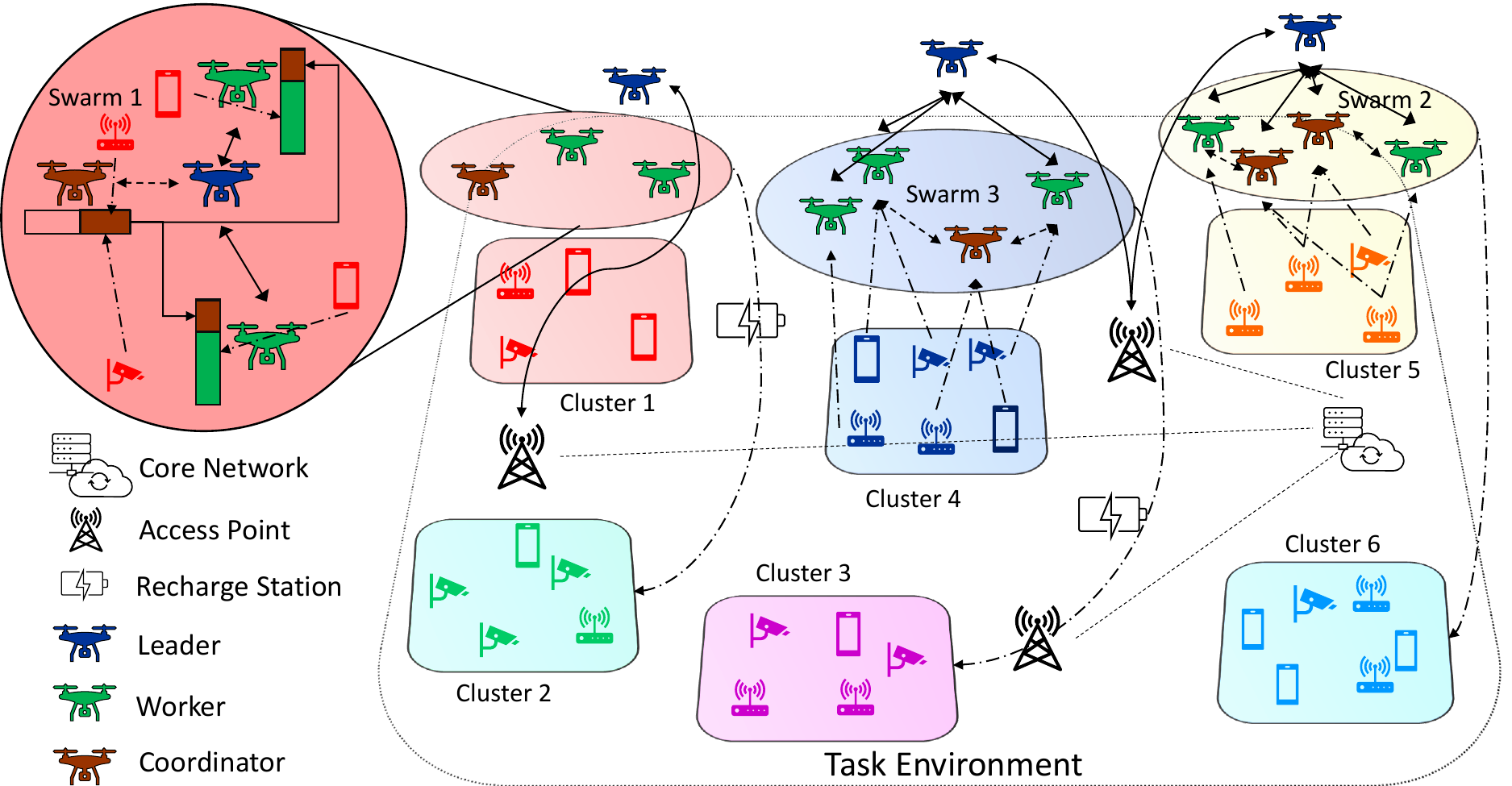}
    \vspace{-1.2mm}
    \caption{{\color{black}Schematic of UAV-enabled {\tt{HN-PFL}}. The network consists of multiple IoT device clusters, stratified UAV swarms, access points, and recharging stations. UAV swarm leaders orchestrate intra-swarm ML training by workers, and the data collection/transfer at workers and coordinator UAVs. Periodically, leaders travel to access points for the core network for global model aggregation.}
    {\color{black} As there are more clusters than swarms, the core network also determines the trajectories of the UAV swarms indicated by the dashed curved arrows.}
    }
    \label{fig:intro_model}
    \vspace{-6mm}
\end{figure}

{\color{black} Unmanned aerial vehicles (UAVs) have been recently incorporated into IoT networks to service applications such as surveillance, aerial base stations, smart agriculture, and search and rescue~\cite{rosabal2022minimization}. } 
{\color{black} Existing work~\cite{pham2022energy} have employed commercially-available UAVs with attached edge computing systems to act as a computational layer for receiving data from IoT devices and performing data processing tasks. 
Current literature is taking initial steps toward integrating such UAVs into FL~\cite{tu2022incentive,zeng2020federated,9210077,yang2020privacy}, which will be particularly useful for geo-distributed IoT settings with sparse cellular connectivity. 
However, these works have mostly focused on implementing the classic FL architecture using UAVs, where either (i) the data is assumed to already be stored on the UAVs~\cite{tu2022incentive,zeng2020federated,9210077} or (ii) the UAVs simply replace cellular base stations as the model aggregators~\cite{yang2020privacy}. 
These works further have not accounted for the time-evolving nature of data distributions in many IoT settings caused by environmental shifts. 
To address this, we propose a paradigm, depicted in Fig.~\ref{fig:intro_model}, that integrates UAVs and UAV swarms into the intelligent IoT ecosystem, where swarms must travel among the device clusters to efficiently collect measurements and conduct model training across time-varying data distributions.} 


Specifically, a UAV-assisted distributed ML paradigm must account for several factors. Online data variations across geo-distributed devices, as well as heterogeneity across the UAVs and devices in terms of data distributions and computation/communication resources, can each have a large impact on performance. On the infrastructure side, device-UAV-core network interfacing and the locations of recharging stations become important, especially across a large geographic region. To address these challenges, the methodology we propose consists of four interrelated parts: (i) a new model for UAV swarms, called \textit{stratified swarms}, suitable for distributed ML data collection and model training; (ii) \textit{hierarchical nested personalized federated learning} ({\tt{HN-PFL}}) to account for data heterogeneity in model updates; (iii) cooperative data processing across the UAVs via \textit{resource pooling}; and (iv) \textit{model/concept drift} tracking, which is tied to the model performance at devices and UAV movement patterns. 
We will develop and solve both UAV-level and swarm-level system optimizations: (a) {\tt{HN-PFL}} model performance is optimized via  efficient orchestration of the UAVs, taking into account the heterogeneity of the network elements, while (b) the swarm trajectories are optimized to account for anticipated model drifts across device clusters that will maximize training performance. 

\vspace{-3mm}
\subsection{Motivations and Applications} \label{ssec:ovr_motive}

The following scenarios will further motivate our system model for UAV-enabled online ML model training, where the UAVs perform both data collection and model training.

\textbf{Community Service Systems via Amazon Sidewalk:} 
Amazon proposes \textit{Sidewalk}~\cite{sidewalk_white_paper} to integrate community networks of IoT devices for household appliance diagnostics (e.g., garage system maintenance, pet finding, smart lighting) in the absence of consistent wi-fi connectivity by leveraging device-to-device (D2D) communications. 
However, efficient training of ML models on IoT devices and sharing them among distant neighborhoods faces four non-trivial challenges: (i) IoT devices may not be idle, plugged into power, or have direct access to a cellular base station (i.e., the core-network), preventing them from performing computationally intensive ML training, 
(ii) distant neighborhoods may not be reachable through D2D, (iii) neighborhoods' collected data may be time varying, and (iv) the neighborhoods may have extremely heterogeneous data distributions. 
We address the first two limitations by introducing UAV resource pooling and transferring model training vertically onto UAV swarms (e.g., the Amazon Prime Air delivery system), the third via model/concept drift at different neighborhoods, and the fourth through model personalization.

\textbf{Distributed Surveillance in Smart Cities:}
Our proposed method has natural applications to smart city surveillance systems. For instance, multiple UAV swarms can be spread throughout a city, extracting data from sensors and cameras as well as those devices in rural areas without direct access to the closed-circuit cellular network to train an ML model~\cite{piza2019cctv}. 
Additionally, these swarms can travel to diverse neighborhoods (e.g., industrial parks vs. academic campuses) within a city,
allowing for model recalibration in the presence of dynamic environments and improving the quality of personalized models
{\color{black} through a global meta-learning approach that captures data commonalities across the city (see Sec.~\ref{sec:HN-PFL}).} 

\textbf{Machine Learning on Wireless Sensor Networks:}
ML techniques have been adapted for wireless sensor networks (WSN) with respect to a wide variety of tasks, e.g., object targeting and event detection. 
Our model can be contextualized, as an example, for integrated-WSN-UAV response systems, wherein wireless sensors collect data, e.g., water level or seismic energy, and offload it to UAVs, which then train ML models~\cite{erdelj2017help}.
The UAV swarms can travel to distant and disconnected WSNs, e.g., across a beach/coast, to gather heterogeneous and time-varying data, e.g., day vs. night tidal measurements, and train ML models for each WSN cluster, thus integrating the system together. 

\vspace{-3mm}
\subsection{Related Work}
This paper contributes to both literature in distributed ML over wireless networks and ML for UAV networks. Below, we provide a summary of related works and highlight the main contributions of our methodology. 

\textbf{Distributed ML over wireless networks:}
{\color{black}Recent literature concerning ML by wireless networks has shifted towards federated learning~\cite{8970161}, and is mostly focused on studying the convergence and behavior of federated learning over wireless networks~\cite{amiri2020federated,wang2019adaptive,9437529,hosseinalipour2022parallel,hosseinalipour2020multi,yang2020federated}. }
Conventional federated learning assumes training a single ML model for all the engaged devices. However, upon having extreme data heterogeneity across the devices, training a single model may result in poor model performance for a portion of devices. This has motivated a new trend of research in ML that aims to train user-specific ML models, 
called personalized federated learning~\cite{fallah2020personalized}, in which meta-gradient updates are introduced to enhance the training efficiency. 
As compared to this literature, we develop and investigate a new distributed ML paradigm over wireless networks called \textit{hierarchical nested personalized federated learning} ({\tt HN-PFL}) inspired by meta-gradient updates, which is different than hierarchical FL architectures, e.g.,~\cite{liu2020client}, which rely on conventional gradient descent updates for local model training. 
Furthermore, we develop a new framework for network-aware {\tt HN-PFL} over UAV-assisted wireless networks that considers model training under heterogeneity of resources in wireless networks. Our resulting optimization formulation balances the tradeoffs between ML model performance and network parameters such as data offloading, training batch sizes, and CPU cycles, and is also part of our contributions.

\textbf{ML for UAV-assisted networks:} Deep learning techniques have been utilized to enhance the efficiency of wireless networks~\cite{8743390}. 
In UAV-assisted  networks, especially when the UAVs are deployed as aerial base stations, reinforcement learning (RL) has been utilized to carry out a variety of tasks, such as trajectory and power control for UAVs~\cite{zhao2020multi}, and UAV server quality~\cite{8807386}. 
A key observation from the aforementioned existing literature is that UAV-assisted networks are difficult to model using traditional closed form methodologies and, as such, benefit from intelligent and autonomous management via RL.  
As compared to current literature, we introduce a new system model for UAV swarms and use a RL method to manage only the macro-level (i.e., swarm trajectories and temporal parameters of ML model training) of our methodology. 
\begin{table*}[t]
{\normalsize
\caption{Summary of Key Notations for Devices, UAVs, Network Optimization, Machine Learning, and Swarm Trajectory Design}
\vspace{-1.5mm}
\label{tab:notation}
\resizebox{.9999\textwidth}{!}{
\setlength\tabcolsep{13pt}
\begin{tabularx}{1.56\textwidth}{m{3em} l | m{3em} l | m{3em} l}
\cmidrule[3pt]{1-6}
\multicolumn{2}{c}{\bf{Device Cluster, UAV Swarm, and Network}} & \multicolumn{4}{c}{\textbf{Data Processing/Offloading Optimization and Network Energy Consumption}}\\ 
$\mathcal{C}$ & Set of all device clusters 
& $\rho_{i,j}(t)$ & Data transfer ratio from device $i$ to UAV $j$
& $\varrho_{i,j}(t)$ & Data transfer ratio from UAV $i$ to $j$\\
$\mathcal{R}$ & Set of recharging stations
& $g_j(t)$ & Adjustable CPU cycle frequency at UAV $j$ 
& $\zeta^{\mathsf{G}}_j(k)$ & Time used by UAV $j$ to gather data \\
$\mathcal{A}$ & Set of access points 
& $\zeta^{\mathsf{P}}_j(k)$ & Time used by UAV $j$ to process data
& $E^{\mathsf{Ba}}_j(s)$ & Starting battery of UAV $j$ at the $s$-th sequence \\ 
$U(s)$ & Number of active swarms at $s$-th sequence
& $M$ & Number of bits per datapoint
& $\widetilde{M}$ & Number of bits used for model parameters\\ \cline{3-6}
$\mathcal{U}(s)$ & Set of active swarms at $s$-th sequence
& \multicolumn{4}{c}{\textbf{Machine Learning Notation}} \\
$\ell_u$ & Leader UAV of $u$-th swarm 
& $F_j$ & Meta loss function at UAV $j$
& $f_j$ & ML loss over all data in $\mathcal{D}_j$ \\
$\mathcal{W}_u$ & Worker UAVs in the $u$-th swarm 
& $\widehat{f}_j$ & ML loss over a datapoint in $\mathcal{D}_j$
& $\mathbf{w}_j(t)$ & Model parameters at UAV $j$ \\
$\widehat{u}$ & Non-leader UAVs in $u$-th swarm 
& $\alpha_{j,1}(t)$ & Inner mini-batch ratio for UAV $j$
& $\alpha_{j,2}(t)$ & Outer mini-batch ratio for UAV $j$ \\
$\widehat{\mathcal{W}}_u$ & Coordinator UAVs in the $u$-th swarm 
& $\alpha_{j,3}(t)$ & Hessian mini-batch ratio for UAV $j$
& $t_s$ & Starting time for the $s$-th sequence \\
$\widetilde{\mathcal{D}}_i(t)$ & Dataset of $i$-th device
& $\tau_s^{\mathsf{L}}$ & Local aggregation period for $s$-th sequence
& $\tau_s^{\mathsf{G}}$ & Global aggregation period for $s$-th sequence \\ 
$\mathcal{D}_j(t)$ & Dataset of $j$-th UAV
& $k$ & Local aggregation index
& $k^{'}$ & Global aggregation index \\ \cline{3-6}
$\widetilde{\mathcal{D}}_c(t)$ & Union of datasets at cluster $c$
& \multicolumn{4}{c}{\bf{Swarm Trajectory Optimization}} \\ 
$B_j^{\mathsf{D}}(t)$ & Max dataset size for UAV $j$
& $\mathcal{X}(s)$ & Swarm positions at $s$-th sequence
& $G_c(s)$ & Online gradient as a result of model drift \\
$T_s$ & Duration of the $s$-th sequence 
& $\mathcal{Z}(s)$ & Network state encoding at $s$-th sequence  
& $\mathcal{H}(s)$ & DRL agent action at the end of the $s$-th sequence \\
$\Lambda_c $ & Model drift of cluster $c$
& $V(s)$ & Reward of the $s$-th sequence 
& $Q_{\theta}$ & Q-network for trajectory design \\
\cmidrule[1pt]{1-6}
\end{tabularx}
 }
}
\vspace{-6mm}
\end{table*}

\vspace{-5mm}
\subsection{Outline and Summary of Contributions}
Our contributions in this work can be summarized as follows:
\begin{itemize}[leftmargin=4mm]
    \item We introduce the framework of UAV-enabled online model training for a set of geo-distributed ground device clusters. We propose \textit{stratified UAV swarms}, which presume different roles for the UAVs: (i) a \textit{leader} that manages UAVs within the swarm (e.g., adjusting the CPU cycles and mini-batch sizes), and determines data transfer configurations, (ii) \textit{workers} that conduct ML model training through \textit{resource pooling}, and (iii) \textit{coordinators} that enable data relaying between the devices and the worker UAVs.
    \item We develop hierarchical nested personalized federated learning ({\tt HN-PFL}), which exploits {\color{black} meta-gradient based learning across disconnected device clusters and yield personalized local models.}
    Through the nesting of intra-swarm updates within inter-swarm aggregations, {\tt HN-PFL} conducts ML model training across the \textit{worker-leader-core} network hierarchy. 
    We analytically characterize the convergence behavior of {\tt HN-PFL}, which leads us to new convergence bounds for distributed ML. 
    \item We integrate network characteristics into ML training/performance by formulating a joint energy and ML performance optimization problem, which aims to configure the data offloading among devices-UAVs and UAVs-UAVs, adjust the CPU cycles of the UAVs, and obtain the mini-batch sizes used at the worker UAVs. This formulation is among the first in literature to consider all these design variables together. 
    We demonstrate that the problem belongs to the category of \textit{complementary geometric programming} problems which are highly non-convex and NP-Hard. We then develop a distributed method, with performance guarantee, based on posynomial condensation to solve the problem for all UAV swarms in parallel. 
    \item  We formulate the problem of UAV swarm trajectory design, alongside of which we also optimize the learning duration of {\tt HN-PFL}. In the formulation, we consider \textit{online model training} under temporal data distribution variations, which is quantified via \textit{model/concept drift}. We demonstrate that the problem solution is intractable and then cast the problem as a sequential decision making problem tackled via a deep reinforcement learning-based method.
\end{itemize}


\vspace{-3mm}
\section{System Model}
\label{sec:sys_model}
\noindent In this section, we introduce the system components, which include IoT device clusters (Sec.~\ref{ssec:clusters}),  UAV swarms, recharging stations, and access points (Sec.~\ref{ss:UAV Swarms}). {\color{black} We also provide an overview of our entire methodology, including the macro- and micro-level design in Sec.~\ref{ssec:full_architecture}.}

\begin{figure*}[t]
    \centering
    \includegraphics[width=0.93 \textwidth]{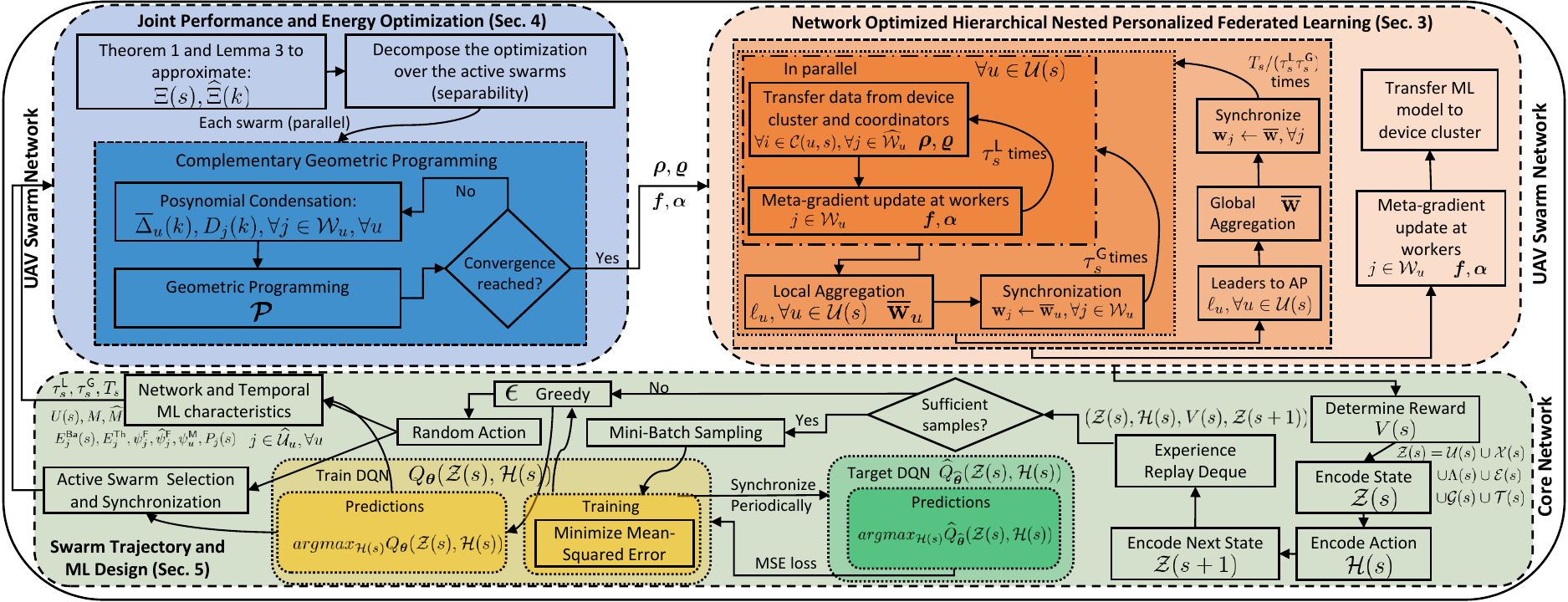}
    \vspace{-1.5mm}
    \caption{Our methodology consists of three components - the HN-PFL architecture (Sec.~\ref{sec:HN-PFL}), the data transfer and processing optimization (Sec.~\ref{sec:netwrok_pfl}), and the swarm trajectory design (Sec.~\ref{sec:swarmTraj}). Our three components effectively reduce the problem into two layers: a \textit{\textbf{micro}}-level wherein ML model training is optimized and subsequently performed via HN-PFL and a \textit{\textbf{macro}}-level wherein the optimality of the macro-system (i.e. the network-wide consideration of all separate device clusters) is achieved. We additionally emphasize that a sole reinforcement learning based architecture would struggle greatly from the large and dynamic state-space of energy-efficient swarm trajectory design, HN-PFL performance and energy maximization, coordinator UAV transmissions, online model drifts, and meta-training convergence determination. 
    }
    \label{fig:overall_flow}
    \vspace{-7mm}
\end{figure*}

\vspace{-4mm}
\subsection{Device Clusters and Data Distributions}
\vspace{-1mm}
\label{ssec:clusters}
We consider a set of geo-distributed devices collected into into $C$ clusters, denoted by $\mathcal{C} = \{\mathcal{C}_1, \cdots, \mathcal{C}_{C}\}$, based on geographic proximity. {\color{black} Hereafter, we will refer to an arbitrary cluster via $c$ for brevity.}
At time instance $t \in \mathcal{T}\triangleq \{1,2,...\}$, we denote $\widetilde{\mathcal{D}}_i(t)$ as the set of datapoints device $i \in c$ has collected at that time. Each $x \in \widetilde{\mathcal{D}}_i(t)$ is a data sample containing model features and (possibly) a target variable. 

{\color{black} Motivated by the real-world applications in Sec.~\ref{ssec:ovr_motive}, wherein data distributions  are expected to change temporally (e.g., at environmental sensors), we focus on \textit{online/dynamic} model training and deployment, which is different from current literature 
that mainly consider static data at the devices
(e.g.,~\cite{amiri2020federated,wang2019adaptive,9437529,hosseinalipour2022parallel}). Our goal is to obtain and periodically update a personalized ML model for each cluster $c$ to use for real-time inference. However, due to significant computation/communication constraints the devices may be unable to train high-dimensional ML models (e.g., surveillance cameras as described in Sec.~\ref{ssec:ovr_motive}).}
{\color{black}{
We thus transpose the ML model training to UAV swarms via data offloading from the devices. We assume that each device $i \in c$ collects data in its buffer $\widetilde{\mathcal{B}}_i$ of finite size, and that each device is capable of transmitting data to nearby UAVs to perform model training. We consider $\widetilde{\mathcal{B}}_i$ as a double-ended queue (i.e., deque) so that, when filled, new data will displace the oldest.}}\footnote{\textbf{Henceforth, we use calligraphic (e.g., $\widetilde{\mathcal{D}}_{i}(t)$) to denote a set, and non-calligraphic (e.g., $\widetilde{D}_{i}(t)$) to denote its cardinality.}}

\vspace{-3.5mm}
\subsection{Swarms, Recharging Stations, and Access Points}
\label{ss:UAV Swarms}
\vspace{-.6mm}
To conduct efficient ML model training at the UAVs, we introduce stratified UAV swarms, where UAVs have different roles for data collection and data processing (Sec.~\ref{sssec:uav_swarms}). 
Since the battery-limited UAVs perform energy-intensive ML model training, we also integrate a recharging methodology via recharging stations in our model (Sec.~\ref{sssec:recharge_station}). Also, to synchronize the ML model training and UAV orchestration, we consider a set of access points in the network (Sec.~\ref{sssec:AP}). 
In the following, we explain each of these components. 
A schematic of our model is depicted in Fig.~\ref{fig:intro_model}. 

\subsubsection{UAV Swarms and Stratification}
\label{sssec:uav_swarms}
We consider a set of $U$ UAV swarms $\mathcal{U} = \{\mathcal{U}_1,\cdots,\mathcal{U}_U\}$, and assume that $U<C$. 
We denote an arbitrary UAV swarm  as $u$ for brevity.
We assume that each UAV swarm $u$ is composed of UAVs with heterogeneous capabilities, e.g., from micro-drones weighing under 200grams that have data storage capabilities to medium-sized fixed/rotatory wing UAVs that have more advanced computational capabilities~\cite{lu2019toward}. Subsequently, to maximize the performance of each swarm, we propose a new \textit{swarm stratification} model, which compared to current literature on UAV trajectory design~\cite{liu2019trajectory} is tailored specifically for ML training tasks. In our swarm stratification, there are three types of UAVs: (i) a leader $\ell_u$ of each swarm $u$, with the set of leaders across swarms denoted by $\mathcal{L}=\cup_{u\in \mathcal{U}} \ell_u$, (ii) a set of workers $\mathcal{W}_u$ in swarm $u$, with the set across swarms denoted by $\mathcal{W}=\cup_{u\in \mathcal{U}} \mathcal{W}_u$, and (iii) a set of coordinators $\widehat{\mathcal{W}}_u$ in $u$, gathered via the set $\widehat{\mathcal{W}}=\cup_{u\in \mathcal{U}}\widehat{\mathcal{W}}_u$. 
For convenience, we refer to the workers and coordinators together as $\widehat{u} = \mathcal{W}_u \cup \widehat{\mathcal{W}}_u= u \backslash \{\ell_u\}$. UAVs in $\widehat{u}$ collect data from nearby IoT devices, with the workers $\mathcal{W}_u$ conducting model training based on their gathered data, and the coordinators $\widehat{\mathcal{W}}_u$ relaying data to other UAVs in $\widehat{u}$ and building a data profile of the device cluster under visit. 

{\color{black} UAV swarms transfer updated models to device clusters at the end of training sequences, and the devices employ these ML models to carry out inference tasks. At the beginning of each training sequence, i.e., when swarms arrive at clusters, each UAV swarm will have been synchronized through the core network with the same global ML model.} When swarm $u$ arrives at some device cluster $c$, the leader $\ell_u$ scatters the UAVs $\widehat{u}$ throughout predetermined locations in $c$. 
The exact positioning of the UAVs is not the focus of this work and can be computed \textit{a priori} at the core network~\cite{zeng2019energyRotary}.
{\color{black} Since there are fewer UAV swarms $\mathcal{U}$ than device clusters $\mathcal{C}$, we will present a methodology to determine UAV swarm trajectories so that all device clusters can be supported in Sec~\ref{sec:swarmTraj}. Here, we focus on defining the interactions between a single UAV swarm and device cluster. } 
All UAVs in $\widehat{u}$ gather data from devices $i \in c$ through a ``wake up and sleep'' paradigm, where the UAV notifies nearby IoT devices and prompts them to upload data. 
We denote $\mathcal{D}_j(t)$ as the dataset UAV $j \in \widehat{u}$, with buffer size $B_j^{\mathsf{D}}(t)$, obtains at time $t$. Workers $\mathcal{W}_u$ form a pool of computational resources above the device cluster and use their gathered data for cooperative ML training, in which they engage in periodic communication with the leader $\ell_u$ regarding their training results.
On the other hand, coordinators $\widehat{\mathcal{W}}_u$ act as \textit{aerial data caches} to relay data to other UAVs in $\widehat u$. 
This arrangement allows more energy-efficient data transfer from IoT devices to worker UAVs via coordinator relaying. We formalize $\mathcal{D}_j(t)$, $\forall j$, through our optimization in Sec.~\ref{ssec:netaware}. 

\vspace{-0.5mm}
\subsubsection{Recharging Stations}
\label{sssec:recharge_station}
We consider a set of geo-distributed recharging stations $\mathcal{R}$ in the network. 
We assume that when any UAV $j \in u$ reaches a minimum battery threshold, the entire swarm $u$ must travel to a recharging station $r \in \mathcal{R}$.

\vspace{-0.6mm}
\subsubsection{Access Points (AP)}
\label{sssec:AP}
We consider a set of gateway APs $\mathcal{A}$, which can be interpreted for example as cellular base stations.
All APs $a\in \mathcal{A}$ are connected through the core network. 
The APs are used by the leader UAVs $\mathcal{L}$ to communicate with the core network, which determines swarm trajectories and synchronizes ML training among the swarms. 

{\color{black} Deployments of this architecture in practice will require three types of signaling: (i) UAV-UAV communication, i.e., among leader, coordinator, and worker UAVs; (ii) device-UAV communication, i.e., data transfers from devices and requests from coordinator/worker UAVs; and (iii) UAV-AP interactions, i.e., for global model aggregations among swarm leaders. We detail potential approaches to these components in Appendix~\ref{sec:signaling}.}

\subsection{Components of Our Methodology and Roadmap} \label{ssec:full_architecture}
{\color{black} We break down our methodology for conducting geo-distributed ML using UAVs into three parts: (i) ML model training through the worker-leader-core network hierarchy, (ii) efficient orchestration of the UAVs inside each swarm, and (iii) energy and performance-driven design of the UAV swarm trajectories across device clusters. For (i), we develop hierarchical nested personalized federated learning ({\tt HN-PFL}) and subsequently derive its performance through theoretical bounds in Sec.~\ref{sec:HN-PFL}. 
For (ii), we formulate a network-aware ML performance and swarm-wide energy consumption optimization problem, and then develop a distributed solver in Sec.~\ref{sec:netwrok_pfl}. 
For (iii), we capture the temporal variation of data across the clusters and obtain the sequence of cluster visits for the UAV swarms (i.e., swarm trajectories) via integrating a deep reinforcement learning architecture in Sec.~\ref{sec:swarmTraj}. 
The interactions between these three components and an overview of their functionality is depicted in Fig.~\ref{fig:overall_flow}. 
From a high-level, the swarm-wide optimization (blue block) determines {\tt HN-PFL}'s meta settings, and the performance of {\tt HN-PFL} (orange block) is used to determine subsequent cluster visits (green block). This process then repeats, cyclically. 
We explain the details of each block of Fig.~\ref{fig:overall_flow} in Sec.~\ref{sec:HN-PFL},~\ref{sec:netwrok_pfl}, and~\ref{sec:swarmTraj}, respectively.}

\vspace{-3mm}
\subsection{Metrics of Interest}
The evaluation of our methodology requires several primary metrics: classification loss at the UAVs (Thm.~\ref{theory:theorem}), the mismatch between the model performance at the UAVs vs. at the device clusters (Prop.~\ref{theory:prop_l2}), and energy consumption (Sec.~\ref{sec:netwrok_pfl}). We will further demonstrate that our learning method provides substantial improvements compared to baselines in terms of convergence rate (Sec.~\ref{ss:proof_of_concept}), network resource savings (Sec.~\ref{ss:optim_results}), and trajectory design efficiency (Sec.~\ref{ss:swarm_trajectories}).

\vspace{-3mm}
\section{Hierarchical Nested Personalized \\ Federated Learning ({\tt HN-PFL})} \label{sec:HN-PFL}

\noindent In this section, we develop our UAV-enabled methodology for personalized federated learning (PFL). We begin with the rationale (Sec.~\ref{ssec:overview}), and then present the {\tt HN-PFL} algorithm (Sec.~\ref{ss:HN-PFL_only}). Finally, we theoretically analyze the convergence of our distributed ML method (Sec.~\ref{ssec:convergence}).

\vspace{-3mm}
\subsection{Overview and Rationale}
\label{ssec:overview}
Conventional federated learning (FL) trains a single ML model suitable for all devices~\cite{mcmahan2017communication}. As devices may exhibit significant heterogeneity in their data distributions, training a global model used for all the devices may lead to poor overall performance. 
This has motivated personalized federated learning (PFL)~\cite{fallah2020personalized}, which trains device-specific ML models by leveraging the \textit{commonality} across the devices' data. 
Conventional FL and PFL both assume a ``star'' learning topology, where {\color{black} workers/devices are connected to and able to communicate directly with a main server~\cite{hosseinalipour2020federated}.} 



\begin{figure}[t]
    \centering
    \includegraphics[width=0.47\textwidth]{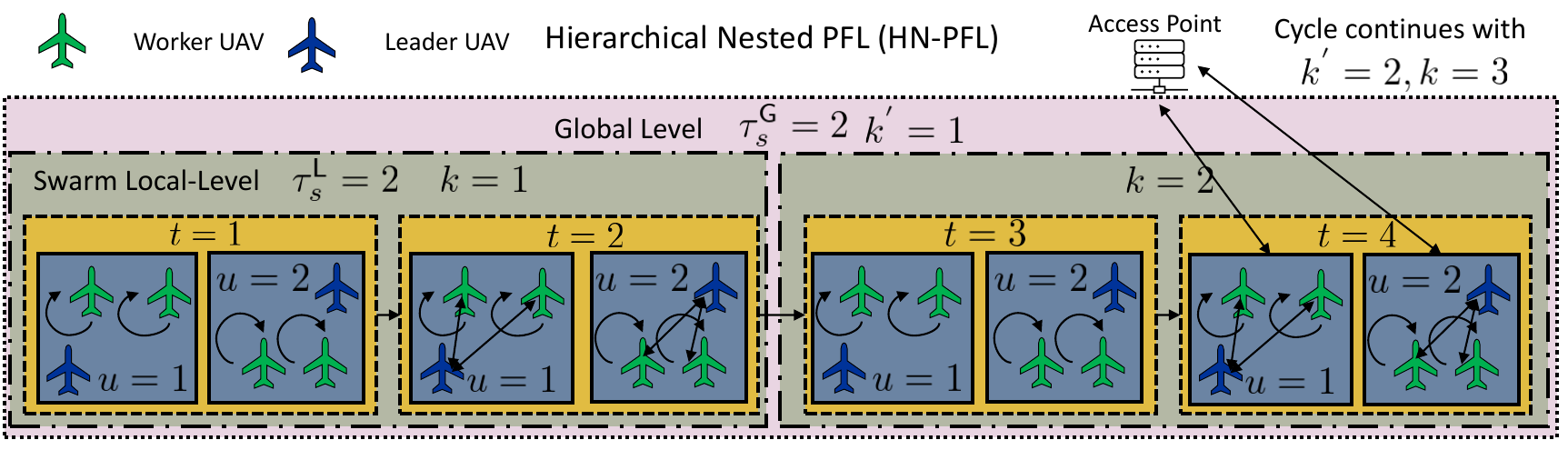}
    \caption{{\color{black}Behavior of workers, leaders, and APs in {\tt HN-PFL}.
    Workers perform local meta-updates, and leaders aggregate their workers' model parameters with period $\tau_s^{\mathsf{L}}$. The core network synchronizes leader parameters every $\tau_s^{\mathsf{G}}$ local aggregations.}} 
    \label{fig:PFL_algo}
    \vspace{-5mm}
\end{figure}

In our setting, the star topology assumed in FL/PFL applies poorly as the IoT device clusters are geo-distributed, and UAVs visiting the clusters may not have direct access to an AP.
On the other hand, UAV-to-UAV communications within a swarm is comparatively low in resource consumption, which motivates \textit{local model aggregations} inside the swarms enabled by the leader UAV. The leaders can then occasionally visit their nearest AP for global aggregation of their associated swarm ML model parameters.
By \textit{nesting} intra-swarm (local) aggregations within inter-swarm (global) aggregations, we develop a new methodology to generalize the star topology in conventional FL/PFL to that of a \textit{hierarchical} tree, called {\tt HN-PFL} where the ML model training is segmented into two layers: (i) swarm-level between leaders and their constituents, and (ii) global-level between access points/core-network and swarm leaders. To the best of our knowledge, {\tt HN-PFL} is the first hierarchical personalized federated learning architecture in literature. 
\vspace{-4mm}
\subsection{{\tt HN-PFL} Algorithm}
\label{ss:HN-PFL_only}
\vspace{-.3mm}

{\tt HN-PFL} breaks down the model training problem into two layers: (i) \textit{workers-leaders}, in which the worker UAVs carry out the ML model training and the leader UAVs perform swarm-wide (local) aggregations; and (ii) \textit{leaders-APs}, in which the leader UAVs engage in global aggregations.


Since our problem requires the swarms $\mathcal{U}$ to travel between IoT device clusters, {\tt HN-PFL} carries out the model learning through consecutive \textit{training sequences}. Each training sequence starts when all active UAV swarms (i.e., non-recharging swarms) arrive at their designated device clusters, and concludes when the swarms finish the model training and begin travelling to their next destination. 
We denote the start of the $s$-th training sequence, $s = 1,2,\cdots$, by $t_s \in \mathcal{T}$, its scheduled interval as $\mathcal{T}_s = \{t_s,\cdots, t_s + T_s - 1 \}$, and the active swarms for the $s$-th sequence as $\mathcal{U}(s) \subseteq \mathcal{U}$. Defining the active swarms $\mathcal{U}(s)$ with respect to the training sequence encompasses the cases where a portion of UAV swarms recharge their batteries and thus are not engaged in ML model training during $\mathcal{T}_s$.




At each time $t \in \mathcal{T}_s$, each worker UAV $j \in \mathcal{W}_u$, $u \in \mathcal{U}(s)$ conducts a \textit{local model update}. This consists of computing its next ML model parameter vector $\mathbf{w}_j(t+1)$ using a meta-gradient update~\cite{finn2017model} defined as:
\vspace{-2mm}
\begin{equation} \label{eq:pfl_meta_gradient}
\begin{aligned}
\mathbf{w}_{j}(t+1) = \mathbf{w}_{j}(t) - \eta_{_{2}} \nabla \widetilde{F}_{j}(\mathbf{w}_{j}(t)), ~t+1\in \mathcal{T}_s,
\end{aligned}
\vspace{-2mm}
\end{equation}
where $\eta_{_{2}}>0$ is the meta-update step-size, and $\nabla \widetilde{ F}_j(\mathbf{w}_j(t))$ is the mini-batch approximation of the meta-gradient $\nabla F_j(\mathbf{w}_j(t))$. 
$\nabla F_j(\mathbf{w}_j(t))$ is the gradient of the meta-function $F_j$, defined as the loss of the gradient descent procedure:
\vspace{-2mm}
\begin{equation}\label{eq:newMet}
    F_j(\mathbf{w}_j(t)) = f_j\big(\underbrace{\mathbf{w}_j(t) - \eta_{_{1}} \nabla f_j(\mathbf{w}_j(t))}_{(a)}\big),
\end{equation}
where $\eta_{_{1}} >0$ is the step size for gradient descent and $f_j$ is the local loss function over the local dataset $\mathcal{D}_j(t)$ at UAV $j$:
\vspace{-1mm}
\begin{equation} \label{eq:worker_loss} 
    f_j(\mathbf{w}_j(t)) = \frac{\sum_{x \in \mathcal{D}_j(t)} \widehat{f}(\mathbf{w}_j(t);x)}{D_j(t)},
    \vspace{-1mm}
\end{equation}
and $\widehat{f}(\mathbf{w}_j(t);x)$ is the loss per datum $x$. This meta-gradient procedure effectively updates the parameters on the loss of the update rule,
{\color{black} which, when connected to other meta-gradient results, yields an ML model that captures the structural data commonality across various data distributions. }
{\color{black}In particular, it results in a set of global parameters that can be better adapted/personalized to local data distributions with an additional gradient descent step, denoted by term (a) in~\eqref{eq:newMet}.} 
We approximate the gradient $\nabla F_j$ by mini-batch methods: 
\vspace{-1mm}
\begin{align}
\label{eq:pfl_combined_minibatches}
\nabla \widetilde{F}_j(\mathbf{w}_{j}(t)) &= 
\nabla\widetilde{f}_j\bigg(\hspace{-.5mm}\mathbf{w}_j(t) - \eta_{_{1}} \nabla \widetilde{f}_j(\mathbf{w}_j(t) \vert \mathcal{D}_{j,1}(t) ) \bigg\vert \mathcal{D}_{j,2}(t)\hspace{-.5mm}\bigg) \nonumber \\
& \qquad \cdot \bigg(\mathbf{I} - \eta_{_{1}} 
\nabla^2 \widetilde{f}_j(\mathbf{w}_j(t) \vert  \mathcal{D}_{j,3}(t))\bigg),
\vspace{-1mm}
\end{align} 
where {\color{black} $\widetilde{f}_j$ is the mini-batch loss defined similarly to~\eqref{eq:worker_loss} over a specific data batch}\footnote{The second argument in $\widetilde{f}_j(.|.)$ denotes the data batch used to compute the respective function.},  $\nabla^2$ is the Hessian operator, and $\mathcal{D}_{j,1}(t)$, $\mathcal{D}_{j,2}(t)$, $\mathcal{D}_{j,3}(t)$ are three independent mini-batches sampled with replacement from $\mathcal{D}_j(t)$.
We denote the mini-batch sampling ratios as $\alpha_{j,1}(t)$, $\alpha_{j,2}(t)$, and $\alpha_{j,3}(t) \in (0, 1)$, i.e., $D_{j,i}(t) = \alpha_{j,i}(t) D_j(t)$, $i=1,2,3$, and the total ratio of data at time $t$ used for processing by worker $j \in \mathcal{W}_u$, $u \in \mathcal{U}(s)$ is denoted by $\alpha_j(t) = \alpha_{j,1}(t) + \alpha_{j,2}(t) + \alpha_{j,3}(t)$.

\begin{remark}\label{rem:1}
The computational burden of the Hessian-based gradient descent in~\eqref{eq:pfl_combined_minibatches} can be alleviated by substituting it with approximate methods such as \textit{first order} or \textit{hessian free} model agnostic training~\cite{finn2017model, fallah2020personalized}, which have been shown to attain a similar performance to that of exact Hessian computation.
\end{remark}


{\tt HN-PFL} performs a series of local and global aggregations during each training sequence $\mathcal{T}_s$.
We conduct local aggregations with period of $\tau_s^{\mathsf{L}}$ (i.e., $\tau_s^{\mathsf{L}}$ local meta-gradient updates prior to each local aggregation), and global aggregations with period of $\tau_s^{\mathsf{L}} \tau_s^{\mathsf{G}}$, defined such that $T_s \triangleq \tau_s^{\mathsf{L}} K_s^{\mathsf{L}} \equiv \tau_s^{\mathsf{L}} \tau_s^{\mathsf{G}} K_s^{\mathsf{G}}$, where $K_s^{\mathsf{L}}$ is the total number of local aggregations, and $K_s^{\mathsf{G}}$ is the total number of global aggregations conducted in $\mathcal{T}_s$. 
Using $k$ for the local aggregation index, $t_s^{\mathsf{L}}(k) = t_s + k \tau_s^{\mathsf{L}}$ is the time of the $k$-th local aggregation in $\mathcal{T}_s$. Using $k'$ as the global aggregation index, $t_s^{\mathsf{G}}(k') = t_s + k' \tau_s^{\mathsf{L}} \tau_s^{\mathsf{G}}$ will denote the time of the $k'$-th global aggregation in $\mathcal{T}_s$. 

{\color{black} 
Within a swarm, workers' datasets will be controlled through a device-to-UAV offloading optimization procedure in Sec. III. Therefore, the swarm averaged ML model should be weighted towards those worker UAVs with the most processed data, as they are likely to have better trained ML models.
So, at the $k$-th \textit{local aggregation} when $t = t_s^{\mathsf{L}}(k)$, {\tt HN-PFL} performs a weighted average at each leader $\ell_u$, $\forall u$:} 
\begin{equation} \label{eq:swarm_aggregation_rule}
\overline{\mathbf{w}}_u(t) = \frac{\sum_{j \in \mathcal{W}_u} \mathbf{w}_j(t) \sum_{t' = t_s^{\mathsf{L}}(k-1) + 1}^{t} \alpha_j(t') D_j(t')}{\sum_{j \in \mathcal{W}_u} \sum_{t' = t_s^{\mathsf{L}}(k-1) +1}^{t} \alpha_j(t') D_j(t')}.
\end{equation}
Leader $\ell_u$ then broadcasts $\overline{\mathbf{w}}_u(t_s^{\mathsf{L}}(k))$ to all worker UAVs $\mathcal{W}_u$. 
This completes our \textit{swarm-wide aggregation}, and we define the swarm-wide/local meta-function $\overline{F}_u$ for $t = t_s^{\mathsf{L}}(k)$ as follows: 
\vspace{-1mm}
\begin{equation} \label{eq:local_meta_function}
\hspace{-.1mm}\resizebox{.92\linewidth}{!}{$
   \overline{F}_{u}(\overline{\mathbf{w}}_u(t)) =  \frac{\sum_{j \in \mathcal{W}_u} F_j(\overline{\mathbf{w}}_u(t)) \sum_{t' = t_s^{\mathsf{L}}(k-1) +1}^{t} \alpha_j(t') D_j(t')}{\sum_{j \in \mathcal{W}_u} \sum_{t' = t_s^{\mathsf{L}}(k-1) +1}^{t} \alpha_j(t') D_j(t')}.$}\hspace{-4mm}
   \vspace{-1mm}
\end{equation}

After $\tau_s^{\mathsf{L}}$ swarm-wide aggregations at the active swarms, the leaders ${\ell}_u$, $u\in \mathcal{U}(s)$, travel to their nearest AP, and transmit their swarm-wide parameters to the core network. At the $k'$-th \textit{global aggregation}, when $t = t_s^{\mathsf{G}}(k')$, the core network determines the global model parameters as:
\vspace{-1mm}
\begin{equation}\label{eq:global_aggregation_rule}
\overline{\mathbf{w}}(t) = \frac{1}{U(s)} \sum_{u \in \mathcal{U}(s)} \overline{\mathbf{w}}_u(t), 
\vspace{-1mm}
\end{equation}
with global meta-function $\overline{F}$ for $t = t_s^{\mathsf{G}}(k')$ as:
\vspace{-2mm}
\begin{equation} \label{eq:global_meta_function}
    \overline{F}(\overline{\mathbf{w}}(t)) = \frac{1}{U(s)} \sum_{u \in \mathcal{U}(s)} \overline{F}_u(\overline{\mathbf{w}}(t)).
    \vspace{-1mm}
\end{equation}
{\color{black}
Note that~\eqref{eq:global_aggregation_rule}\&\eqref{eq:global_meta_function} employ an unweighted aggregation across swarms, as opposed to the weighted aggregation~\eqref{eq:swarm_aggregation_rule}\&\eqref{eq:local_meta_function} within swarms. 
This is due to the fact that the distributions of processed data will be non-i.i.d. across swarms. 
\eqref{eq:global_aggregation_rule}\&\eqref{eq:global_meta_function} allow the determination of a global model that uses the data across all the active clusters without bias towards any specific cluster and data distribution. 
A global model formed via unbiased meta-function aggregation seeks to capture the structural data commonalities across different UAV swarms, which allows both efficient model personalization and adaptability to online ML environments at device clusters.} 
The access points $\mathcal{A}$ then broadcast $\overline{\mathbf{w}}(t_s^{\mathsf{G}}(k'))$ to the leaders, which return to their swarms and synchronize worker parameters. 

{\color{black} At the final global aggregation of training sequence $s$, i.e., $t = t_s + T_s - 1$, swarms in $\mathcal{U}(s)$ \textit{personalize} the global model to their respective visiting device cluster data distribution by performing a \textit{single} stochastic gradient update (accounting for term (a) in~\eqref{eq:newMet}) followed by a swarm-wide aggregation as in~\eqref{eq:swarm_aggregation_rule}, and then swarm leaders transfer the resulting ML model to the devices. 
Next, active swarms from both $s$ and $s+1$ initialize the next sequence. For each $u\in \mathcal{U}(s)\cup \mathcal{U}(s+1)$, leader $\ell_u$ will receive (i) final global parameters $\overline{\mathbf{w}}(t_s + T_s - 1)$, and (ii) trajectory decisions from the APs. Swarms then travel to their next cluster (for swarms $u \in \mathcal{U}(s) \setminus \mathcal{U}(s+1)$, this would be a recharging station). Upon arrival, swarm leaders $l_u$, $u \in \mathcal{U}(s+1)$, will begin training sequence $s+1$ by synchronizing workers with the latest global parameters.
Our developed {\tt HN-PFL} algorithm is summarized in Fig.~\ref{fig:PFL_algo}. 
}

{\color{black} We assume that data distributions at the IoT devices are stationary during training sequences, but vary between training sequences. 
Distribution changes at clusters may lead to worse ML performance (i.e., a weaker ML model) over time, which incentivizes swarms to re-visit and re-calibrate their ML models. 
We model online data distributions at clusters by introducing heterogeneous \textit{model/concept drift}, which primarily affects the UAV trajectories (see Sec.~\ref{sec:modelDrift})}. 
{\color{black} We next obtain the convergence bound of {\tt HN-PFL}, which will be employed in the optimization formulation in Sec.~\ref{sec:netwrok_pfl}.}

\subsection{Convergence Analysis of {\tt HN-PFL}}
\label{ssec:convergence}
In the following, we derive the performance bound of {\tt HN-PFL} for non-convex loss functions. 
To this end, in addition to the meta functions defined in~\eqref{eq:local_meta_function} and~\eqref{eq:global_meta_function}, we define the swarm-average loss function $\overline{f}_u(\mathbf{w})$ at $t \in \mathcal{T}_s$ as:
\begin{equation} \label{eq:swarm-loss}
     \overline{f}_u(\mathbf{w}(t)) = 
     \sum_{j \in \mathcal{W}_u} \frac{\Delta_{j}(k)}{\overline{\Delta}_{u}(k)} f_j(\mathbf{w}(t)),
\end{equation}
where $f_j$ is defined as in~\eqref{eq:worker_loss}, $k$ is the most recent local aggregation index, $\Delta_{j}(k) = \sum_{t' = t_s^{\mathsf{L}}(k-1)+1}^{t} \alpha_j(t') D_j(t')$,
and
    $\overline{\Delta}_{u}(k) = \sum_{j \in \mathcal{W}_u} \Delta_{j}(k)$.
Also, we define the global average loss function $\overline{f}(\mathbf{w})$ at time $t \in \mathcal{T}_s$ as:
\begin{equation} \label{eq:global-loss}
\overline{f}(\mathbf{w}(t)) = \frac{1}{U} \sum_{u \in \mathcal{U}} \overline{f}_u(\mathbf{w}(t)).
\end{equation}

While~\eqref{eq:swarm_aggregation_rule},\eqref{eq:local_meta_function},\eqref{eq:swarm-loss} are only realized by {\tt HN-PFL} at a local aggregation, $t = t_s^{\mathsf{L}}(k)$, and~\eqref{eq:global_aggregation_rule},\eqref{eq:global_meta_function},\eqref{eq:global-loss} are only realized at a global aggregation, $t = t_s^{\mathsf{G}}(k^{'})$, defining them for each $t \in \mathcal{T}_s$ will be useful in our theoretical analysis. 

Our convergence analysis employs some standard assumptions~\cite{fallah2020convergence} on non-convex loss functions:

\begin{assumption}[Loss function characteristics] We make the following assumptions on $f_j$ at worker $j \in \mathcal{W}_u$, $\forall u \in \mathcal{U}$:
\label{eq:assumptions_all}

\begin{enumerate}[leftmargin=5mm]
    \item $f_j$ is bounded below: $f_j(\mathbf{w}) > -\infty, \forall \mathbf{w}$. \label{eq:assumption_bounded_loss}
    
    \item \label{eq:assumption_loss_smooth_twice_differentiable}
    {\color{black} The gradient of $f_j$ is $\mu^{\mathsf{G}}_j$-Lipschitz and bounded by $B_j$}: (i) $\Vert \nabla f_j(\mathbf{w}) - \nabla f_j(\mathbf{w}') \Vert \leq \mu^{\mathsf{G}}_j \Vert \mathbf{w}-\mathbf{w}' \Vert$, $\forall \mathbf{w},\mathbf{w}'$, (ii) $\Vert \nabla f_j(\mathbf{w}) \Vert \leq B_j$. 
    
    \item \label{eq:assumption_hessian_lipschitz}
    {\color{black} $f_j$ is twice continuously differentiable, and the Hessian of $f_j$ is $\mu^{\mathsf{H}}_j$-Lipschitz continuous}: $\Vert \nabla^2 f_j(\mathbf{w}) - \nabla^2 f_j(\mathbf{w}')\Vert \leq \mu^{\mathsf{H}}_j \Vert \mathbf{w} - \mathbf{w}' \Vert$, $\forall \mathbf{w},\mathbf{w}'$.
    
    \item Bounded local data variability: 
       $\mathbb{E}[\Vert \nabla \widehat{f}(x,\mathbf{w}) - \nabla f_j(\mathbf{w}) \Vert^2] \leq \sigma^{\mathsf G}_{j}$
    and 
        $\mathbb{E}[\Vert \nabla^2 \widehat{f}(x,\mathbf{w}) - \nabla^2 f_j(\mathbf{w}) \Vert^2] \leq \sigma^{\mathsf H}_{j},~\forall \mathbf{w}, x$.
    \label{eq:assumption_stochastic_gradient_sigma}
    
    \item 
    The variance of the gradient and Hessian of $f_j(\mathbf{w})$ are bounded: 
    $\sum_{j \in \mathcal{W}_u}
    \frac{  \Delta_{j}(k)}{\overline{\Delta}_{u}(k)} 
    \Vert
    \nabla f_j(\mathbf{w}) - \nabla \overline{f}_u(\mathbf{w}) 
    \Vert^2  
    \leq \gamma_{u}^{\mathsf G}$,
        $\sum_{j \in \mathcal{W}_u} \frac{\Delta_{j}(k)}{\overline{\Delta}_{u}(k)}
        \Vert         
        \nabla^2 f_j(\mathbf{w}) - \nabla^2 \overline{f}_u(\mathbf{w}) 
        \Vert^2
        \leq \gamma_u^{\mathsf H}, \forall \mathbf{w}, k$.
    \label{eq:assumption_bounded_grad_hessian_gamma}
    
    \item \label{eq:assumption_bounded_grad_hessian_gamma_global}
    The variance of the gradient and Hessian of $\overline{f}_u(\mathbf{w})$ are bounded: 
    $\frac{1}{U} \sum_{u \in \mathcal{U}}  
    \Vert
    \nabla \overline{f}_u(\mathbf{w}) - \nabla \overline{f}(\mathbf{w}) 
    \Vert^2 
    \leq \gamma^{\mathsf G}$,
    $\frac{1}{U} \sum_{u \in \mathcal{U}} 
    \Vert
    \nabla^2 \overline{f}_u(\mathbf{w}) - \nabla^2 \overline{f}(\mathbf{w}) 
    \Vert^2 
    \leq \gamma^{\mathsf H}, \forall \mathbf{w}$.

\end{enumerate}
\end{assumption}

We also require an assumption to characterize the loss function behavior at the device clusters:
\vspace{-1mm}
\begin{assumption} [Device cluster loss function characteristics] 
\label{theory:untrained_clusters}
For device cluster $c$ and a model parameter $\mathbf w$, let $f^{\mathsf{D}}_c(\mathbf{w}) = {\sum_{x \in \widetilde{\mathcal{D}}_c(t)} \hat{f}(x;\mathbf{w})}/{\widetilde{D}_c(t)}$, where $\widetilde{\mathcal{D}}_c(t) \triangleq \cup_{i \in c} \widetilde{D}_i(t)$,  denote the local loss at time $t$.
We assume that (i) the gradient of the loss function is $\mu^{\mathsf{G}}_{c}$-Lipschitz continuous, (ii) the Hessian is $\mu^{\mathsf{H}}_{c}$-Lipschitz continuous, (iii) $\mathbb{E}[\Vert \nabla \hat{f}(x,\mathbf{w}) - \nabla f_{c}^{\mathsf{D}}(\mathbf{w}) \Vert^2] \leq \sigma_{c}^{\mathsf{G}}$ $\forall x, \mathbf{w}$, and (iv) $\mathbb{E}[\Vert \nabla^2 \hat{f}(x,\mathbf{w}) - \nabla^2 f_{c}^{\mathsf{D}}(\mathbf{w}) \Vert^2] \leq \sigma_{c}^{\mathsf{H}}$ $\forall x, \mathbf{w}$.
\end{assumption}

The loss functions of many common ML models (e.g., neural networks with continuous activation functions~\cite{wang2019adaptive}) will satisfy these assumptions. In the following analysis, we let $B = \max_j\{B_j\}$, $\mu^{\mathsf{G}} = \max_{j} \{\mu_j^{\mathsf{G}} \}$, and $\mu^{\mathsf{H}} = \max_{j} \{ \mu_j^{\mathsf{H}} \}$.

Our main result in this section will be the convergence behavior of the global meta-function $\overline{F}$ in {\tt HN-PFL} (Theorem~\ref{theory:theorem}). To obtain this, we first derive bounds on the expected error of the meta-gradient approximations at worker UAVs (Lemma~\ref{theory:lemma_sigma}), and on the meta-gradient variability between workers, leaders, and the core network (Lemma~\ref{theory:lemma_gamma}). 
\vspace{-1mm}
\begin{lemma}[Mini-batch versus full batch meta-gradients] \label{theory:lemma_sigma}
During each training interval $\mathcal{T}_s$, the expected error of the mini-batch approximation $\nabla \widetilde{F}_j(\mathbf{w}(t))$ of the true meta-gradient $\nabla F_j(\mathbf{w}(t))$ at each worker UAV $j \in \mathcal{W}_u$, $\forall u \in \mathcal{U}(s)$ satisfies:
\vspace{-1mm}
\begin{equation} \label{eq:lemma1}
\mathbb{E} [ \Vert \nabla \widetilde{F}_j(\mathbf{w}(t)) - \nabla F_j(\mathbf{w}(t)) \Vert^2 ] \leq \sigma_j^{\mathsf{F}}(t),
\vspace{-1mm}
\end{equation}
where
\vspace{-1mm}
\begin{align}
\sigma_j^{\mathsf{F}}(t) &=
\underbrace{\frac{3 \eta_{_{1}}^2 \sigma_j^{\mathsf{H}}}{\alpha_{j,3}(t)D_j(t)}}_{(a)}
\underbrace{\bigg[B^2 + \frac{\sigma_j^{\mathsf{G}}\left(\alpha_{j,1}(t) + (\mu^{\mathsf{G}} \eta_{_{1}})^2 \alpha_{j,2}(t)\right)}{\alpha_{j,1}(t)\alpha_{j,2}(t)D_j(t)} \bigg]}_{(b)} \nonumber \\
& \quad + \underbrace{\frac{12\sigma_j^{\mathsf{G}} \left( \alpha_{j,1}(t) + (\mu^{\mathsf{G}} \eta_{_{1}})^2 \alpha_{j,2}(t) \right) }{\alpha_{j,1}(t)\alpha_{j,2}(t)D_j(t)}}_{(c)}.
\label{eq:lemma1-1}
\vspace{-1mm}
\end{align}
\end{lemma}
\begin{proof}
See Appendix~\ref{app:sk_l1}.
\end{proof}


We make a few observations from Lemma~\ref{theory:lemma_sigma} regarding the mini-batch ratios $\alpha_{j,1}(t)$, $\alpha_{j,2}(t)$, and $\alpha_{j,3}(t)$. Intuitively, if any of these are $0$, then the upper bound in~\eqref{eq:lemma1} should diverge, which is what we observe in~\eqref{eq:lemma1-1}.
Next, the groups $(a), (b),$ and $(c)$ in~\eqref{eq:lemma1-1} show that each mini-batch ratio has a unique impact on the bound. In particular, $\alpha_{j,3}(t)$ weighs the Hessian data variability $\sigma_j^{\mathsf{H}}$ in $(a)$, while $\alpha_{j,1}(t)$ and $\alpha_{j,2}(t)$ scale the gradient data variability $\sigma_j^{\mathsf{G}}$ in $(b)$ and $(c)$. Due to the multiplicative effect of the Hessian and gradient in the meta-gradient computation (see~\eqref{eq:pfl_combined_minibatches}), the effects of the mini-batch ratios are coupled between $(a)$ and $(b)$, through which $\alpha_{j,3}(t)$ also weights the impact of $\sigma_j^{\mathsf{G}}$. Hence, when we are faced with a limited budget for data processing in a UAV swarm, the mini-batch ratios must be allocated carefully, which we will address through our optimization problem in Sec.~\ref{ssec:netaware}. 
\vspace{-1mm}
\begin{lemma}[Intra- and inter-swarm meta-gradient variability]
\label{theory:lemma_gamma}
During each training interval $\mathcal{T}_s$, the intra-swarm meta-gradient variability at each UAV swarm $u \in \mathcal{U}(s)$ obeys the following upper bound at local aggregation $k$:
\vspace{-1mm}
\begin{equation}
\begin{aligned}
&
\sum_{j \in \mathcal{W}_u}
\frac{\Delta_{j}(k)}
{\overline{\Delta}_{u}(k)}
\left\Vert 
\nabla F_j(\mathbf{w}) - \nabla \overline{F}_u(\mathbf{w}) \right\Vert^2
\leq \gamma_u^{\mathsf{F}},
\end{aligned}
\vspace{-1mm}
\end{equation}
where
$
\gamma_u^{\mathsf{F}} = 3B^2\eta_{_{1}}^2 \gamma_u^{\mathsf{H}} + 192 \gamma_u^{\mathsf{G}}.
$
Also, the variability of inter-swarm meta-gradients is upper bounded as follows:
\vspace{-1mm}
\begin{equation}
\begin{aligned}
&
\frac{1}{U(s)} \sum_{u \in \mathcal{U}(s)} 
\left\Vert
\nabla \overline{F}_u(\mathbf{w}) - \nabla F(\mathbf{w}) 
\right\Vert^2  
\leq \gamma^{\mathsf{F}},
\end{aligned}
\vspace{-1mm}
\end{equation}
where
$
\gamma^{\mathsf{F}} = 3B^2\eta_{_{1}}^2 \gamma^{\mathsf{H}} + 192 \gamma^{\mathsf{G}}.
$

\end{lemma}
\begin{proof}
See Appendix~\ref{app:sk_l2}.
\end{proof}
\vspace{-1mm}

Using Lemmas~\ref{theory:lemma_sigma} and~\ref{theory:lemma_gamma}, we can bound the variance of {\color{black} {\tt HN-PFL}} model parameters across worker UAVs after a given global aggregation, which is one of our key theoretical results. 
For brevity, all proofs are given as sketches, with the key steps emphasized. The full proofs are provided in the appendices. 
\vspace{-1mm}
\begin{proposition}[Parameter variability across UAVs] \label{theory:prop_l2}
During interval $\mathcal{T}_s$, the variance in model parameters across the active worker UAVs in the network at the $k'$-th global aggregation, i.e., $t = t_s^{\mathsf{G}}(k')$, satisfies the following upper bound:
\begin{equation} \label{eq:l2_result}
\hspace{-.1mm}
\resizebox{.92\linewidth}{!}{$
 \mathbb{E}\bigg[ \bigg\Vert
\frac{1}{U(s)}\sum\limits_{u \in \mathcal{U}(s)} \sum\limits_{j \in \mathcal{W}_u} 
\frac{\Delta_{j}(k'\tau_{s}^{\mathsf{G}})}{\overline{\Delta}_{u}(k'\tau_{s}^{\mathsf{G}})}
\bigg(\hspace{-1mm}\mathbf{w}_j(t) - \overline{\mathbf{w}}(t)\hspace{-1mm}\bigg) 
\bigg\Vert^2 \bigg] \hspace{-1mm}\leq \Upsilon(t),
 $}\hspace{-6mm}
\end{equation}
where $\Upsilon(t)$ is defined in~\eqref{eq:upsilon_def} with
$\mu^{\mathsf F}\triangleq 4\mu^{\mathsf{G}} + \eta_{_{1}} \mu^{\mathsf{H}} B$, and $\sigma_u^{\mathsf{F}}(t) = \sum_{j \in \mathcal{W}_u} 
\frac{\Delta_j(k'\tau_{s}^{\mathsf{G}}) }{\overline{\Delta}_{u}(k'\tau_{s}^{\mathsf{G}})}
\sigma_j^{\mathsf{F}}(t)$.

\begin{table*}[t]
      \vspace{-.5mm}
        \begin{equation}\label{eq:upsilon_def}
        \begin{aligned} 
\Upsilon(t) =&
\underbrace{\bigg(
 \frac{16 \eta_{_{2}}^2}{U(s)}\sum_{u \in \mathcal{U}(s)} \sum_{j \in \mathcal{W}_u} \frac{\Delta_{j}(k' \tau_s^{\mathsf{G}})}{\overline{\Delta}_{u}(k' \tau_s^{\mathsf{G}})}
\sum_{y=1}^{\tau_s^{\mathsf{L}}}
\sigma_j^{\mathsf{F}}(t - y)
+
\frac{24 \eta_{_{2}}^2}{U(s)} \sum_{u \in \mathcal{U}(s)}\gamma_u^{\mathsf{F}}
\bigg)
\frac{1 - (8 + 48 (\eta_{_{2}} \mu^{\mathsf{F}})^2)^{\tau_s^{\mathsf{L}}}}{1- (8 + 48 (\eta_{_{2}} \mu^{\mathsf{F}})^2)}}_{(a)}
\\[-0.8em]&+
\underbrace{\bigg(
 \frac{16 \eta_{_{2}}^2}{U(s)}\sum_{u \in \mathcal{U}(s)} \sum_{y=1}^{\tau_{s}^{\mathsf{L}}\tau_{s}^{\mathsf{G}}} \sigma_u^{\mathsf{F}}(t - y) 
 +
24 \eta_{_{2}}^2 \gamma^{\mathsf{F}}
\bigg)
\frac{1 - (8 + 48 (\eta_{_{2}} \mu^{\mathsf{F}})^2)^{\tau_{s}^{\mathsf{L}} \tau_{s}^{\mathsf{G}}}}{1- (8 + 48 (\eta_{_{2}} \mu^{\mathsf{F}})^2)}}_{(b)}
\end{aligned}
      \vspace{-1.5mm}
        \end{equation}
        \hrulefill
        \vspace{-2mm}
\end{table*}

\end{proposition}
\begin{proof}
See Appendix~\ref{app:prop2}.
\end{proof}


Considering the two terms in~\eqref{eq:upsilon_def}, term $(a)$ captures the impact of the local aggregation frequency $\tau_s^{\mathsf{L}}$ while term $(b)$ captures the \textit{nested} impact of the local-global aggregation frequency  $\tau_s^{\mathsf{L}} \tau_s^{\mathsf{G}}$. 
It can be seen that, to reduce the variance of ML models $\mathbf{w}_j$ across workers $j \in \mathcal{W}_u$ $\forall u \in \mathcal{U}(s)$, which is desirable as we will show in Theorem~\ref{theory:theorem}, the local or global aggregation periods $\tau_s^{\mathsf{L}}, \tau_s^{\mathsf{G}}$ need to be reduced, i.e., more frequent aggregations. 
Proposition~\ref{theory:prop_l2} suggests that the improvement is exponential. 
However, global aggregations consume more network resources than local aggregations in {\tt HN-PFL} - they require bidirectional communications between worker and leader UAVs, and between leaders and APs during which non-leader UAVs continue to consume energy by remaining idle in the air. We balance this trade-off by obtaining the optimal UAV orchestration for a choice of $\tau_s^{\mathsf{L}}$ and $\tau_s^{\mathsf{G}}$ (Sec.~\ref{sec:netwrok_pfl}), and then jointly optimizing $\tau_s^{\mathsf{L}}$ and $\tau_s^{\mathsf{G}}$ and UAV trajectories under optimal UAV orchestration (Sec.~\ref{sec:swarmTraj}).

{\color{black}Additionally, Proposition~\ref{theory:prop_l2} captures the impact of non-i.i.d. data distributions on parameter variability. In particular, we see that $\Upsilon(t)$ is directly proportional to the intra-swarm meta-gradient variability $\gamma_u^F$, the inter-swarm variability $\gamma^F$, and the mini-batch approximation error (the $\sigma_j^F(t)$ and $\sigma_u^F(t)$ terms). For the same setting of system control parameters, each of these will increase with the level of data heterogeneity present across devices, both within and between clusters.}

Finally, we apply Proposition~\ref{theory:prop_l2} to obtain our main result, which characterizes the decreasing magnitude of the gradient of~\eqref{eq:global_meta_function}, i.e., the global meta-function in {\tt HN-PFL}, over training sequences. Since we consider non-convex ML models, the main metric of interest for learning performance is the norm squared of the meta-gradient.
\vspace{-1mm}
\begin{theorem} \label{theory:theorem}[Global meta-gradient over training sequences] For training sequence $\mathcal{T}_s$, if  $\eta_2 < \frac{1}{6\mu^{\mathsf{F}}}$, we have the following upper bound on the expected cumulative average magnitude of the global meta-gradient across the active UAV swarms:
\vspace{-2mm}
\begin{equation}\label{eq:th1Main}
\begin{aligned}
\frac{1}{T_s}\sum_{t = t_{s-1} }^{t_s-1} \mathbb{E}\left[\left\Vert \nabla \overline{F} (\mathbf{w}(t)) \right\Vert^2\right] \leq \Xi(s),
\end{aligned}
\vspace{-1.5mm}
\end{equation}
where $\Xi(s)$ is given in~\eqref{eq:Xi}.
\begin{table*}[t]
\begin{minipage}{0.99\textwidth}
    \vspace{-1mm}
\begin{equation}\label{eq:Xi}
\begin{aligned}
\hspace{-4mm}
\Xi(s) \triangleq &
{\frac{1}{\frac{\eta_{_{2}}}{2} - 6\eta_{_{2}}^2 \frac{\mu^{\mathsf{F}}}{2}}}
\vast[
{\frac{F^{\mathsf{U}} (\mathbf{w}(t_{s-1})) - (F^{\mathsf{U}})^{\star}}{T_s}}
+ 
\frac{1}{T_s} \sum_{k'=1}^{K^{\mathsf{G}}_{s}} \hspace{2mm} 
\sum_{k = (k'-1)\tau_{s}^{\mathsf{G}} +1}^{\tau_{s}^{\mathsf{G}} k'}
\hspace{2mm}
\sum_{t = (k-1)\tau_{s}^{\mathsf{L}} }^{\tau_{s}^{\mathsf{L}} k-1} 
\Bigg[
{\left(3\eta_{_{2}}^2 \frac{\mu^{\mathsf F}}{2} + \eta_{_{2}}\right)} \\
&
\times
\bigg( 
{\frac{1}{U(s)} \sum_{u \in \mathcal{U}(s)} \sum_{j \in \mathcal{W}_u} \frac{\Delta_j(k' \tau_s^{\mathsf{G}} ) }{\overline{\Delta}_u(k' \tau_s^{\mathsf{G}} )}
\sigma_j^{\mathsf{F}}(t)}
+ {(\mu^{\mathsf{F}})^2 \Upsilon(t_s^{\mathsf{G}}(k')) }
\bigg)
+
{3\eta_{_{2}}^2 \mu^{\mathsf{F}} \frac{1}{U(s)} \sum_{u \in \mathcal{U}(s)} \gamma_u^{\mathsf{F}}}
\Bigg]
\vast]
\end{aligned}
\vspace{-1.5mm}
\end{equation}
\hrulefill
\end{minipage}
      \vspace{-5mm}
\end{table*}
\end{theorem}
\begin{proof}
See Appendix~\ref{app:thm1}.
\end{proof}
\vspace{-1mm}

\textbf{Main takeaways.} Theorem~\ref{theory:theorem} yields a general bound on the average gradient for {\tt HN-PFL} with  time-varying mini-batch sizes. Smaller values of the bound are desired, as it indicates closeness to a stationary point.
This bound quantifies how several parameters (some controllable and others a factor of the environment) affect training performance. Specifically, the bound in~\eqref{eq:Xi} is dependent on the mini-batch ratios and processed data sizes at the UAVs (embedded in $\Delta_j$, and $\Delta_u$), the initial performance of the ML model for the $s$-th training sequence (embedded in $F^{\mathsf{U}}(\mathbf{w}(t_{s-1}))$), the data variability (embedded in $\sigma_j^{\mathsf{F}}$, $\sigma_u^{\mathsf{F}}$, and $\Upsilon$), the gradient/Hessian characteristics (embedded in $\mu^{\mathsf{F}}$, $\gamma_u^{\mathsf{F}}$, and $\gamma^{\mathsf{F}}$), the local/global aggregation periods (through the nested sums), and the inner and outer step-sizes ($\eta_{_1}$, $\eta_{_2}$).

Since {\tt HN-PFL} conducts ML model training at the UAVs, the bound in Theorem~\ref{theory:theorem} is based on the meta-functions defined at the UAVs. 
{\color{black} However, as the ML model is used by the devices and UAVs train on a subset of the cluster's total data, we need to bound the difference between the ML model performance at the UAVs and the devices.}
Henceforth, since the data at UAVs is assumed to be constant within a local aggregation period $k$, we refer to $D_j(t)$ $\forall j$ as $D_j(k)$.
\vspace{-1mm}
\begin{lemma}[Meta-gradient mismatch between clusters and swarms]\label{lemma:mismatch} Let $F_{c}^{\mathsf{D}}$ denote the meta-function defined based on $f^{\mathsf{D}}_c$ in Assumption~\ref{theory:untrained_clusters}, $\mathcal{C}(s)$ denote the set of actively trained device clusters for training sequence $s$, and $\overline{F}^{\mathsf{D}}(\mathbf{w}(t)) = \frac{1}{|\mathcal{C}(s)|} \sum_{c \in \mathcal{C}(s)} F_c^{\mathsf{D}} (\mathbf{w}(t))$, $t \in \mathcal{T}_s$, be the average meta-function for all actively trained device clusters for sequence $s$ for a given parameter $\mathbf{w}(t)$ at time $t$. 
The difference between the meta-gradient computed at the UAVs vs. those of their respective device clusters for local aggregation $k$ is bounded by:
\vspace{-1mm}
\label{theory: worker_vs_full_cluster}
\begin{equation} \label{eq:cluster_grad_bound}
\hspace{-.2mm}
\begin{aligned} 
& \mathbb{E} \Bigg[ \bigg\Vert \nabla \overline{F} (\mathbf{w}(t)) - \nabla \overline{F}^{\mathsf{D}}(\mathbf{w}(t)) \bigg\Vert^2 \Bigg]
\leq \widehat{\Xi}(k), ~ t=\tau_s^{\mathsf L}(k),
\end{aligned}
\hspace{-6mm}
\vspace{-1mm}
\end{equation}
where $\widehat{\Xi}(k)$ is given by:
\vspace{-1mm}
\begin{equation} \label{eq:xi_hat}
    \begin{aligned}
    & \widehat{\Xi}(k) \triangleq \frac{1}{U(s)} \sum_{u \in \mathcal{U}(s)} \sum_{j \in \mathcal{W}_u} \frac{\Delta_j(k)}{\overline{\Delta}_u(k)}
    \Bigg[\frac{3\eta_1^2 \sigma_{C(u,s)}^{\mathsf H}}{D_j(k)}\bigg[B^2 \\& + \frac{\sigma_{C(u,s)}^{\mathsf{G}} (1 + \mu^2 \eta_1^2 ) } { D_j(k) } \bigg] 
    + \frac{12\sigma_{C(u,s)}^{\mathsf{G}}( 1 + \mu^2 \eta_1^2 )  }{D_j(k) } \Bigg],
    \end{aligned}
    \vspace{-1mm}
\end{equation}
where $C(u,s)$ denotes the cluster which UAV swarm $u$ trains during $s$ and $\sigma_{.}^{\mathsf{H}}$ and $\sigma_{.}^{\mathsf{G}}$ are defined in Assumption~\ref{theory:untrained_clusters}. 
\vspace{-1mm}
\begin{proof}
See Appendix~\ref{app:l3_proof}.
\end{proof}
\end{lemma}
\vspace{-1mm}
The bound in Lemma~\ref{theory: worker_vs_full_cluster} {\color{black} captures the effect of total data at the worker UAVs on the mismatch between the cluster and UAV swarm meta-gradients. In particular, as more data from the devices are transferred to the UAVs (increasing $D_j(k)$), the error in~\eqref{eq:xi_hat} due to sampling decreases.}

\vspace{-2mm}
\section{Data Processing Optimization for {\tt HN-PFL}}\label{sec:netwrok_pfl}
\noindent 
Given our {\tt HN-PFL} algorithm {\color{black} and its convergence behavior} in Sec.~\ref{ss:HN-PFL_only}, we next aim to develop efficient ML model training for our system. To this end, we need to obtain the optimal orchestration of UAVs once they engage in model training (micro-level design) and the swarm trajectories (macro-level design). 
We break these down into two main components: (i) data transfer and processing configurations at UAVs during the training sequences (the blue block in Fig.~\ref{fig:overall_flow}) and (ii) learning sequence duration and UAV swarm trajectory/movement patterns in-between training sequences (the green block in Fig.~\ref{fig:overall_flow}). These two parts are intertwined, i.e., the model training performance at clusters affects the UAV trajectory design, and vice versa.

{\color{black}In this section, we address the first component, 
by formulating the UAV data processing and transfer optimization problem, and subsequently develop the second component in Sec.~\ref{sec:swarmTraj}. 
{\color{black}Our formulation will account for the interplay between the overall energy cost of the system and the developed {\tt HN-PFL} performance metrics from Sec.~\ref{sec:HN-PFL} via a control optimization of key network parameters (Sec.~\ref{ssec:netaware})}. 
Then, we show that the resulting optimization can be characterized as a complementary geometric program, and develop an iterative \textit{distributed optimization} method for solving it (Sec.~\ref{ssec:gp-solve}).} 

\vspace{-3mm}
\subsection{Data Processing/Offloading Configuration}
\label{ssec:netaware}
\subsubsection{Offloading and processing models}
At each training sequence $s$, each active UAV swarm $u \in \mathcal{U}(s)$ is located above a device cluster denoted by $\mathcal{C}(u,s) \in \mathcal{C}$. Let $A(u,s) \in \mathcal{A}$ denote the nearest AP to UAV swarm $u$, which leader $\ell_u$ will periodically visit during the model training to conduct global model aggregation. 
Also, let $\psi_j^{\mathsf{F}}$ denote the energy consumed per unit time by worker/coordinator UAV $j \in \mathcal{W}_u$ for flying/hovering, and $\hat{\psi}_u^{\mathsf{F}}$ that of leader UAV $\ell_u$. We denote the geographical distance between leader $\ell_u$ and its nearest AP as $d(\ell_u,A)$, and denote the movement energy consumption of the leader UAV per unit distance as $\psi_{\ell_u}^{\mathsf{M}}$.

Device-UAV and UAV-UAV data transfers are carried out at the beginning of local aggregation rounds, i.e., when $t = t_s^{\mathsf{L}}(k)$.
The data received by the worker UAVs is used throughout the local aggregation period.
Upon engaging in data transmission, each device $i \in \mathcal{C}(u,s)$ samples data points uniformly at random from its local buffer and transmits them to the UAVs. We let $\rho_{i,j}(k)\in[0,1]$ denote the fraction of datapoints in the local dataset $\widetilde{D}_i(k)$ of the device that is transmitted to worker/coordinator UAV $j \in \widehat{u}$.
Coordinator UAVs act as data caches that facilitate multi-hop data relaying between the devices to worker UAVs used for ML model training. We let $\varrho_{h,j}(k)\in[0,1]$ denote the fraction of the local dataset at the coordinator UAV $h \in \widehat{\mathcal{W}}_u$ that is forwarded to another UAV $j \in \mathcal{W}_u$. Each worker UAV ${j} \in \mathcal{W}_u$ processes data with CPU frequency $g_j(k)\in [g^{\mathsf{min}}_j,g^{\mathsf{max}}_j]$.

For sequence $s$, we denote the transmit powers of device $i$, UAV $j$, and leader $\ell_u$ by $P_i(s)$, $P_j(s)$, and $P_{\ell_u}(s)$, respectively. Through transmissions, either data or model parameters are transferred. We denote the number of bits used to represent one data point as $\widetilde{M}$, and the number of bits used to represent the one model parameter vector as $M$.

UAV-to-UAV data transmissions are carried out through  air-to-air (A2A) channels, which are considered to be line-of-sight (LoS). Device-to-UAV and leader UAV-to-AP transmissions are performed through ground-to-air (G2A) and air-to-ground (A2G) channels, respectively, which are a mixture of LoS and non-line-of-sight (NLoS) links. Denoting $d(a,b)$ as the geographical distance between two nodes $a$ and $b$ ($a,b\in \mathcal{U} \cup \mathcal{C}$), the path-loss model for the LoS link between two nodes $a$ and $b$ is then given by~{\cite{al2014modeling}}:
\vspace{-1mm}
\begin{equation}\label{eq:A2A}
 L_{a,b}^{\textrm{A2A}}=\eta^{\mathsf{LoS}}(\mu^{\mathsf{Tx}} d(a,b))^{\alpha^{\mathsf{PL}}} ,
 \vspace{-1mm}
\end{equation}
where $\eta^{\mathsf{LoS}}>1$ 
denotes the excessive path loss factor for the LoS link,   $\alpha^{\mathsf{PL}}$ is the path-loss exponent, and $\mu^{\mathsf{Tx}} =  4\pi f^{\mathsf{Tx}} / c^{\mathsf{light}}$ with $c^{\mathsf{light}}$ 
denoting the speed of light and $f^{\mathsf{Tx}}$ denoting the carrier frequency.
For the G2A/A2G channel, the probability of having an LoS link between two nodes $a$ and $b$ is given by~{\cite{al2014modeling, mozaffari2017mobile}}:
$
P_{a,b}^{\textrm{LoS}}= \left({1+\psi^{\mathsf{Tx}} \exp(-\beta^{\mathsf{Tx}}[\theta_{ab}-\psi^{\mathsf{Tx}} ] ) }\right)^{-1},
$
where $\psi^{\mathsf{Tx}}$ and $\beta^{\mathsf{Tx}}$ are constants depending on the carrier frequency and the conditions of the environment, and $\theta_{a,b}$ is the elevation angle between the respective nodes defined as:
$
\theta_{a,b}=\frac{180}{\pi} \times \sin^{-1}\left(\frac{ \Delta h_{a,b}}{d(a,b)}\right),
$
with $\Delta h_{a,b}$ denoting the difference in altitude between nodes $a$ and $b$. The probability of NLoS link is given by $P^{\textrm{NLoS}}_{a,b}=1-P^{\textrm{LoS}}_{a,b}.$ With this, the path-loss of an A2G/G2A link from node $a$ to node $b$ can be obtained as:
\vspace{-1mm}
\begin{equation}\label{eq:G2A}
 L_{a,b}^{\textrm{A2G}}=(\mu^{\mathsf{Tx}} d(a,b))^{\alpha^{\mathsf{PL}}} [P^{\textrm{LoS}}_{a,b} \times \eta^{\mathsf{LoS}}+P^{\textrm{NLoS}}_{a,b} \times \eta^{\mathsf{NLoS}}],
 \vspace{-1mm}
\end{equation}
where $\eta^{\mathsf{NLoS}}>\eta^{\mathsf{LoS}}$ 
denotes the excessive path loss factor for the NLoS link.
 Finally, the data-rate between two nodes $a$ and $b$ is given by:
 \vspace{-1mm}
 \begin{equation}
     R_{a,b} = \bar{B}_{a,b} \log_2 (1+ \frac{P_a/L_{a,b}}{\sigma^2}), ~a,b\in \mathcal{U} \cup \mathcal{C},
     \vspace{-1mm}
 \end{equation}
where $\bar{B}_{a,b}$ denotes the bandwidth, $\sigma^2=N_0 \bar{B}_{a,b}$ denotes the noise power with $N_0$ as the noise spectral density, $P_a$ is the transmit power of node $a$, and $L_{a,b}$ is the path-loss obtained through either~\eqref{eq:A2A} or~\eqref{eq:G2A}. 
{\color{black}As the channel conditions may change over time, we incorporate the training sequence index into the data rate notation denoted by $R_{a,b}(s)$ to represent the data rate between two nodes $a$ and $b$ during the training sequence $s$.
To focus our optimization formulation on the interplay between ML convergence and energy consumption, we simplify the contributions of specific modulation and signaling schemes by ignoring interference from simultaneous transmissions to the UAVs as they are stationary and can use orthogonal frequency bands~\cite{7502656} for communications.}

We denote the time used for data gathering at UAV $j$ as $\zeta^{\mathsf{G}}_j(k)$, and the time used for data processing as $\zeta_{j}^{\mathsf{P}}(k)$. We also define $\zeta^{\mathsf{Local}}$ as the maximum allowable time for data gathering and local computation before each UAV transmits its parameters to the leader UAV for aggregation. 
 
\subsubsection{Joint energy and performance optimization}
With the aforementioned models in hand, we formulate the following optimization problem for determining data offloading/processing configuration at training sequence $s$:
\vspace{-2mm}

{\small
\begin{equation}
\hspace{-10mm}
    \begin{aligned}\label{eq:f1_obj}
&  \bm{(\mathcal{P})}:~ \hspace{-1mm}
\min_{\boldsymbol{\rho},\boldsymbol{\varrho},\boldsymbol{\alpha},\boldsymbol{g}} 
(1-\theta) \underbrace{\bigg(\theta_1 \Xi(s) + \theta_2\sum_{k=1}^{K_{s}^{\mathsf{L}}} \widehat{\Xi}(k) \bigg)}_{(a)}  \\
&
+ \theta \underbrace{\sum_{k=1}^{K_{s}^{\mathsf{L}}} \sum_{u \in \mathcal{U}(s) }\hspace{-1mm}\bigg[ 
\sum_{j \in \widehat{\mathcal{W}}_u} 
\hspace{-1mm} E^{\mathsf{Tx},\mathsf{U}}_j(k)
+ \sum_{j \in \mathcal{W}_u} \hspace{-1mm} E^{\mathsf{P}}_{j}(k) 
+ \sum_{i \in \mathcal{C}(u,s)} \hspace{-1.5mm} E^{\mathsf{Tx},\mathsf{C}}_i(k) \bigg]  
}_{(b)} 
\end{aligned}
\hspace{-10mm}
\end{equation}
}
\vspace{-0.3in}
\begin{align}
&\textrm{s.t.} \nonumber \\
& E^{\mathsf{P}}_{j}(k) = \tau_s^{\mathsf{L}} \frac{a_j c_j}{2} \left( \alpha_{j}(k) \right) D_{j}(k) {g_j^2(k)}, j \in \mathcal{W}_u, \forall u \label{eq:f1_con1}\\ 
& E^{\mathsf{Tx},\mathsf{C}}_{i}(k) = \sum_{j \in  \widehat{u}}  \frac{\rho_{i,j}(k) \widetilde{D}_i(k) \widetilde{M} }{R_{i,j}(s)} P_i(s), i \in \mathcal{C}(u,s), \forall u \label{eq:f1_con3} \\ 
& E^{\mathsf{Tx},\mathsf{U}}_{j}(k) = \sum_{h \in \mathcal{W}_u} \frac{\varrho_{j,h}(k) D_j(k) \widetilde{M} }{R_{j,h}(s)} P_j(s) , j \in \widehat{\mathcal{W}}_u, \forall u \label{eq:f1_con4}\\  
& E^{\mathsf{Tx},\mathsf{W}}_j = K_{s}^{\mathsf{L}}  P_j(s)M/R_{j,\ell_u}(s)  , j \in {\mathcal{W}}_u, \forall u \label{eq:f1_con5}\\ 
& E^{\mathsf{Tx},\mathsf{L}}_{\ell_u} = K_{s}^{\mathsf{L}} P_{\ell_u}(s) \max_{j \in \mathcal{W}_u}\Big\{ M/ R_{\ell_u,j}(s) \Big\} ,  \forall u \label{eq:f1_con6}\\ 
& E^{\mathsf{F,U}}_j = {T}_s  \psi_j^{\mathsf{F}},  j \in \widehat{u}, \forall \widehat{u} \label{eq:f1_con7}\\
& E^{\mathsf{F,L}}_{\ell_u} = {T}_s  \hat{\psi}_u^{\mathsf{F}} + 2 K_{s}^{\mathsf{G}} \psi_u^{\mathsf{M}}(d(\ell_u,A)), \forall u \label{eq:f1_con8}\\
& \sum_{k=1}^{K_{s}^{\mathsf{L}}} \hspace{-0.8mm} E^{\mathsf{P}}_j(k) \hspace{-0.8mm} + \hspace{-0.8mm} E^{\mathsf{Tx,W}}_j \hspace{-1.1mm} + \hspace{-0.8mm} E^{\mathsf{F,U}}_j \hspace{-0.8mm} \leq \hspace{-0.8mm} E^{\mathsf{Ba}}_{j}(s) - E^{\mathsf{Th}}_j , j \in \mathcal{W}_u, \forall u \hspace{-4.3mm} \label{eq:f1_con9}\\
& \sum_{k=1}^{K_{s}^{\mathsf{L}}} E^{\mathsf{Tx,U}}_j(k) + E^{\mathsf{F,U}}_j \leq E^{\mathsf{Ba}}_{j}(s) - E^{\mathsf{Th}}_j , j \in \widehat{\mathcal{W}}_u, \forall u \label{eq:f1_con10}\\
& E^{\mathsf{Tx,L}}_{\ell_u} + E^{\mathsf{F,L}}_{\ell_u}\leq E^{\mathsf{Ba}}_{\ell_u}(s) - E^{\mathsf{Th}}_{\ell_u} , \forall u \label{eq:f1_con11}\\
&  \sum_{j \in \widehat{u}} \rho_{i,j}(k) \leq 1, i \in \mathcal{C}(u,s), \forall u \label{eq:f1_con12} \\
& \sum_{h \in \mathcal{W}_u} \varrho_{j,h}(k) \leq 1, j \in \widehat{\mathcal{W}}_u, \forall u \label{eq:f1_con13} \\
& D_{j}(k) = \sum_{i \in \mathcal{C}(u,s)} \rho_{i,j}(k) \widetilde{D}_{i}(k) 
\nonumber \\&~~~~~~~~~~+ \sum_{h \in \widehat{\mathcal{W}}_u} \varrho_{h,j}(k) D_{h}(k) \leq B_j^{\mathsf{D}}(k), j \in \mathcal{W}_u, \forall u \label{eq:f1_con14}\\
& D_{j}(k) = \sum_{i \in \mathcal{C}(u,s)} \rho_{i,j}(k) \widetilde{D}_{i}(k) \leq B_j^{\mathsf{D}}(k), j \in \widehat{\mathcal{W}}_u, \forall u \label{eq:f1_con14b} \\ 
& \zeta^{\mathsf{G}}_j(k) + \zeta^{\mathsf{P}}_j(k) \leq \zeta^{\mathsf{Local}},  j \in u, \forall u \label{eq:f1_con17} \\ 
& \zeta^{\mathsf{G}}_j(k) = \sum_{i \in \mathcal{C}(u,s) } \rho_{i,j}(k) \widetilde{D}_i(k) \widetilde{M} / R_{i,j}(s) \nonumber \\&~~~~~~~~~~ + \sum_{h \in \widehat{\mathcal{W}}_u} \varrho_{h,j}(k) D_h(k) \widetilde{M} / R_{h,j}(s),  j \in \widehat{u}, \forall \widehat{u} \label{eq:f1_con16}
\\
& \zeta^{\mathsf{P}}_j(k) = \tau_s^{\mathsf{L}} c_j \alpha_{j}(k) D_j(k) / g_j(k), j \in \mathcal{W}_u, \forall u  \label{eq:f1_con15}\\
& g^{\mathsf{min}}_j \leq g_j(k) \leq g^{\mathsf{max}}_j, j \in {\mathcal{W}}_u, \forall u \label{eq:f1_con18} \\
& 0 \leq \rho_{i,j}(k) \leq 1, i \in \mathcal{C}(u,s), j \in \widehat{u}, \forall u \label{eq:f1_con19} \\
& 0 \leq \varrho_{h,j}(k) \leq 1, h \in \widehat{\mathcal{W}}_u, j \in \mathcal{W}_u \label{eq:f1_con20} \\ 
& \alpha_j(k) = \alpha_{j,1}(k) + \alpha_{j,2}(k) + \alpha_{j,3}(k),~ j \in \mathcal{W}_u, \forall u  \label{eq:f1_con21} \\ 
& 0 < \alpha_{j,1}(k), \alpha_{j,2}(k), \alpha_{j,3}(k) \leq 1, j \in \mathcal{W}_u, \forall u  \label{eq:f1_con2}
\end{align}
\textbf{Objective and variables.} The objective function in~\eqref{eq:f1_obj} captures the tradeoff between the expected ML model performance (term $(a)$) and the energy consumption during the training sequence (term $(b)$).
Term $(a)$ encompasses both the ML performance at the active UAV swarms (through $\Xi$) 
and the mismatch to the ML performance at the device cluster (through $\widehat{\Xi}$), which we quantified in Theorem~\ref{theory:theorem} and Lemma~\ref{lemma:mismatch}. We weigh the importance of the two terms in $(a)$ with normalized positive coefficients $\theta_1$ and $\theta_2$. 
The coefficient $\theta\in [0,1]$ weighs the importance of the terms in the objective function. The variables in the problem are the device-to-UAV data transfer configurations $\boldsymbol{\rho}=\{\rho_{i,j} \}_{i\in\mathcal C(u,s),j\in \widehat{u}, u\in \mathcal{U}(s)}$, the UAV-to-UAV data transfer configurations $\boldsymbol{\varrho}=\{\varrho_{h,j} \}_{h \in \widehat{\mathcal{W}}_u, j \in \mathcal{W}_u,u\in \mathcal{U}(s)} $, the mini-batch  ratios $\boldsymbol{\alpha}=\{\alpha_{j,1},\alpha_{j,2},\alpha_{j,3}\}_{j\in \mathcal{W}_u,u\in \mathcal{U}(s)}$, and the UAV CPU frequency cycles $\boldsymbol{g}=\{g_j\}_{j \in \mathcal{W}_u,u\in \mathcal{U}(s)}$.

For the energy objective terms, $E_{j}^{\mathsf{P}}(k)$ in~\eqref{eq:f1_con1} denotes the energy consumption used for \underline{p}rocessing data during each meta-gradient update at worker UAV $j \in \mathcal{W}_u$, where $a_j$ is the effective capacitance coefficient of UAV $j$'s chipset~\cite{9261995}, and $c_j$ is the number of CPU cycles to process each datapoint. 
In~\eqref{eq:f1_con3}, $E_{i}^{\mathsf{Tx}, \mathsf{C}}(k)$ is the energy used for data transmission (\underline{Tx}) from device $i$ in \underline{c}luster $\mathcal{C}(u,s)$ to the UAVs. 
In~\eqref{eq:f1_con4}, we define $E_{j}^{\mathsf{Tx},\mathsf{U}}(k)$ as the energy used for data transmissions from coordinator \underline{U}AV $j \in \widehat{\mathcal{W}}_u$ to other UAVs.  
In~\eqref{eq:f1_con5},  $E_{j}^{\mathsf{Tx},\mathsf{W}}$ denotes the total energy consumed for ML model parameter transmission from \underline{w}orker UAV $j \in \mathcal{W}_u$ to its associated leader, which occurs $K_{s}^{\mathsf{L}}$ times during the training sequence.
In~\eqref{eq:f1_con6}, $E_{\ell_u}^{\mathsf{Tx},\mathsf{L}}$ captures the energy consumption used for parameter transmission through broadcasting from \underline{l}eader $\ell_u$ to worker UAVs $j \in \mathcal{W}_u$.
In~\eqref{eq:f1_con7}, $E_{j}^{\mathsf{F},\mathsf{U}}$ represents the total energy used for \underline{f}lying/hovering by \underline{U}AV $j \in \widehat{u}$ for the $s$-th training sequence, which can be obtained using the result of~\cite{zhangEnergy} for fixed-wing UAVs,  and  the result of~\cite{zeng2019energyRotary} for rotatory wing UAVs.
Similarly, in~\eqref{eq:f1_con8}, $E_{\ell_u}^{\mathsf{F},\mathsf{L}}$ captures the energy consumption for \underline{f}lying by \underline{l}eaders $\ell_u$, accounting for levitation and round-trips to the nearest APs at the global aggregation instances.

\textbf{Constraints.} Constraint~\eqref{eq:f1_con9} ensures that the total energy consumed by each worker UAV $j$ for data processing, parameter transmission and flying is less than $E_{j}^{\mathsf{Ba}}(s) - E_{j}^{\mathsf{Th}}$, where $E_{j}^{\mathsf{Ba}}(s)$ represents the battery energy at $j$ at the start of the $s$-th training sequence and $E_{j}^{\mathsf{Th}}$ encompasses
both (i) surplus idle energy needed for extra hovering time caused by potential asynchronocity due to the heterogeneity of leader UAV to AP travel times, and (ii) the minimum energy threshold for $j$ to reach the nearest recharging station after the conclusion of the training sequence.
Constraint~\eqref{eq:f1_con10} imposed on the coordinator UAVs is similar to~\eqref{eq:f1_con9}, except that coordinator UAVs only conduct data transmission while flying.
Constraint~\eqref{eq:f1_con11} imposed on the leader UAVs guarantees that there is enough energy remaining after parameter broadcasting and flying to reach the nearest recharging station.
Constraints~\eqref{eq:f1_con12} and~\eqref{eq:f1_con13} ensure that the total amount of data offloaded from each device $i \in \mathcal{C}(u,s)$ and each coordinator UAV $h \in \widehat{\mathcal{W}}_u$ is less than the size of the available data set.
As a result of this offloading,~\eqref{eq:f1_con14} and~\eqref{eq:f1_con14b} capture the total number of datapoints at the UAVs.
Constraint~\eqref{eq:f1_con17} ensures that the accumulated time used for data gathering $\zeta_{j}^{\mathsf{G}}(k)$ and local data processing $\zeta_{j}^{\mathsf{P}}(k)$ at each worker UAV $j$ during one local aggregation is lower than the maximum allowable time for parameter transfer to the corresponding leader UAV $\zeta^{\mathsf{Local}}$.
These two quantities are in turn defined in~\eqref{eq:f1_con16} and~\eqref{eq:f1_con15}.
Finally, constraints~\eqref{eq:f1_con18}-\eqref{eq:f1_con2} ensure the feasibility of the optimization variables.

\textbf{Complementary geometric program.} $\bm{\mathcal{P}}$ is a non-convex problem, since products of the optimization variables are present in the objective. For instance, in the definition $E^{\mathsf{P}}_{j}(n)$ in~\eqref{eq:f1_con1}, there are multiplications between all four variable types: $\alpha$, $g$, $\rho$, and $\varrho$ (the last two are encompassed in term $D_j$ according to~\eqref{eq:f1_con14}). In fact, this problem can be categorized as a \textit{complementary geometric program}~\cite{chiang2007power}, a particular class of non-convex and NP-hard optimization problems. Based on this classification, we will develop a distributed, iterative approach to solve $\bm{\mathcal{P}}$ where iterative approximation of the problem in the format of a geometric program (GP) is considered. 

\begin{table*}[t]
\begin{minipage}{0.99\textwidth}
\begin{equation} \label{eq:actual_delta_u}
    \overline{\Delta}_u(k) \triangleq 
    \sum_{j \in \mathcal{W}_u} \bigg( \alpha_{j,1}(k) + \alpha_{j,2}(k) + \alpha_{j,3}(k) \bigg) \left(\sum_{i \in \mathcal{C}(u,s)} \rho_{i,j}(k) \widetilde{D}_i(k)
    + \sum_{h \in \widehat{\mathcal{W}}_u} \sum_{i \in \mathcal{C}(u,s)}  \varrho_{h,j}(k) \rho_{i,h}(k) \widetilde{D}_i(k) \right)
    \vspace{-3mm}
\end{equation}

\begin{equation} \label{eq:delta_u_approx}
    \hspace{-.1mm}
    \resizebox{.96\linewidth}{!}{$
    \begin{aligned}
    &  \overline{\Delta}_u(k)  \geq \widehat{\overline{\Delta}}_u(k)
    \triangleq 
    \prod_{j\in \mathcal{W}_u} \vast[
    \prod_{i \in \mathcal{C}(u,s)} 
    \Bigg[
    \bigg(\frac{ \widetilde{\delta}_{i,j,1}(k)
    \left[ \overline{\Delta}_u(k;m) \right] } 
    { \widetilde{\delta}_{i,j,1}(k;m)}\bigg)^{\frac{ \widetilde{\delta}_{i,j,1}(k; m) } {\overline{\Delta}_u(k; m) } }
    \bigg(\frac{ \widetilde{\delta}_{i,j,2}(k)
    \left[ \overline{\Delta}_u(k; m) \right] } 
    { \widetilde{\delta}_{i,j,2}(k; m)}\bigg)^{\frac{ \widetilde{\delta}_{i,j,2}(k; m) } {\overline{\Delta}_u(k; m) } }
    \bigg(\frac{ \widetilde{\delta}_{i,j,3}(k)
    \left[ \overline{\Delta}_u(k;m) \right] } 
    { \widetilde{\delta}_{i,j,3}(k;m)}\bigg)^{\frac{ \widetilde{\delta}_{i,j,3}(k; m) } {\overline{\Delta}_u(k; m) } } \Bigg] \\
    &  
    \prod_{h \in \widehat{\mathcal{W}}_u} \prod_{i \in \mathcal{C}(u,s)} 
    \Bigg[
    \bigg( \frac{ \delta_{j,h,i,1}(k)
    \left[ \overline{\Delta}_u(k;m ) \right] }
    { \delta_{j,h,i,1}(k; m ) } \bigg)^{ \frac{ \delta_{j,h,i,1}(k; m) } {\overline{\Delta}_u(k; m) } }
    \bigg( \frac{ \delta_{j,h,i,2}(k)
    \left[ \overline{\Delta}_u(k; m ) \right] }
    { \delta_{j,h,i,2}(k; m ) } \bigg)^{ \frac{ \delta_{j,h,i,2}(k; m) } {\overline{\Delta}_u(k; m) } } 
    \bigg( \frac{ \delta_{j,h,i,3}(k)
    \left[ \overline{\Delta}_u(k; m ) \right] }
    { \delta_{j,h,i,3}(k; m ) } \bigg)^{ \frac{ \delta_{j,h,i,3}(k; m) } {\overline{\Delta}_u(k; m) } }
    \Bigg]
    \vast],
    \end{aligned}
    $}\hspace{-3mm}
\end{equation}
where
\vspace{-1mm}
\begin{equation} \label{eq:little_delta_defs}
\widetilde{\delta}_{j,i,n}(k) \triangleq \alpha_{j,n}(k)\rho_{i,j}(k)\widetilde{D}_i(k),
~~\delta_{j,h,i,n}(k) \triangleq \alpha_{j,n}(k)\varrho_{h,j}(k) \rho_{i,h}(k) \widetilde{D}_i(k),~~n \in \{1,2,3\}.
\end{equation}
\vspace{-3mm}
\hrule
\end{minipage}
\end{table*}

\begin{table*}
\begin{minipage}{0.99\textwidth}
\vspace{-2mm}
\begin{equation}\label{eq:actualD_j}
    D_j(k) \triangleq \sum_{i \in \mathcal{C}(u,s)} \rho_{i,j}(k)\widetilde{D}_i(k) 
    + \sum_{h \in \widehat{\mathcal{W}}_u} \sum_{i \in \mathcal{C}(u,s)} \varrho_{h,j}(k) \rho_{i,h}(k) \widetilde{D}_i(k),
    \vspace{-2mm}
\end{equation}
\begin{equation}\label{eq:D_j_approx}
    \begin{aligned} 
    &  D_j(k)\geq 
    \widehat{D}_j(k) \triangleq 
    \prod_{i \in \mathcal{C}(u,s)} \bigg( \frac{  \widetilde{\lambda}_{i,j} (k)  
    D_j(k;m) }
    { \widetilde{\lambda}_{i,j} (k;m) }\bigg)^ {\frac{ \widetilde{\lambda}_{i,j}(k;m)}{D_j(k;m)}}
    \prod_{h \in \widehat{\mathcal{W}}_u} \prod_{i \in \mathcal{C}(u,s)} 
    \bigg(
    \frac{ \lambda_{h,i,j}(k)
    D_j(k; m) }
    { \lambda_{h,i,j}(k;m) }
    \bigg) ^ { \frac{ \lambda_{h,i,j}(k;m) }{
    D_j(k;m) }},
    \end{aligned}
    \end{equation}
where
\begin{equation} \label{eq:little_lambda_defs}
    \lambda_{i,j}(k) \triangleq \rho_{i,j}(k) \widetilde{D}_i(k),~~~ \widetilde{\lambda}_{h,i,j}(k) \triangleq \varrho_{h,j}(k) \rho_{i,h}(k) \widetilde{D}_i(k).
    \vspace{-2mm}
\end{equation}
\hrulefill
\end{minipage}
\vspace{-4mm}
\end{table*}

\vspace{-3mm}
\subsection{Distributed Algorithm for Data Processing/Offloading} 
\label{ssec:gp-solve}
To solve $\bm{\mathcal{P}}$, we first make two key observations:

\textbf{Observation 1: Complementary Geometric Programming.}
GP is a method for converting a non-convex optimization problem into convex form when the objective and constraints are composed of \textit{monomials} and \textit{posynomials}. We provide an overview of this for the interested reader in Appendix~\ref{app:gp}. Although the objective in $\bm{\mathcal{P}}$ is composed of multiplications between the variables, it does not follow the format of GP: the expressions of $\Xi$ and $\widehat{\Xi}$ consist of terms that are in the format of ratio of two posynomials, which are not posynomials (e.g., $\frac{\Delta_j}{\overline{\Delta}_u}$, 
$\Upsilon$, and $\sigma_j^{\mathsf{F}}$, each of which contain arithmetic relationships of $\rho$ and $\varrho$ in the denominators). Rather, $\bm{\mathcal{P}}$ is a complementary GP, which cannot readily be translated to convex form~\cite{chiang2007power}. We exploit a method based on \textit{posynomial condensation} to approximate the ratios of two posynomials appearing in the expressions of $\Xi$ and $\widehat{\Xi}$ (via approximating $\overline{\Delta}_u$ and $D_j$ where needed) as the ratio between a posynomial and a monomial.
Since the ratio of a posynomial and a monomial is a posynomial, we then can convert the approximations into geometric programs. 

\textbf{Observation 2: Potential of Distributed Implementation.}
In $\bm{\mathcal{P}}$, all of the constraints are separable with respect to the UAV swarm index $u$, and the objective function (including the terms $\Xi$ and $\hat{\Xi}$) can be written as a sum of separable functions with respect to the UAV swarm index. In practice, however, $\gamma^{\mathsf{F}}$ (defined in Lemma~\ref{theory:lemma_gamma}) and $\mu^{\mathsf{F}}$ (defined in Proposition~\ref{theory:prop_l2}) cannot be locally and independently computed by UAV swarms $u \in \mathcal{U}(s)$, as it depends on the maximum over all worker UAVs in the network. We can approximate $\gamma^{\mathsf{F}}$ and $\mu^{\mathsf{F}}$ by estimating it at the core network at the instance of global aggregations, and subsequently broadcast it to the leader UAVs though the APs. With the knowledge of $\gamma^{\mathsf{F}}$ and $\mu^{\mathsf{F}}$ at the swarm leaders, the problem can then be decomposed among the local aggregation instances $k \in \{1, \cdots, K_s^{\mathsf{L}}\}$ and solved distributively by each swarm $u \in \mathcal{U}(s)$ at its leader.
\begin{algorithm}[t]
 	\caption{Distributed network-aware optimization of {\tt HN-PFL} under partial global knowledge}\label{alg:cent}
 	\SetKwFunction{Union}{Union}\SetKwFunction{FindCompress}{FindCompress}
 	\SetKwInOut{Input}{input}\SetKwInOut{Output}{output}
 	 	{\footnotesize
 	\Input{Convergence criterion between two consecutive iterations. Estimated Lipschitz continuity factors (i.e., $\mu^{\mathsf{G}}$, $\mu^{\mathsf{H}}$, and $\mu^{\mathsf{F}}$),  estimated data variability (i.e., $\sigma^{\mathsf{H}}$ and $\sigma^{\mathsf{G}}$), and estimated data quantities (i.e., $D_j(k)$) at devices, all of which are obtained based on historical data observations.} 
 	 // \textit{Procedure at the UAV swarms}\\
 	 \For{$k' \in \{1, \dots, K_{s}^{\mathsf{G}} \}$} {
 	 // \textit{Procedure at the core network}\\
 	 \For{$\ell_u \in \mathcal{L}(s)$}{
 	 // \textit{Procedure conducted across the UAV swarms in parallel using their leader}\\
 	 \For{$k \in \{(k'-1) \tau_s^{\mathsf{G}} +1, \cdots, k' \tau_s^{\mathsf{G}} \} $}{
 	 Initialize the iteration count $m=0$.\\
 	 Choose an initial feasible point $\bm{x}{[0]}=[\boldsymbol{\rho}[0],\boldsymbol{\varrho}[0],\boldsymbol{\alpha}[0],\boldsymbol{g}[0]]$ for $\bm{\widetilde{\mathcal{P}}}_0$ at $k$.\\
 	 \While{convergence criterion is not met \textbf{OR} $m=0$}{
 	 Obtain the scalar values for $\overline{\Delta}_u(k,m)$, $\widetilde{\delta}_{j,i,n}(k;m)$, $\delta_{j,h,i,n}(k;m)$, $D_j(k,m)$, and $\lambda_{i,j}(k;m)$, $\widetilde{\lambda}_{h,i,j}(k;m)$ by substituting $\bm{x}[m]$ in~\eqref{eq:actual_delta_u},~\eqref{eq:little_delta_defs},~\eqref{eq:actualD_j}, and~\eqref{eq:little_lambda_defs}. \\
 	Use the above scalars to obtain the monomial approximations given in~\eqref{eq:delta_u_approx} and~\eqref{eq:D_j_approx}, and replace them in~$\bm{\widetilde{\mathcal{P}}}_m$.\\
 	 Apply the logarithmic change of variables and take the $\log$ from the constraints of $\bm{\widetilde{\mathcal{P}}}_m$ and convert it to a convex optimization problem (as in~\eqref{GPtoConvex} in Appendix~\ref{app:gp}).\\
 	 Solve the resulting convex optimization problem to obtain $\bm{x}[m+1]$ (e.g., using  CVX~\cite{diamond2016cvxpy}), and set  	 $m=m+1$.\label{Alg:Gpconvexste}}
 	  Choose the obtained point as  $\bm{x}^{\star}=\bm{x}[m]=[\boldsymbol{\rho}^{\star},\boldsymbol{\varrho}^{\star},\boldsymbol{\alpha}^{\star},\boldsymbol{g}^{\star}]$.} 
 	  The leader broadcasts the solution $\bm{x}^{\star}$ among its constituent workers/coordinators/devices to start their respective data transfer procedures, and specifically for workers to tune their CPU cycles and mini-batch sizes.
 	  }}}
  \end{algorithm}
  
\textbf{Developing the solver:}
First, we must convert the ratio of posynomials in the objective of $\bm{\mathcal{P}}$ into that of a posynomial and a monomial. To do so, we iteratively approximate the posynomial denominators, which are $\Delta_u$ and $D_j$ in our case, using the arithmetic-geometric mean inequality (see Lemma~\ref{Lemma:ArethmaticGeometric} in Appendix~\ref{app:gp}). The resulting approximations for iteration $m+1$ are given in~\eqref{eq:delta_u_approx} and~\eqref{eq:D_j_approx}, where $m$ is the iteration index. 
Here, the solution to the problem at the $m$-th approximation iteration is denoted $\mathbf{x}[m]= \{\bm{\varrho}[m],\bm{\rho}[m],\bm{\alpha}[m], \bm{g}[m] \}$. $\overline{\Delta}_u(k;m)$, $\delta_{j,h,i,n}(k;m)$, and $\widetilde{\delta}_{i,j,n}(k;m)$, $n \in \{1,2,3\}$ in~\eqref{eq:delta_u_approx} and $D_j(k;m)$, $\lambda_{h,i,j}(k;m)$, and $\widetilde{\lambda}_{i,j}(k;m)$ in~\eqref{eq:D_j_approx} are scalar values obtained by substituting the solution at iteration $m$ (i.e., $\mathbf{x}[m]$) in the corresponding expressions in~\eqref{eq:actual_delta_u},~\eqref{eq:little_delta_defs},~\eqref{eq:actualD_j}, and~\eqref{eq:little_lambda_defs}, respectively. 
It can be verified that the approximations in~\eqref{eq:delta_u_approx} and~\eqref{eq:D_j_approx} are in fact the best local monomial approximations to their respective posynomials near the fixed point $\mathbf{x}[m]$ in the sense of the first-order Taylor approximation. 

We solve $\bm{\mathcal{P}}$ in a distributed manner at each UAV swarm through sequentially applying the above approximations to obtain problem $\widetilde{\bm{\mathcal{P}}}_m$ at iteration $m$. In  $\widetilde{\bm{\mathcal{P}}}_m$, each constraint is an inequality on posynomials, and the objective function is a sum of posynomials, admitting the GP format:
\vspace{-1mm}
\begin{align*}\label{eq:ApproxP1}
  \hspace{-3mm} & (\bm{\widetilde{\mathcal{P}}}_m):~\hspace{-1mm} \min_{\boldsymbol{\rho},\boldsymbol{\varrho},\boldsymbol{\alpha},\boldsymbol{g}}~ (1-\theta)\widetilde{(a)}_m + \theta (b) ~~
\textrm{s.t.}~~~ \eqref{eq:f1_con1}-\eqref{eq:f1_con2},
\hspace{-3mm} 
\end{align*} 
where term $\widetilde{(a)}_m$ is obtained from term ${(a)}$ in problem $\mathcal{P}$ by: (i) decomposing ${(a)}$ into a sum of separable functions with respect to UAV swarm indexes; and then (ii) using the expressions in~\eqref{eq:delta_u_approx} and~\eqref{eq:D_j_approx} for iteration $m$ in $(a)$. Note that term $(b)$ defined in~\eqref{eq:f1_obj} is indexed by UAV swarm $u$, and the constraints of $\bm{\widetilde{\mathcal{P}}}$ are with respect to UAV swarm. 

The pseudocode of our resulting sequential approximation method is given in Algorithm~\ref{alg:cent}. The problem is solved at the beginning of each global aggregation interval $k'$ by the leader UAVs in each active swarm. The following proposition shows that the algorithm has the most desirable convergence properties for a non-convex solver:
\vspace{-1mm}
\begin{proposition}\label{prop:optOpt}
Algorithm~\ref{alg:cent} generates a sequence of improved
feasible solutions for problem $\bm{\widetilde{\mathcal{P}}}_m$ that converge to a point $\bm{x}^\star$
satisfying the Karush-Kuhn-Tucker (KKT) conditions of $\bm{\mathcal{P}}$.
\end{proposition}
\vspace{-1mm}
\begin{proof}
See Appendix~\ref{app:optOpt}.
\end{proof}

\normalsize
\vspace{-3mm}
\section{Swarm Trajectory and {\color{black}Temporal Design}}
\label{sec:swarmTraj}
\noindent 
{\color{black} 
Our developed {\tt HN-PFL} methodology (Sec.~\ref{sec:intro}) and ML performance and energy optimization (Sec.~\ref{sec:sys_model})  are the fundamental components of our UAV-assisted ML methodology.
In fact, in cases where the swarms are deployed and \textit{stationary} above the clusters (e.g.,~\cite{zeng2020federated}), these two components are enough to achieve optimal ML performance and energy efficiency.

Nevertheless, we go further by generalizing our methodology to a more realistic scenario where there are more clusters than UAV swarms, which requires UAV swarms to travel among the clusters. 
It is this generalization that motivates swarm trajectory and temporal (e.g., length of ML training sequence) design components (the green block of Fig.~\ref{fig:overall_flow}). 
Together our three components, depicted in Fig.~\ref{fig:overall_flow} and developed in Sec.~\ref{sec:HN-PFL},~\ref{sec:netwrok_pfl}, and~\ref{sec:swarmTraj}, complete our methodology.

Upon completion of a training sequence using our {\tt HN-PFL} and performance optimization components (Sec.~\ref{sec:HN-PFL} and~\ref{sec:netwrok_pfl}), {\color{black} the core network instructs the UAV swarms to travel to their next device cluster, which will result in an update to both the global and cluster specific ML models, or to a recharging station.}
{\color{black} The core network also instructs the UAV swarms to discard their gathered data, as the local training and personalization conducted by each swarm should be specific to the unique underlying data distribution of its next visited swarm.}
We next design the swarm trajectories, sequence duration and aggregation period to maximize the ML performance. }

\vspace{-2mm}
\subsection{Online Model Training under Model/Concept Drift}\label{sec:modelDrift}
Since we consider \textit{online model training}, where the data distributions at the devices are time varying, the performance of the local model changes over time. To capture this effect, we introduce the notion of \textit{model/concept drift}.
\vspace{-1mm}
\begin{definition} [Model/Concept Drift]
For device cluster $c$ with local meta-loss function $F_c$, we denote the online model drift at time $t$ by $\Lambda_c(t)\in\mathbb{R}^+$, which captures the maximum variation of the gradient for any given model parameter across two consecutive time instances; mathematically:
\vspace{-1.5mm}
\begin{equation}\label{eq:conceptDrift}
   \hspace{-.1mm} \resizebox{.92\linewidth}{!}{$
   \left\Vert \nabla F_c \left(\mathbf w\bigg|\tilde{\mathcal{D}}(t)\right)-\nabla F_c \left(\mathbf w\bigg|\tilde{\mathcal{D}}(t-1)\right) \right\Vert^2 \hspace{-1mm}\leq \Lambda_c(t), \forall \mathbf{w}.$}\hspace{-3mm}
   \vspace{-1mm}
\end{equation}
\end{definition}
Device clusters with higher values of model drift are likely to require more frequent recalibration of their local models (achieved by revisiting with UAV swarms), as their local models become obsolete faster. 
In contrast, clusters with smaller model drift, i.e., small fluctuations in local gradient, may not be worth revisiting, due to marginal rewards (in terms of model performance gains) per energy consumed. 
Also, 
if clusters experience large and consistent model/concept drift, then the learning duration $T_s$ should be smaller as the network will require swarms to re-calibrate local models more frequently. Model drift can be estimated every time a UAV swarm returns to a previously visited cluster by comparing historical meta-gradient vs. current meta-gradient computed by upon arrival.

Next, we demonstrate how model drift can be utilized to estimate local model performance given the current data distribution at a device cluster:
\vspace{-1mm}
\begin{lemma}[Estimating online gradient via model drift]\label{lemma:conceptDrift}
Let $t^{\mathsf{V}}_c$ denote a time instance when device cluster $c$ was visited by a UAV swarm,  $\mathbf w_c(t^{\mathsf{V}}_c)$ denote the corresponding local model, and $\nabla F_c (\mathbf w_c(t^{\mathsf{V}}_c) | \widetilde{\mathcal{D}}_c(t^{\mathsf{V}}_c))$ denote the local gradient. Consider time instance $t\geq t^{\mathsf{V}}_c+1$, where during the time interval $[t^{\mathsf{V}}_c+1,t]$ cluster $c$ remains unvisited by UAV swarms. 
Given the updated data at device cluster $c$, i.e., $\widetilde{\mathcal{D}}_c(t)$, the local meta gradient at the device cluster for the outdated local model is upper bounded as:
\vspace{-3mm}

{\small
\begin{align}\label{eq:resConsDrift}
\hspace{-1.5mm}\left \Vert \nabla F_c \left(\mathbf w_c(t^{\mathsf{V}}_c) \big| \widetilde{\mathcal{D}}_c(t) \right)\right\Vert^2 \hspace{-1.7mm}\leq& \underbrace{(t-t^{\mathsf{V}}_c+1) \left\Vert\nabla F_c \left(\mathbf w_c(t^{\mathsf{V}}_c) \big| \widetilde{\mathcal{D}}_c(t^{\mathsf{V}}_c)\right)\right\Vert^2}_{(a)} \nonumber\\[-0.3em]\hspace{-3mm}
&+\underbrace{(t-t^{\mathsf{V}}_c+1)\sum_{t'=t^{\mathsf{V}}_c+1}^{t} \Lambda_c(t')}_{(b)},
\end{align}
}
\vspace{-3mm}

\noindent where $\Lambda_c(\cdot)$ is defined in~\eqref{eq:conceptDrift}. Subsequently, assuming an upper bound on the model drift at the device cluster $\Lambda_c(t')\leq \Lambda_c$, $t'\in[t^{\mathsf{V}}_c+1,t]$, we have: {\small
$ \left \Vert \nabla F_c \left(\mathbf w_c(t^{\mathsf{V}}_c) \big| \widetilde{\mathcal{D}}_c(t) \right)\right\Vert^2 \leq(t-t^{\mathsf{V}}_c+1) \left\Vert\nabla F_c \left(\mathbf w_c(t^{\mathsf{V}}_c) \big| \widetilde{\mathcal{D}}_c(t^{\mathsf{V}}_c)\right)\right\Vert^2 + (t-t^{\mathsf{V}}_c+1)(t-t^{\mathsf{V}}_c)\Lambda_c$}.
\end{lemma}
\vspace{-1mm}
\begin{proof}
See Appendix~\ref{app:lemConc}.
\end{proof}

The bound in~\eqref{eq:resConsDrift} demonstrates that the real-time performance of the outdated model, measured through the value of the meta gradient, becomes obsolete linearly with time (term (a)) and cumulative value of the model drift at the device cluster (term (b)). We will incorporate this result in the design of UAV swarm trajectories. 
\vspace{-3mm}
\subsection{Problem Formulation and DRL Characteristics} \label{ss:DRL_formulation}
\subsubsection{Background} 
In the following, we first formulate the problem of UAV swarm trajectory design and learning sequence duration. 
We then cast the problem as a sequential decision making problem. Subsequently, we develop a deep reinforcement learning (DRL) methodology by encoding the real-time characteristics of our network into \textit{states} for the DRL, defining a calculation methodology for the \textit{actions} available for the DRL, and linking the actions to DRL \textit{reward} calculations by embedding the optimization methodology from Section~\ref{sec:netwrok_pfl}.

For training sequence $s$, let $X_u(s)$ denote the location of UAV swarm $u$, $\Lambda_c(s)$ denote the latest estimation of the model drift of device cluster $c$, $E_u(s)$ denote the minimum remaining battery among the UAVs belong to swarm $u$, and $G_c(s)$ denote the latest value of the gradient at device cluster $c$. 
Since the duration of ML model training $T_s$, $\forall s$, is usually far smaller than the delay between consecutive visits to the same device cluster, we assume that the effect of model drift during model training is negligible, as we do so in Sec.~\ref{sec:HN-PFL}, and instead integrate the effects of model drift in our DRL design caused by delay between consecutive visits of the device cluster. 
For device cluster $c$, if it has been visited during sequence $s$, i.e., $c\in \mathcal{C}(s)$, $G_c(s)$ captures its recent gradient: {\small
$ G_c(s)=\left \Vert \nabla F_c \left(\mathbf w_c(t_s+T_s) \big| \widetilde{\mathcal{D}}_c(t_s+T_s) \right)\right\Vert^2$} measured at the end of the training sequence; however, if it is not visited, i.e., $c\in \overline{\mathcal{C}}(s)$, then {\small$ G_c(s)=\left \Vert \nabla F_c \left(\mathbf w_c(t^{\mathsf{V}}_c) \big| \widetilde{\mathcal{D}}_c(t) \right)\right\Vert^2$}, which is computed via the bound in~\eqref{eq:resConsDrift}. 
Also, we let $E^{\mathsf{M}}(s)$ denote the sum of energy of movement of the UAV swarms to travel to their current location in training sequence $s$ from their previous locations in sequence $s-1$, and let $O(s)$ denote the value of the final objective function of $\bm{\mathcal{P}}$ solved in Section~\ref{sec:netwrok_pfl} encompassing both the energy used for model training and the gradient of those devices under model training.

\subsubsection{Problem Formulation}
Given a total of $S$ ML model training sequences, we propose the following formulation to determine the optimal trajectory design and temporal ML characteristics:
\vspace{-2mm}
\begin{align}
   &\hspace{-3mm}(\bm{\widehat{\mathcal{P}}}):\hspace{-2mm} \min_{\substack{\{\mathcal{X}(s)\}_{s=1}^{S}\\ \{
   \mathcal{T}(s)\}_{s=1}^{S}}} \hspace{-1mm} \frac{1}{S}\sum_{s=1}^{S} \left[c_1 E^{\mathsf{M}}(s) +  c_2 O(s) + c_3 \hspace{-2mm}\sum_{c\in\overline{C}(s)}\hspace{-2mm} G_c(s)  \right]\hspace{-3mm} \label{eq:drl_min_obj} \\
   & \textrm{s.t.} \nonumber \\
   & E_u(s) \geq E^{\mathsf{T}},  u \in \mathcal{U}(s), s \in \{1, \cdots,S \} \label{eq:f2_con1}\\
   &   \mathcal{T}(s) \in \mathcal{T}^{\mathsf{F}},  s \in \{1, \cdots,S\} \label{eq:f2_con2},
\end{align}
\vspace{-5mm}

\noindent where {\small $\mathcal{X}(s)\triangleq \{X_1(s),\cdots,X_U(s)\}$}, and  {\small $\mathcal{T}(s) \triangleq \left\{ T_{s} \cup \tau^{\mathsf{L}}_{s} \cup \tau^{\mathsf{G}}_{s}\right\}$}. 
The objective function in~\eqref{eq:drl_min_obj} consists of three parts, the energy required for  swarms to move to their next destination ({\small$E^{\mathsf{M}}(s)$}), the data offloading/processing objective function result ({\small$O(s)$} derived from  {\small$\bm{{\mathcal{P}}}$}), and the estimated online gradient as a result of model drift at clusters without UAV training ({\small$G_c(s), c\in\overline{C}(s)$}). 
Constraint~\eqref{eq:f2_con1} ensures that swarms have sufficient energy  {\small$E^{\mathsf{T}}$}, at all times, to travel to a recharging station before they can no longer sustain their flight, and constraint~\eqref{eq:f2_con2} guarantees that the temporal ML characteristics  {\small$\mathcal{T}(s) \triangleq \left\{ T_{s} \cup \tau^{\mathsf{L}}_{s} \cup \tau^{\mathsf{G}}_{s}\right\}$}, $\forall s$, are always within some feasible set {\small$\mathcal{T}^{\mathsf{F}}$}. 
Finally, $c_1$, $c_2$, and $c_3$ in~\eqref{eq:drl_min_obj} scale the objective components. 


Solving $\bm{\widehat{\mathcal{P}}}$ faces the following challenges: (i) the effects of training at a previous device cluster carries over to all future training sequences, at any device cluster, (ii) $\bm{\widehat{\mathcal{P}}}$ is a combinatorial problem, which suffers from the \textit{curse of dimensionality}, since at each sequence the core network must assign the swarms to either device clusters or recharging stations, and (iii) $\bm{\widehat{\mathcal{P}}}$ is defined over an unknown environment, i.e., neither the data distributions at the device clusters nor the model/concept drift are known apriori. 
These facts motivate us to cast the problem as a \textit{sequential decision making problem}, which is solved at the core network for each training sequence by encoding the network aspects as reinforcement learning objectives.
Classical reinforcement learning methods rely on a pre-built Q-table to determine future actions and associated network states~\cite{mnih2015human}, but, due to the aforementioned challenges of our problem, in particular the curse of dimensionality, building a Q-table is infeasible. {\color{black}Motivated by the success that recurrent neural networks (RNN) exhibited as the deep Q-network for sequential decision making~\cite{9448143}, we adapt an RNN based DRL featuring LSTM layers.} 

\subsubsection{State of the DRL} 
We encode the locations of the UAV swarms, the model drift and gradients at device clusters, and the temporal ML characteristics (i.e., total time, local-global aggregation periods) as the state of the DRL. 
Formally, we define the state at the end of training sequence $s$, $\mathcal{Z}(s)$, as:
\vspace{-1.5mm}
\begin{equation}
    \mathcal{Z}(s)=\mathcal{U}(s) \cup \mathcal{X}(s) \cup \Lambda(s) \cup \mathcal{E}(s) \cup \mathcal{G}(s) \cup \mathcal{T}(s),
    \vspace{-1.5mm}
\end{equation}
where {\small$\mathcal{U}(s)$} denotes the set of active UAV swarms, $\mathcal{X}(s)$ denotes the locations of all the swarms,  {\small$\Lambda(s)=\{\Lambda_1(s),\cdots,\Lambda_C(s)\}$} are the latest model drift estimates observed at the device clusters, {\small$\mathcal{E}(s)=\{{E}_1(s),\cdots, {E}_U(s)\}$}, {\small$\mathcal{G}(s)=\{{G}_1(s),\cdots,{G}_C(s)\}$}, and {\small$\mathcal{T}(s)=\left\{ T_{s} \cup \tau^{\mathsf{L}}_{s} \cup \tau^{\mathsf{G}}_{s}\right\}$}. 

\subsubsection{Action of the DRL} 
The core network, as the DRL agent, determines the active UAV swarms {\small$\mathcal{U}(s+1)$}, their locations {\small$\mathcal{X}(s+1)$}, and the temporal behavior {\small$\mathcal{T}(s+1)$} of the next training sequence. In particular, we define the DRL action, $\mathcal{H}(s)$, as:
\vspace{-1.5mm}
\begin{equation}
    \mathcal{H}(s)= \mathcal{X}(s+1) \cup \mathcal{U}(s+1) \cup \mathcal{T}(s+1).
    \vspace{-1mm}
\end{equation}
\vspace{-5mm}
\subsubsection{Reward of the DRL} 
The DRL agent aims to maximize the reward of each action $\mathcal{H}(s)$ with respect to the current state $\mathcal{Z}(s)$ via the objective value of $\widehat{\bm{\mathcal{P}}}$. 
Formally, we define the reward of the agent as:
\vspace{-2mm}
\begin{equation}\label{eq:rewardDRL}
\hspace{-.1mm}
\resizebox{.93\linewidth}{!}{$
    V(s)=  C\underbrace{\left({c_1{{O(s)}}+ c_2 \sum_{c\in \overline{\mathcal{C}}(s)} {{{G}_c}(s)} +c_3 E^{\mathsf{M}}(s)}\right)^{-1}}_{(a)} - \underbrace{P \sum_{u\in \mathcal{U}} \mathbbm{1}_{\{E_u\leq E^{\mathsf{T}}\}}}_{(b)},$}\hspace{-4mm}
    \vspace{-2mm}
\end{equation}
where $(a)$ captures the (inverse) value of the objective function of~$\bm{\widehat{\mathcal{P}}}$ for a particular training sequence $s$, and $(b)$ is a penalty function with $P \gg 1$, capturing the case where UAVs' battery level drops below $E^{\mathsf{T}}$, upon which the indicator $\mathbbm{1}_{\{E_u\leq E^{\mathsf{T}}\}}$ takes the value of one for UAV swarm $u$. 

\begin{figure}[t]
    \centering
    \includegraphics[width = 0.48\textwidth]{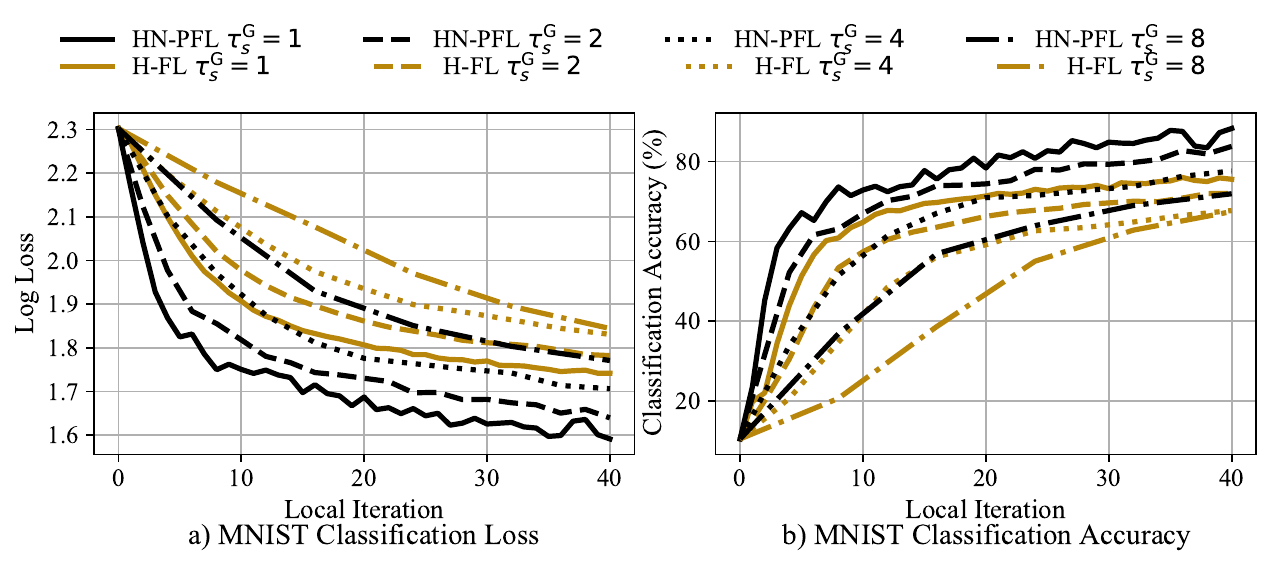}
        \vspace{-2.5mm}
    \caption{ML architecture comparisons for MNIST with fixed $\tau_s^{\mathsf{L}} = 1$. {\tt HN-PFL} demonstrates faster convergence and better final classification accuracy than {\tt H-FL} for various $\tau_s^{\mathsf{G}}$.}
    \label{fig:mnist_fixed_swarm_aggs}
    \vspace{-5mm}
\end{figure}

\textbf{Main Takeaways.} Our reward function in~\eqref{eq:rewardDRL} captures multiple possibly competing objectives: (i) it motivates visiting device clusters that benefit from network-aware ML model training, i.e., those that require less energy to achieve a better model, (ii) it avoids leaving clusters with larger model drifts (via $G_c(s)$) unvisited for long periods of time, (iii) it promotes {\color{black} visiting clusters with high performance gain (via $O(s)$) as compared to ML energy consumption}, and (iv) it avoids situations where {\color{black} swarm battery levels would drop below a threshold (via term $(b)$).}

\subsubsection{DRL Learning Architecture} 
We exploit an {\color{black} RNN with LSTM layers} to approximate the optimal action-value function based on the Bellman equation:
\vspace{-2mm}
\begin{equation} \label{eq:optim_q_fxn}
\begin{aligned}
       & Q^{\star}(\mathcal{Z}(s),\mathcal{H}(s)) = \mathbb{E}\bigg[ V(s)\\
       &  + \gamma \displaystyle\max_{\mathcal{H}(s+1)}Q^{\star}\left(\mathcal{Z}(s+1),\mathcal{H}(s+1)\right) \big| \mathcal{Z}(s),\mathcal{H}(s)\Big] \bigg],
\end{aligned}
\vspace{-1mm}
\end{equation}
where 
$\gamma$ is the future reward discount. We use one {\color{black} RNN} to approximate $Q^{\star}$ called the \textit{train Q-network} $Q_{\boldsymbol{\theta}}(\mathcal{Z}(s),\mathcal{H}(s))$, which we train by adjusting its parameters $\boldsymbol{\theta}$ to reduce the mean-squared error (MSE). 
Classical deep Q-network techniques determine the MSE of the train Q-network with respect to a target reward $y(s)$ that also depends on the train Q-network's model parameters $\boldsymbol{\theta}$. This self-coupling produces over-estimation, so we use an additional {\color{black}RNN} called \textit{target Q-network} $\widehat{Q}_{\widehat{\boldsymbol{\theta}}}$ with parameters $\widehat{\boldsymbol{\theta}}$ to produce accurate loss measurements and periodically synchronize the target Q-network to the train Q-network. 
Furthermore, to prevent correlations in the environment observation sequence from influencing the parameters $\boldsymbol{\theta}$ in the \textit{train Q-network}, we use \textit{experience replay} to sample a randomized mini-batch of $\mathcal{M}^{\mathsf{Q}}$ experience tuples, each of which is of the form $(\mathcal{Z}(s), \mathcal{H}(s), V(s), \mathcal{Z}(s+1))$, to calculate MSE. 
In particular, we compute the mean-squared error as:
\vspace{-2mm}
\begin{equation}
\begin{aligned}
    L(\boldsymbol{\theta}) = \frac{1}{|\mathcal{M}^{\mathsf{Q}}|} \sum_{s \in \mathcal{M}^{\mathsf{Q}}} \big(Q_{\boldsymbol{\theta}}(\mathcal{Z}(s),\mathcal{H}(s)) - y(s)\big)^2,
\end{aligned}
\vspace{-2mm}
\end{equation}
where {\small $y(s) = V(s) + \gamma \max_{\mathcal{H}(s+1)} \widehat{Q}_{\widehat{\boldsymbol{\theta}}}(\mathcal{Z}(s+1),\mathcal{H}(s+1))$} is the target reward. 
Using the MSE, we then perform SGD to adjust the parameters $\boldsymbol{\theta}$. 
Since this procedure requires at least $M^{\mathsf{Q}}$ previous experience tuples (saved in a deque-style buffer of size $B^{\mathsf{Q}}$), and each training sequence $s$ only generates a single tuple, the {\color{black}RNN} training requires at least $M^{\mathsf{Q}}$ training sequences before it begins. 
In order to generate representative experience tuples from the environment before the {\color{black}RNN} is properly trained and to ensure that the {\color{black}DRL process is better able to find the global maximum (rather than getting stuck in a local maxima)}, we utilize an \textit{$\epsilon$-greedy policy}~\cite{stadie2015incentivizing}, wherein with probability $\epsilon$ the agent will select an action randomly and with probability $(1-\epsilon)$ the agent will determine the action based on the train Q-network $Q_{\boldsymbol{\theta}}$. As the train Q-network improves over time, $\epsilon$ decreases with limit $\epsilon^{\mathsf{min}}\leq \epsilon$. 

\begin{figure}[t]
    \centering
    \includegraphics[width = 0.48\textwidth]{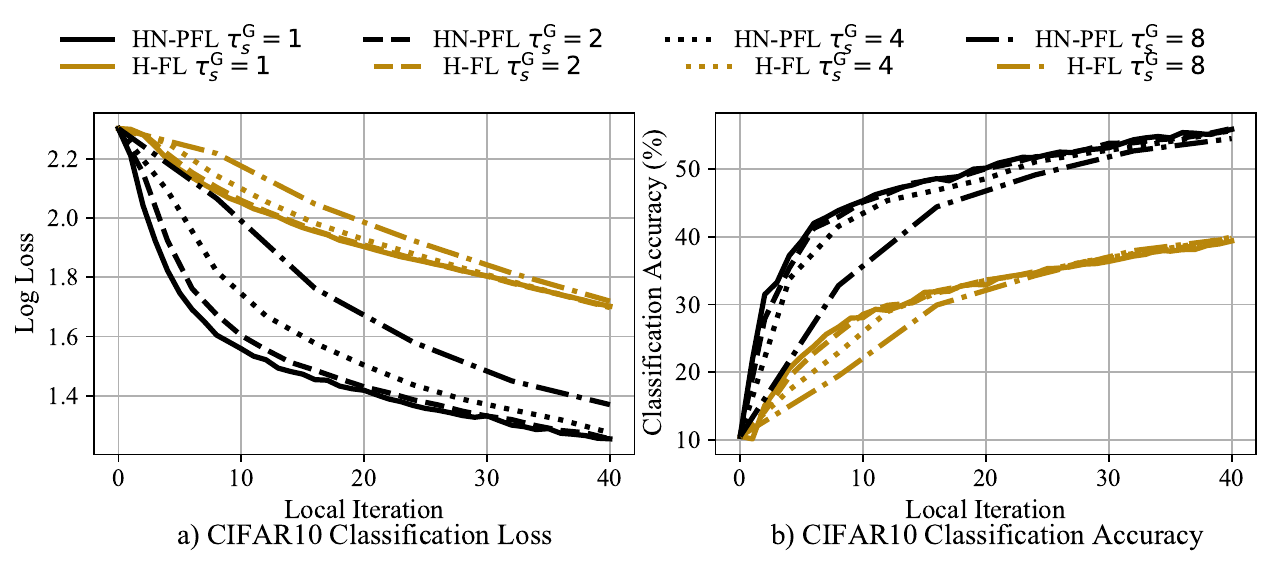}
    \vspace{-2.5mm}
    \caption{{\color{black}ML architecture comparisons for CIFAR-10 with fixed $\tau_s^{\mathsf{L}} = 1$. {\tt HN-PFL} demonstrates faster convergence and better final classification accuracy than {\tt H-FL} for various $\tau_s^{\mathsf{G}}$.}}
    \label{fig:cifar10_fixed_swarm_aggs}
    \vspace{-5mm}
\end{figure}

\vspace{-4mm}
\subsection{System Integration: A Solution for UAV-enabled Online ML Over Heterogeneous Networks}
\label{ssec:system_integration}

In the past few sections, we developed our methodology for network-aware UAV-enabled online model training, which comprises {\tt HN-PFL} from Sec.~\ref{sec:HN-PFL}, the data processing/offloading optimization from Sec.~\ref{sec:netwrok_pfl}, and the swarm trajectory and ML design from Sec.~\ref{sec:swarmTraj}. 
The interconnection between the different components is visualized in Fig.~\ref{fig:overall_flow}, which we provided a high level discussion on in Sec.~\ref{ssec:full_architecture}. Given our introduced variables and parameters in Sec.~\ref{sec:HN-PFL},~\ref{sec:netwrok_pfl}, and~\ref{sec:swarmTraj}, we can now further comment on the interdependence.

Specifically, the {\tt HN-PFL} methodology (orange block) is implemented by the UAV swarms, and operates based on the reception of transfer/processing parameters (i.e., $\boldsymbol{\rho}, \boldsymbol{\varrho}, \boldsymbol{\alpha}, \mathbf{f}$) from the data processing optimization at the swarm leaders (blue block). 
{\color{black} The ML training results and UAV battery statuses are embedded into network states and form the bases of our
swarm trajectory (i.e., $\mathcal{X}(s)$) and temporal training decisions (i.e., $\mathcal{T}(s), \tau_s^{\mathsf{L}}, \tau_s^{\mathsf{G}}$) for the training sequences (green block), which is then shared from the core network to the swarm leaders. With these core network decisions, the next sequence of ML training will begin, and the cycles continues.}

\vspace{-2mm}
\section{Numerical Evaluation and Discussion}
\label{sec:simulations}
\noindent 
In this section, we conduct numerical evaluations of our proposed methodology.
When literature contains existing techniques, such as hierarchical federated learning ({\tt H-FL})~\cite{liu2020client}, we compare our methodology against it; otherwise, we develop heuristic algorithms as baselines. 
{\color{black} For brevity, we present results here based on two datasets, MNIST (digits) and CIFAR-10 (color objects). Results for two other datasets, FMNIST (grayscale objects) and RADIOML (wireless signals) are deferred to Appendix A. Descriptions of each dataset and UAV network parameters can also be found in Appendix A. The key findings for each dataset are qualitatively consistent.} 

\begin{figure}[t]
    \centering
    \includegraphics[width = 0.48\textwidth]{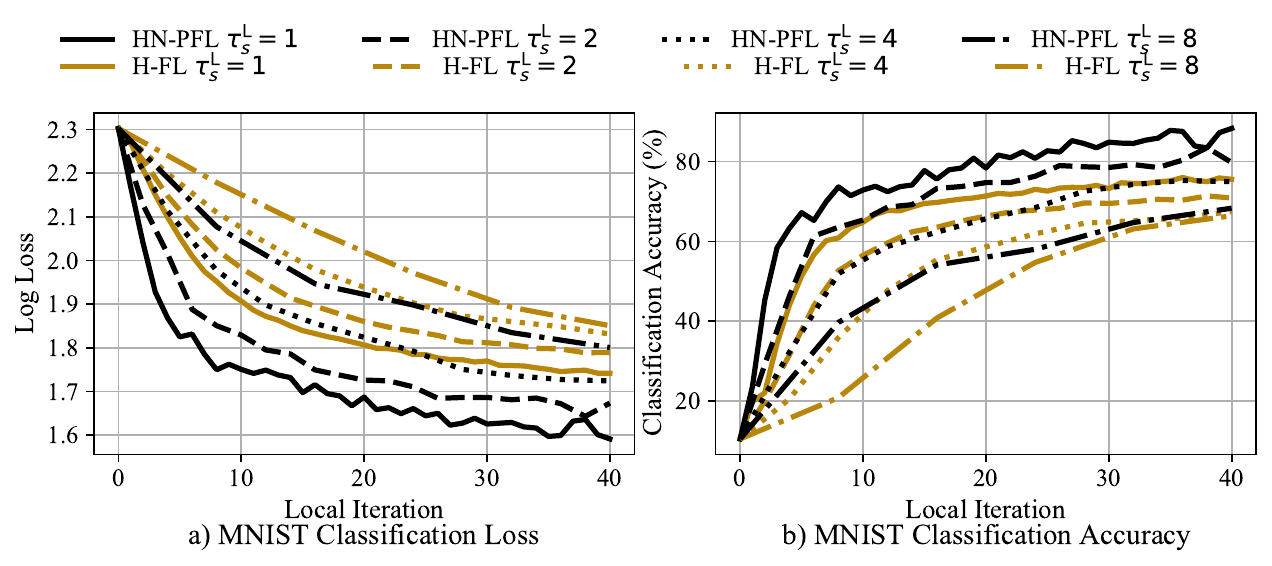}
    \vspace{-2.5mm}
    \caption{Using the same network for MNIST as Fig.~\ref{fig:mnist_fixed_swarm_aggs}, this experiment fixes $\tau_s^{\mathsf{G}} = 1$ to vary $\tau_s^{\mathsf{L}}$ instead. Our methodology {\tt HN-PFL} continues to outperform {\tt H-FL} consistently.}
    \label{fig:mnist_fixed_global_aggs}
    \vspace{-0mm}
\end{figure}

\vspace{-4mm}
\subsection{{\tt HN-PFL} Proof-of-Concept} \label{ss:proof_of_concept}
\vspace{-.5mm}
We start by investigating the {\tt HN-PFL} procedure from Sec.~\ref{sec:HN-PFL}, by comparing its performance (measured via classification accuracy, loss, and energy consumption) to that of hierarchical federated learning ({\tt H-FL}) proposed in~\cite{liu2020client}. 
For this simulation, we consider a network composed of 4 UAV swarms with $2$-$3$ workers, where each swarm has data from only $3$ labels (thus, non-iid data distributions) and data quantity determined randomly from a Gaussian distribution: $\mathcal{N}(2500,250)$ for MNIST and {\color{black}$\mathcal{N}(3500,350)$ for CIFAR-10}. 
As a proof-of-concept, we strictly isolate the performance of the distributed ML methodologies to the UAV layers, i.e., we assume the data has already been transferred to the worker UAVs. 
For a fair comparison between {\tt HN-PFL} and {\tt H-FL}, we ensure that both methodologies train over the same amount of data for each of their iterations by defining the batch size for {\tt H-FL} as $\alpha^{\mathsf{HFL}}$ and setting each batch ratio for {\tt HN-PFL} as $\alpha_{j,i} \triangleq \floor{\frac{1}{3} \alpha^{\mathsf{HFL}}}~\forall i \in \{1,2,3 \}$. 
Finally, we use the following settings for ML model training: $\eta = \eta_1 =10^{-3}$ and $\eta_2 = 10^{-2}$, where $\eta$ is the learning rate for gradient descent in {\tt H-FL}, and to avoid lengthy Hessian computations, we use the Hessian first-order approximation (see Remark~\ref{rem:1}).

\begin{table}[!t]
\caption{Energy consumption for MNIST to reach $65\%$ accuracy}
\vspace{-1.5mm}
\label{tab:personalized_mnist} 
\begin{tabularx}{0.48\textwidth}{
p{\dimexpr.05\linewidth-2\tabcolsep-1.3333\arrayrulewidth}
p{\dimexpr.05\linewidth-2\tabcolsep-1.3333\arrayrulewidth}
p{\dimexpr.1\linewidth-2\tabcolsep-1.3333\arrayrulewidth}
p{\dimexpr.12\linewidth-2\tabcolsep-1.3333\arrayrulewidth}
p{\dimexpr.15\linewidth-2\tabcolsep-1.3333\arrayrulewidth}
| 
p{\dimexpr.05\linewidth-2\tabcolsep-1.3333\arrayrulewidth}
p{\dimexpr.05\linewidth-2\tabcolsep-1.3333\arrayrulewidth}
p{\dimexpr.1\linewidth-2\tabcolsep-1.3333\arrayrulewidth}
p{\dimexpr.12\linewidth-2\tabcolsep-1.3333\arrayrulewidth}
p{\dimexpr.15\linewidth-2\tabcolsep-1.3333\arrayrulewidth}
}
\toprule[.2em]
\multicolumn{2}{c}{\bf{Ratio}} & \multicolumn{3}{c}{\bf{MNIST} (kJ)} & \multicolumn{2}{c}{\bf{Ratio}} & \multicolumn{3}{c}{\bf{MNIST} (kJ)} \\
\cmidrule(lr){1-2} \cmidrule(lr){3-5} \cmidrule(lr){6-7} \cmidrule(lr){8-10} 
$\centering \tau_s^{\mathsf{L}}$ & $\centering \tau_s^{\mathsf{G}}$ & \bf{{\scriptsize HFL}} & \centering \bf{{\scriptsize HNPFL}} & \bf{{\scriptsize Savings}} & $\centering \tau_s^{\mathsf{L}}$ & $\centering \tau_s^{\mathsf{G}}$ & \bf{{\scriptsize HFL}} & \centering \bf{{\scriptsize HNPFL}} & \bf{{\scriptsize Savings}}\\
\midrule
\centering 1 & \centering 1 & 4.64 & 2.32 & 50.0\% 
& 1 & 1 & 4.64 & 2.32 & 50.0\% \\ 
\centering 1 & \centering 2 & 8.13 & 4.06 & 50.1\% 
& 2 & 1 & 9.29 & 4.64 & 50.1\% \\ 
\centering 1 & \centering 4 & 14.51 & 8.13 & 44.0\% 
& 4 & 1 & 15.10 & 8.13 & 46.2\% \\ 
\centering 1 & \centering 8 & 19.73 & 17.99 & 8.8 \% 
& 8 & 1 & 18.57 & 14.51 & 21.9\% \\ 
\bottomrule
\end{tabularx}
\vspace{-6mm}
\end{table}

\begin{figure}[t]
    \centering
    \includegraphics[width = 0.48\textwidth]{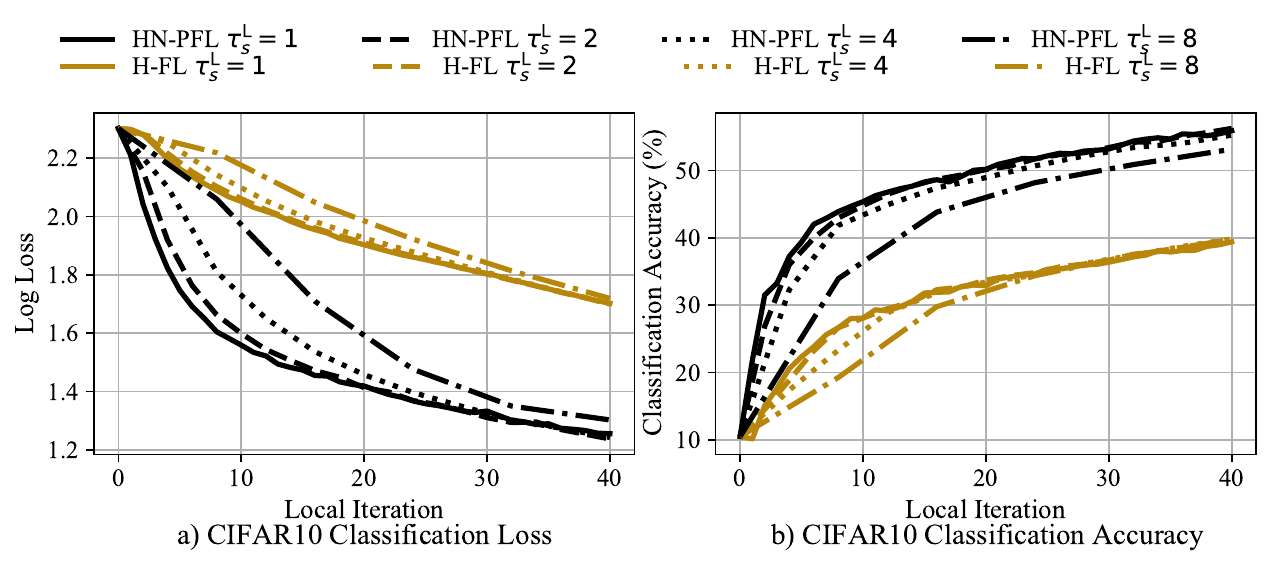}
    \vspace{-2.5mm}
    \caption{{\color{black}Using the same network for CIFAR10 as Fig.~\ref{fig:cifar10_fixed_swarm_aggs}, this experiment fixes $\tau_s^{\mathsf{G}} = 1$ to vary $\tau_s^{\mathsf{L}}$ instead. Our methodology {\tt HN-PFL} continues to outperform {\tt H-FL} consistently.}}
    \label{fig:cifar10_fixed_global_aggs}
    \vspace{-0mm}
\end{figure}

We investigate the impact of local and global aggregations on the performance and efficiency of the ML training separately. 
We first investigate the effects of varying global aggregation period $\tau_s^{\mathsf{G}}$ and fix local aggregation period (i.e., $\tau_s^{\mathsf{L}}=1$) in Fig.~\ref{fig:mnist_fixed_swarm_aggs} for MNIST. 
We repeat this experiment with fixed $\tau_s^{\mathsf{G}} = 1$ and vary $\tau_s^{\mathsf{L}}$ in Fig.~\ref{fig:mnist_fixed_global_aggs}, also for MNIST. 
{\color{black}The corresponding CIFAR-10 experiments are in Fig.~\ref{fig:cifar10_fixed_swarm_aggs} and~\ref{fig:cifar10_fixed_global_aggs}. }
Due to the non-iid and non-convex natures of our problem, the noisy convergence seen in Figs.~\ref{fig:mnist_fixed_swarm_aggs}-\ref{fig:cifar10_fixed_global_aggs} is expected. 

{\color{black}On both MNIST and CIFAR-10, we see that {\tt HN-PFL} attains better final model accuracy and classification loss over its {\tt H-FL} counterpart. For example, when $\tau_s^{\mathsf{L}} = \tau_s^{\mathsf{G}} = 1$, {\tt HN-PFL} outperforms {\tt H-FL} in trained classification accuracy by at least $10\%$ on both MNIST and CIFAR-10.} 
When we increase the aggregation period, the model performances obtained at any given local iteration are lower because the total iterations are fixed and thus longer aggregation periods result in fewer aggregations within the same timeframe. Fewer aggregations leads to worse performance as a result. 
Nonetheless, {\tt HN-PFL} is able to maintain its advantage as $\tau_s^{\mathsf{G}}$ or $\tau_s^{\mathsf{L}}$ increases for both MNIST and CIFAR-10. 
{\color{black}Furthermore, the ability of {\tt HN-PFL} to attain higher accuracy values with fewer training iterations than {\tt H-FL} results in less network energy consumption in the form of flight, processing, and communication energy among UAVs in order to reach desired accuracy thresholds.}
We demonstrate the corresponding energy savings of {\tt HN-PFL} in Table~\ref{tab:personalized_mnist} for MNIST and Table~\ref{tab:personalized_cifar10} for CIFAR-10 to reach a specific training accuracy. For each table, we selected model accuracies that are reachable for all combinations of $\tau_s^{\mathsf{L}}$ and $\tau_s^{\mathsf{G}}$. We chose $65\%$ for MNIST and {\color{black}$25\%$ for CIFAR-10}. 
{\color{black}On average, our {\tt HN-PFL} method saves $38.7\%$ and $40.6\%$ of the energy used for {\tt H-FL} for MNIST and CIFAR-10, respectively.} 


\vspace{-3mm}
\subsection{Data Offloading/Processing Optimization}
\label{ss:optim_results}
\begin{figure*}[t] 
    \centering
    \includegraphics[width = 0.95 \textwidth]{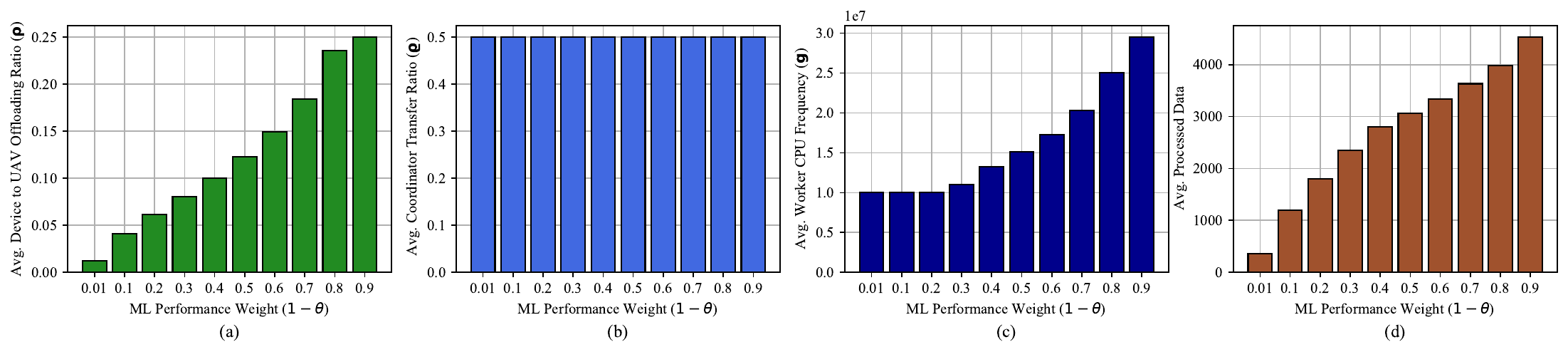}
    \vspace{-1mm}
    \caption{The behavior of the optimization variables: $\boldsymbol{\rho}, \boldsymbol{\varrho}, \boldsymbol{\alpha}, \boldsymbol{g}$ depends on the ML learning importance factor $1-\theta$ in the optimization formulation. As $1-\theta$ increases, i.e., the importance of ML model performance increases, the network responds by offloading more data on average, increasing the CPU frequencies of workers, and processing more data on average as a result.}
    \label{fig:optimizer_performance}
    \vspace{-3mm}
\end{figure*}

\begin{figure}[t] 
    \centering
    \vspace{-4mm}
    \includegraphics[width=0.45\textwidth]{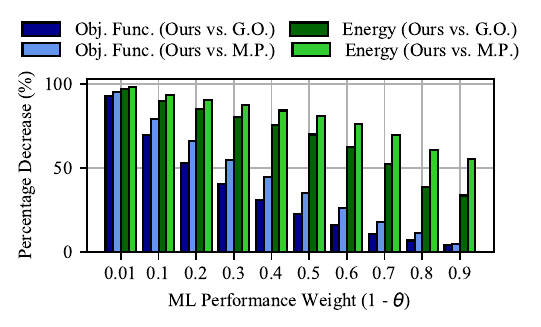}
    \caption{Our data offloading/processing optimization minimizes the network objective better than the two baselines, greedy offloading (G.O.) and maximum processed (M.P.). For comparison, we show the percentage decrease of energy consumption and objective function that our optimization achieves when compared to the baselines.} 
    \label{fig:greed_rho}
    \vspace{-3mm}
\end{figure}

{\color{black}Next, we focus on the optimization (i.e., blue) block from Fig.~\ref{fig:overall_flow}, which relates to the network control aspect of our problem.}
Note that our optimization formulation from Sec.~\ref{sec:netwrok_pfl} adjusts the performance and energy consumption of {\tt HN-PFL} and is separable with respect to each swarm.
We investigate the performance of our optimization solver on a single swarm composed of two workers and two coordinators training for a device cluster composed of ten devices in Figs.~\ref{fig:optimizer_performance}-\ref{fig:greed_rho}, averaging over 10 experiments. 
{\color{black}In practice, a network operator can vary $1 -\theta$, which scales the importance of the ML model performance vs. energy consumption in the objective of $\bm{\mathcal{P}}$. 
In Fig.~\ref{fig:optimizer_performance}, our optimization responds by adjusting {\tt HN-PFL}'s parameters: 
device-to-UAV data offloading ($\boldsymbol{\rho}$), coordinator-to-worker data offloading ($\boldsymbol{\varrho}$), aggregate data processing (i.e.,  $D_j\times \alpha_j$), and worker CPU frequency determination ($\boldsymbol{g}$).}
As the network operator places greater importance on ML model performance (i.e., increasing $1-\theta$), our solver increases data offloading, i.e., larger average $\bm{\rho}$ in Fig.~\ref{fig:optimizer_performance}(a), and swarm leaders also instruct their workers to increase their CPU frequencies in Fig.~\ref{fig:optimizer_performance}(c). {\color{black} In this manner, more data is offloaded and subsequently processed within the same time frame, which increases the ML model performance.} 
Fig.~\ref{fig:optimizer_performance}(b) also demonstrates that the coordinator UAVs never retain data for themselves as the average $\bm{\varrho}$ is $0.5$, which implies that all the data is getting offloaded to the 2 workers. This is the case since the coordinator UAVs are only used for data relaying. 
The joint effect of larger average $\boldsymbol{\rho}$ and $\boldsymbol{g}$ is more average total data processed, seen in Fig.~\ref{fig:optimizer_performance}(d). 

\begin{table}[!t]
\caption{{\color{black}Energy consumption for CIFAR-10 to reach $25\%$ accuracy.}}
\vspace{-1.5mm}
\label{tab:personalized_cifar10} 
\begin{tabularx}{0.48\textwidth}{
p{\dimexpr.05\linewidth-2\tabcolsep-1.3333\arrayrulewidth}
p{\dimexpr.05\linewidth-2\tabcolsep-1.3333\arrayrulewidth}
p{\dimexpr.1\linewidth-2\tabcolsep-1.3333\arrayrulewidth}
p{\dimexpr.12\linewidth-2\tabcolsep-1.3333\arrayrulewidth}
p{\dimexpr.15\linewidth-2\tabcolsep-1.3333\arrayrulewidth}
| 
p{\dimexpr.05\linewidth-2\tabcolsep-1.3333\arrayrulewidth}
p{\dimexpr.05\linewidth-2\tabcolsep-1.3333\arrayrulewidth}
p{\dimexpr.1\linewidth-2\tabcolsep-1.3333\arrayrulewidth}
p{\dimexpr.12\linewidth-2\tabcolsep-1.3333\arrayrulewidth}
p{\dimexpr.15\linewidth-2\tabcolsep-1.3333\arrayrulewidth}
}
\toprule[.2em]
\multicolumn{2}{c}{\bf{Ratio}} & \multicolumn{3}{c}{\bf{CIFAR-10 (kJ)}} & \multicolumn{2}{c}{\bf{Ratio}} & \multicolumn{3}{c}{\bf{CIFAR-10 (kJ)}} \\
\cmidrule(lr){1-2} \cmidrule(lr){3-5} \cmidrule(lr){6-7} \cmidrule(lr){8-10} 
$\tau_s^{\mathsf{L}}$ & $\tau_s^{\mathsf{G}}$ & \bf{{\scriptsize HFL}} & \bf{{\scriptsize HNPFL}} & \bf{{\scriptsize Savings}} 
& $\tau_s^{\mathsf{L}}$ & $\tau_s^{\mathsf{G}}$ & \bf{{\scriptsize HFL}} & \bf{{\scriptsize HNPFL}} & \bf{{\scriptsize Savings}} \\
\midrule
1 & 1 & 3.48 & 1.74 & 50.0\% 
& 1 & 1 & 3.48 & 1.74 & 50.0\% \\ 
1 & 2 & 3.48 & 2.32 & 33.3\% 
& 2 & 1 & 3.48 & 2.32 & 33.3\% \\
1 & 4 & 5.22 & 2.32 & 55.6\% 
& 4 & 1 & 5.22 & 2.32 & 55.6\% \\ 
1 & 8 & 8.13 & 2.90 & 64.3\% 
& 8 & 1 & 8.13 & 2.90 & 64.3\% \\ 
\bottomrule
\end{tabularx}
\vspace{-6mm}
\end{table}

{\color{black} Next, in Fig.~\ref{fig:greed_rho}, we verify that our optimization formulation optimizes the performance of {\tt HN-PFL}.} As no alternative methodology for our optimization problem exists, we develop two methods, called greedy offloading (G.O.) and maximum processed (M.P.), and use them as baselines to compare the effectiveness of our data offloading/processing optimization in Fig.~\ref{fig:greed_rho}. In G.O., the devices always offload their entire datasets to the UAVs, and, in M.P., the UAVs overclock their CPU frequencies (reaching 2.3GHz for all UAVs) and maximize mini-batch ratios. Both G.O. and M.P. are determined in a fashion that adheres to the constraints in~\eqref{eq:f1_con1}-\eqref{eq:f1_con2}, and \textit{we use our solver to determine the rest of the optimization variables in each baseline}. 
We show the percentage savings {\color{black} in objective function of $\bm{\mathcal{P}}$} of our method over the two baselines in Fig.~\ref{fig:greed_rho}. When the network operator places greater emphasis on energy efficiency (i.e., small $1-\theta$), our joint optimization will decrease data offloading and processing to conserve energy. As a result, our method achieves over $80\%$ decrease in objective function value and energy consumption compared to either baseline when $1-\theta = 0.01$. 
Even when $1-\theta =0.9$, i.e., the network aims to process more data in order to improve the ML component of the objective function, our method retains a roughly $6\%$ improvement for the objective function and a $25\%$ improvement for the energy consumption against both baselines. 


\vspace{-3mm}
\subsection{Trajectory Optimization with Model/Concept Drift} 
\label{ss:swarm_trajectories}
\begin{figure}[t] 
    \centering
    \includegraphics[width=0.49 \textwidth]{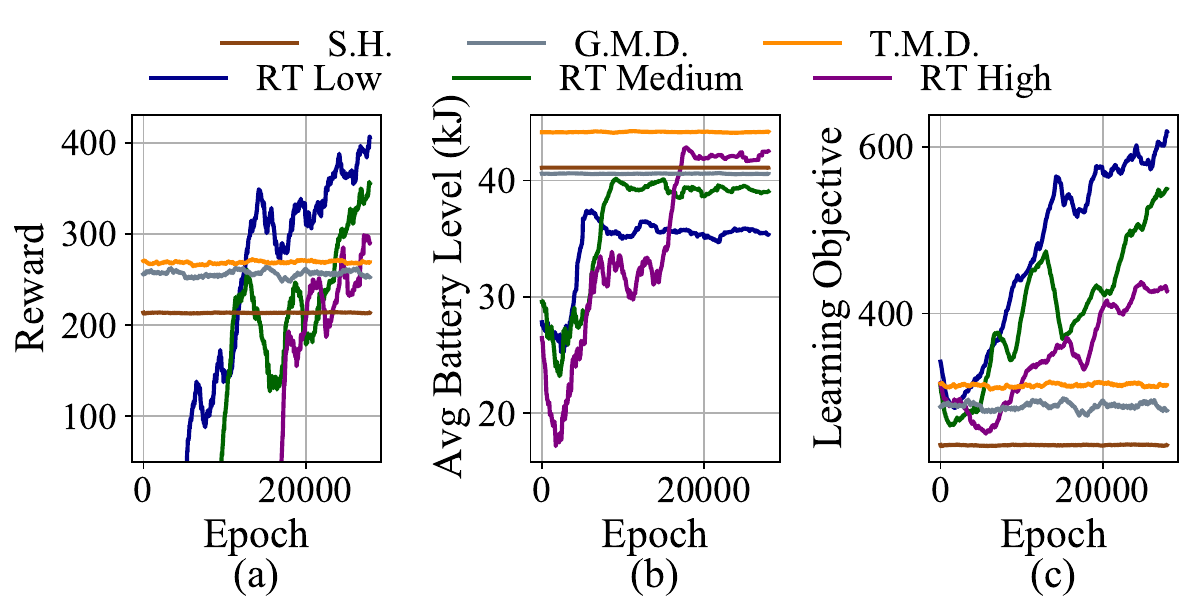}
    \vspace{-6mm}
    \caption{{\color{black}Comparison of our method for swarm trajectory and temporal ML design against several baselines. 
    In the legend, the first row denotes baseline algorithms and the second row denotes our method under three separate recharging thresholds (RT Low, RT Medium, and RT High). 
    For all cases, our method learns optimal patterns for efficient swarm trajectories and outperforms the baselines. } }
    \label{fig:DRL_performance}
    \vspace{-4mm}
\end{figure}

\vspace{-.9mm}
Next, we turn to the trajectory optimization component of Sec.~\ref{sec:swarmTraj} (green block in Fig.~\ref{fig:overall_flow}). 
We consider $2$ recharging stations and $8$ device clusters separated by distances in $[500,2000]$m, where each device cluster has $[8,10]$ devices. At the UAV level, we consider $3$ swarms, each of which has $[3,5]$ workers and $[1,2]$ coordinators. 
We refer to recharging stations as R:1 and R:2, and clusters as C:1, $\cdots$, C:8. 
First, we evaluate the performance of the DRL methodology by calculating the moving average reward from~\eqref{eq:rewardDRL}, average UAV battery levels, and learning objective 
(sum of the objective function $O(s)$ and the estimated online gradient $G_c(s)$) 
for our RNN-based DRL method with $\epsilon=\gamma=0.7$ and three baselines in Fig.~\ref{fig:DRL_performance}. 
Since existing baselines for our problem do not exist, we developed three baseline algorithms: (i) sequential heuristic (S.H.), (ii) greedy minimum distance (G.M.D.), and (iii) threshold minimum distance (T.M.D.), and calculated their rewards using the reward function in~\eqref{eq:rewardDRL}. 
We explain these baseline algorithms in  Appendix~\ref{sec:app_sim_settings}. 
{\color{black}To analyze the impact of UAV downtime due to recharging, we consider our methodology under three values of $E^{\mathsf{T}}$, i.e., battery recharging threshold (RT): low (16.88 kJ), medium (25.32 kJ), and high (33.76 kJ). 
A large RT requires UAV swarms to recharge more frequently, inducing downtime in place of active model training, while a small RT permits UAV swarms to participate in more training sequences per recharge, which should lead to higher reward and learning objective for the overall network. We present the behavior of the reward, average battery level, and learning objective of these three recharging thresholds in Fig.~\ref{fig:DRL_performance}. We quantify their respective recharging downtime by measuring the average recharging station visits per epoch: 0.125 for RT Low, 0.163 for RT Medium, and 0.228 for RT High, where a higher value indicates more downtime. We see that less recharging downtime (i.e., RT Low) indeed leads to the best overall reward and learning objective among test cases. We additionally note that our methodology is able to outperform each of the baselines by at least 7\% in reward and 30\% in learning objective across these values of RT.}
{\color{black}We further evaluate the sensitivity of our RNN-based DRL method to varying $\epsilon$ and $\gamma$ in Appendix~\ref{sec:app_sim_settings}.} 

As the other goal for our DRL-based trajectory optimization is adaptability to changing model drifts, we use the same network as that in Fig.~\ref{fig:DRL_performance} with time-varying model drifts for each device cluster, and measure the cluster visit rate per 1k epochs in Fig.~\ref{fig:visit_freq_all}.
Each cluster has a unique affine function to model its model drift growth. Initially, the 8 clusters have scaled model drifts of $[2,4,6,8,10,12,14,16]$, but end at $[15,15,21,11,12,17,19,17]$. So, initially, the cluster visit rate favors C:8 and C:7, which have the highest model drifts. {\color{black}However, as the epochs increase, the model drift begins skewing towards C:3, and our methodology responds by increasing its visit rate from $0.21$ to $0.45$. Inversely, C:4 has the smallest final model drift and our method is able to adjust its visit rate from $0.22$ to $0.17$, showing our methodology's ability to provide more network services and UAV swarms to those device clusters that need it the most.}

\vspace{-1mm}
\section{Conclusion and Future Work}
\noindent
We developed a holistic framework for integrating UAV swarm networks for online distributed machine learning. 
This involved a number of unique modelling decisions and analysis. We  proposed a swarm stratification architecture tailored for our distributed machine learning framework.
Our introduced distributed machine learning architecture, hierarchical nested personalized federated learning {\tt HN-PFL}, nests meta-function based gradient descent into local and global aggregations through the worker-leader-core network hierarchy, for which we characterized the performance bound. 
Finally, we proposed and developed a holistic framework for network-aware UAV-enabled model training, consisting of two intertwined parts: (i) data offloading and processing optimization, for which we developed a distributed algorithm with performance guarantee, and (ii) learning duration and trajectory design, for which we developed a solution based on deep reinforcement learning.

{\color{black}One important direction for future work is the problem of swarm dimensioning, i.e., optimizing the number and type of UAVs comprising each swarm. 
UAVs could be intelligently exchanged in-between training periods to further optimize the tradeoff between model performance and energy consumption.}

\begin{figure}[t] 
    \centering
    \includegraphics[width=0.45 \textwidth]{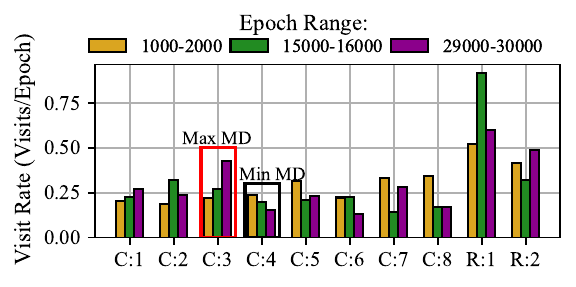}
    \vspace{-3mm}
    \caption{{\color{black}Swarm visit statistics per thousand epochs at various training stages. Our method learns to favor visiting device clusters with the largest model drift (MD), even when the model drift is time-varying. Over time, cluster C:3 has the largest model drift of $21$, and our method increases its visit rate from $0.21$ to $0.45$. Conversely, the final model drift at cluster C:4 is only $11$, and our method decreases its visit rate from $0.22$ to $0.17$.}}
    \label{fig:visit_freq_all}
    \vspace{-3mm}
\end{figure}

\vspace{-3mm}
\bibliographystyle{IEEEtran}
\bibliography{References}
\vspace{-13.5mm}
\begin{IEEEbiography}[\vspace{-11mm}{\includegraphics[width=0.8in,height=0.8in,clip]{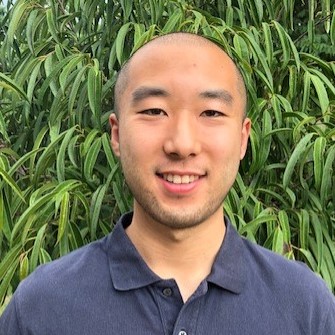}}]{Su Wang} is a PhD student in ECE at Purdue University. He received his BS in Electrical Engineering from Purdue in 2018. 
\end{IEEEbiography}
\vspace{-28.5mm}
\begin{IEEEbiography}[\vspace{-11mm}{\includegraphics[width=0.8in,height=0.8in,clip]{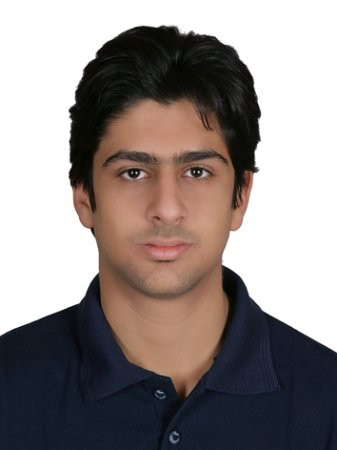}}]{Seyyedali Hosseinalipour} received his Ph.D. in EE from NCSU in 2020. He has won the 2020 ECE
Doctoral Scholar of the Year Award and 2021 ECE
Distinguished Dissertation Award at NCSU. He was a postdoctoral researcher at Purdue University from 2020 to 2022. He is currently an assistant professor of EE at University at Buffalo (SUNY).
\end{IEEEbiography}
\vspace{-24.5mm}
\begin{IEEEbiography}[\vspace{-11mm}{\includegraphics[width=0.8in,height=0.8in,clip]{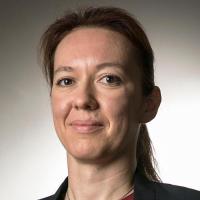}}]{Maria Gorlatova} is a Nortel Networks Assistant
Professor of ECE at Duke University. She received her PhD from Columbia University in 2013. She won the NSF CAREER Award, the ACM/IEEE IPSN Best Research Artifact Award, and the IEEE Communications Society Young Author Best Paper
Award.
\end{IEEEbiography}
\vspace{-23.5mm}
\begin{IEEEbiography}[\vspace{-11mm}{\includegraphics[width=0.8in,height=0.8in,clip]{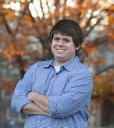}}]{Christopher G. Brinton} is an assistant professor of ECE at Purdue University. He received his Ph.D. in EE from Princeton University in 2016. He is the recipient of the NSF CAREER Award, the ONR Young Investigator Program Award, the DARPA Young Faculty Award, and the Intel Rising Star Faculty Award.
\end{IEEEbiography}
\vspace{-22.5mm}
\begin{IEEEbiography}[\vspace{-11mm}{\includegraphics[width=0.8in,height=0.8in,clip]{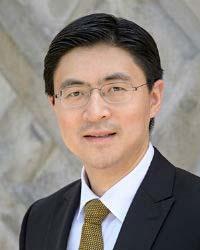}}]{Mung Chiang} is the John A. Edwardson Dean of the College of Engineering and Executive Vice President at Purdue University. Previously, he was the Arthur LeGrand Doty Professor of EE at Princeton University. He received his Ph.D. from Stanford University in 2003. He won the 2013 Alan T. Waterman Award, the highest honor to U.S. young scientists and engineers.
\end{IEEEbiography}

\newpage
\begingroup
\let\clearpage\relax 
\onecolumn
\appendices

\begin{changemargin}{-0.3cm}{-0.3cm} 


\section{{\color{black}Simulation Settings and Additional Simulations}}
\label{sec:app_sim_settings}
\textbf{Network Characteristics and Parameters.}
To calculate data rates among pairs of networked devices/UAVs, i.e.,~\eqref{eq:A2A},~\eqref{eq:G2A}, 
we use the following set of values~\cite{mozaffari2017mobile,zeng2017energy}:  $N_0=-174\textrm{dBm/Hz}$, $\alpha^{\mathsf{PL}}=2$, $\eta^{\textrm{LoS}}=3\textrm{dB}$, $\eta^{\textrm{NLoS}}=23\textrm{dB}$, $f = 2\textrm{GHz}$, $\psi=11.95, \beta=0.14$, $\overline{B}=2\textrm{MHz}$. 
We set the transmit power of the devices in the range $[23,25]\textrm{dBm}$ and the transmit power of UAVs to $20\textrm{dBm}$. The UAVs' altitude are also selected from $[25,30]\textrm{m}$. 
\newline

{\color{black}
To determine the coefficients presented in Assumption~1 (i.e. $B$, $\gamma^{\mathsf{H}}_u$, $\gamma^{\mathsf{G}}_u$, $\gamma^{\mathsf{H}}$, $\gamma^{\mathsf{G}}$, $\sigma^{\mathsf{G}}_j$, $\sigma^{\mathsf{H}}_j$), we performed ML model training on our datasets, which yielded empirical estimates of these values. Then to ensure that they are upper bounds for more general test cases, we increase them by an order of magnitude for our simulations.
To summarize, we used $B=500$, $\gamma^{\mathsf{H}}_u=0.1$, $\gamma^{\mathsf{G}}_u=0.05$, $\gamma^{\mathsf{H}}=0.1$, $\gamma^{\mathsf{G}}=0.05$, $\sigma^{\mathsf{G}}_j=50$, and $\sigma^{\mathsf{H}}_j=50$.}

\hfill \break

\textbf{Datasets.} We consider the MNIST (\url{http://yann.lecun.com/exdb/mnist/}), Fashion-MNIST (FMNIST) (\url{https://github.com/zalandoresearch/fashion-mnist}), {\color{black} CIFAR-10 (\url{http://www.cs.toronto.edu/~kriz/cifar.html})}, and {\color{black}RADIOML2016.10b~\cite{o2016radio} datasets.} 
MNIST and FMNIST datasets contain $70$K images ($60$K for training, $10$K for testing), where each image belongs to one of 10 labels of hand-written digits and fashion products, respectively. FMNIST can be considered as a harder classification task as compared to MNIST due to its more complex images. 
{\color{black}
While MNIST and FMNIST have grayscale images of 28x28 pixels, CIFAR-10 images are of 32x32 pixels and feature three channels (R,G,B), with the full dataset containing $60$K images ($50$K for training, $10$K for testing). 
On the other hand, RADIOML2016.10b (RADIOML) consists of waveform data, not images, and is commonly employed to assess ML algorithms for automatic modulation classification, i.e., classifying the modulation format of a particular signal.
Similarly to existing works on RADIOML (e.g.,~\cite{sahay2021deep}), we pre-process the data to obtain only the waveforms relevant to the CPFSK, GFSK, PAM4, and QPSK modulation schemes. 
Our resulting dataset size for RADIOML is $40$K waveforms ($32$K for training, $8$K for testing). 
For clarity, we reproduce a few constituent images from the datasets that we use for our simulations in Fig.~\ref{fig:dsets_exposition}.
\newline 

\textbf{Neural Network Architectures.}
For the MNIST and FMNIST datasets, we use a CNN with two convolutional layers followed by two linear layers. Both convolutional layers have kernel size 2, with the first layer outputting 16 maps and the second layer outputting 32 maps. The linear layers are correspondingly adjusted to fit the output dimensions of the second convolutional layer. 
For CIFAR-10 and RADIOML, we use a CNN with the same number of layers as MNIST and FMNIST, but with modified dimensions. 
Specifically, our CNN for CIFAR-10 uses kernel size 5 for the two convolutional layers, with 20 maps for the first convolutional layer and 50 maps for the second convolutional layer. 
Our CNN for RADIOML uses non-uniform kernel sizes of (2,5) and 16 maps for the first layer, and kernel sizes of (1,4) and 32 maps for the second layer.
The linear layers of our CNNs for CIFAR-10 and RADIOML are correspondingly adjusted to fit the output dimensions of the second convolutional layer. 
}

\hfill \break

\begin{figure}[h]
    \centering
    \includegraphics[width=0.8\textwidth]{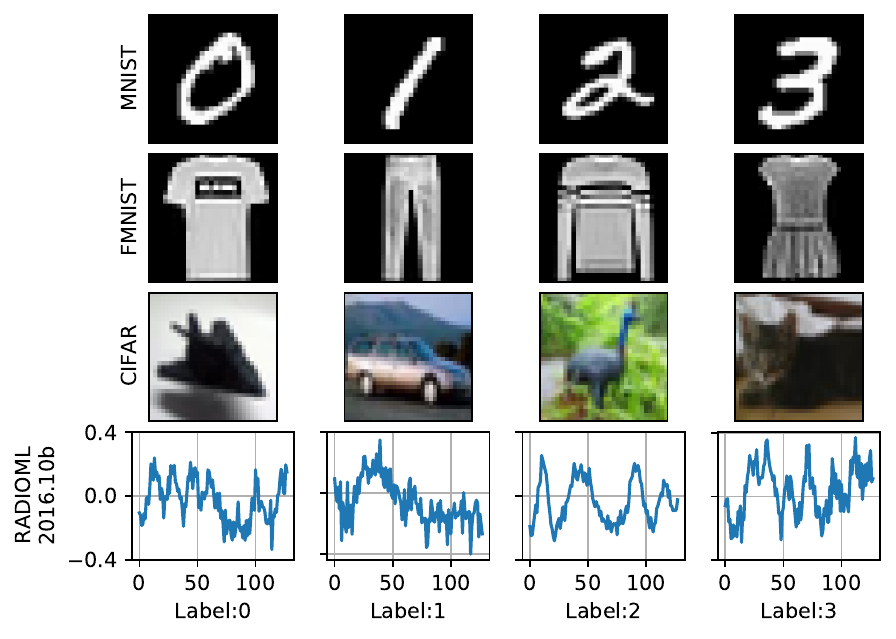}
    \vspace{-2mm}
    \caption{{\color{black}The datasets used for our simulations. We chose to use a classical digit recognition dataset for ML, MNIST, two object detection datasets, FMNIST (Fashion-MNIST) and CIFAR-10, an object detection task with RGB images, and a network management dataset RADIOML2016.10b (RADIOML).}}
    \label{fig:dsets_exposition}
\end{figure}

\textbf{Reinforcement Learning Integration and Training.}
To ensure that our reinforcement learning training is based on the results of the optimization of {\tt HN-PFL}, we simulate and then store the results of various configurations of our optimization formulation $\bm{\mathcal{P}}$. For each unique configuration (e.g., $T_s = 100, \tau_s^{\mathsf{L}} = \tau_s^{\mathsf{G}} = 2$), we save the ML model performance result as the learning reward $O(s)$, and utilize the energy consumed to complete training to update the minimum UAV battery levels $E_u(s)$ for a swarm. In this way, when we start training the RNN-based deep reinforcement learning agent, the agent determines the optimal swarm trajectories and temporal ML characteristics based on the results of {\tt HN-PFL} and our optimization formulation $\bm{\mathcal{P}}$.
\newline

To train the DRL agent, we use train-target DQNs with an update period of 20 epochs and an experience replay deque of size 20. As the experience replay storage is initially empty, we initialize the training by running 20 random instances to fill the experience replay deque with the swarm trajectories and temporal decisions for {\tt HN-PFL} and their associated reward. Thereafter, we train/update our DQN parameters after every epoch by calculating the mean square error (i.e., the loss) between the train and target network outputs, and then use gradient descent with learning rate $0.001$ to update the train DQN's parameters. After every 20 epochs of training, we re-synchronize the train and target DQNs' model parameters. 
\newline

At this point, our methodology will select subsequent swarm trajectory and temporal decisions for {\tt HN-PFL} based on an epsilon-greedy policy, with $\epsilon=0.7$. An epsilon-greedy policy means that, with probability $1-\epsilon$, the DQN is used to determine the next swarm trajectory and temporal decisions. 
As our $\epsilon$ is linearly decaying with the number of training epochs, reaching a minimum of $0.005$, our DQN is initially trained by biasing against random results, and, eventually when the DQN is well trained, our methodology will primarily adjust the DQN based on its own decisions. 
\newline

Our swarm/UAV flight and communication characteristics used for the reinforcement learning training are the same as those used in Fig.~\ref{fig:mnist_fixed_swarm_aggs}-~\ref{fig:fmnist_fixed_global_aggs}. For the additional reward coefficients found in~\eqref{eq:rewardDRL}, we use: $C = 10^5$, $c_1 = 0.2$, $c_2 = 0.25$, and $c_3 = 0.005$.
\newline

{\color{black}
\textbf{Baseline Comparison Algorithms for Reinforcement Learning Method.}
In sec.~\ref{sec:simulations}, we compare our RNN-based DRL methodology against three baseline algorithms: (i) sequential heuristic (S.H.), (ii) greedy minimum distance (G.M.D.), and (iii) threshold minimum distance (T.M.D.). 
Due to space constraints, we explain them here.
S.H. cycles through the clusters sequentially from C:1 to R:2, in the order presented in the x-axis of Fig.~\ref{fig:visit_freq_all}. 
On the other hand, our minimum distance methods determine the next destination for a swarm based on closest proximity, with G.M.D. always selecting the minimum distance and T.M.D. choosing between a random option or the minimum distance based on a probability threshold (set at $10\%$ probability to select the random option). Both minimum distance based methods will reroute swarms to the nearest recharging station when needed.} 
\newline

\textbf{{\color{black}Additional Simulations.}}
{\color{black}In the following subsection, we will first present additional simulations on Fashion-MNIST (FMNIST) and RADIOML2016.10b (RADIOML) to verify the superiority of our {\tt HN-PFL} methodology. 
Then, we will further demonstrate the veracity of our RNN-based DRL methodology via additional simulations, which separately vary $\epsilon$ and $\gamma$.}
\newline

{\color{black}{\tt HN-PFL} \textit{Proof of Concept:}} 
{\color{black}We further validate the improvement in accuracy and energy obtained by the {\tt HN-PFL} methodology on FMNIST and RADIOML, and provide these simulation results here. 
Similar to the experiments for MNIST and CIFAR-10, we consider a network composed of 4
UAV swarms with 2-3 workers, where each swarm has data
from only 3 labels (thus, non-iid data distributions) and data
quantity determined randomly from a Gaussian distribution:
N(3500, 350) for FMNIST and N(4500, 450) for RADIOML. 
The comparisons of classification accuracy for FMNIST are presented in Fig.~\ref{fig:fmnist_fixed_swarm_aggs} and~\ref{fig:fmnist_fixed_global_aggs}, and the comparison of energy consumption is presented in Table~\ref{tab:personalized_fmnist}. 
We see that our {\tt HN-PFL} attains at least 5\% better final classification accuracy relative to the baseline {\tt H-FL} for variety of $\tau^{\mathsf{G}}_s$ and $\tau^{\mathsf{L}}_s$ combinations. Table~\ref{tab:personalized_fmnist} also shows that {\tt HN-PFL} attains more than $35\%$ energy savings over {\tt H-FL} for a variety of $\tau^{\mathsf{G}}_s$ and $\tau^{\mathsf{L}}_s$ combinations to reach 45\% classification accuracy. 
\newline

{\tt HN-PFL} performs similarly well on RADIOML, the results of which are shown in Fig.~\ref{fig:radioml_fixed_swarm_aggs} and~\ref{fig:radioml_fixed_global_aggs} and Table~\ref{tab:personalized_radioml}. 
We note that, on RADIOML, the training for both {\tt HN-PFL} and {\tt H-FL} started from a starter model, which was trained centrally via meta-gradient descent for 10 iterations. 
Fig.~\ref{fig:radioml_fixed_swarm_aggs} and~\ref{fig:radioml_fixed_global_aggs} show that {\tt HN-PFL} attains at least 7\% improvement over the baseline {\tt H-FL} for a variety of $\tau^{\mathsf{G}}_s$ and $\tau^{\mathsf{L}}_s$ combinations.
Furthermore, due to its faster convergence speed, {\tt HN-PFL} saves over $48\%$ of the energy used by {\tt H-FL} to reach $80\%$ classification accuracy. 
}
\newline

\begin{figure}[h]
    \centering
    \begin{minipage}[h]{0.48\textwidth}
    \includegraphics[width=\textwidth]{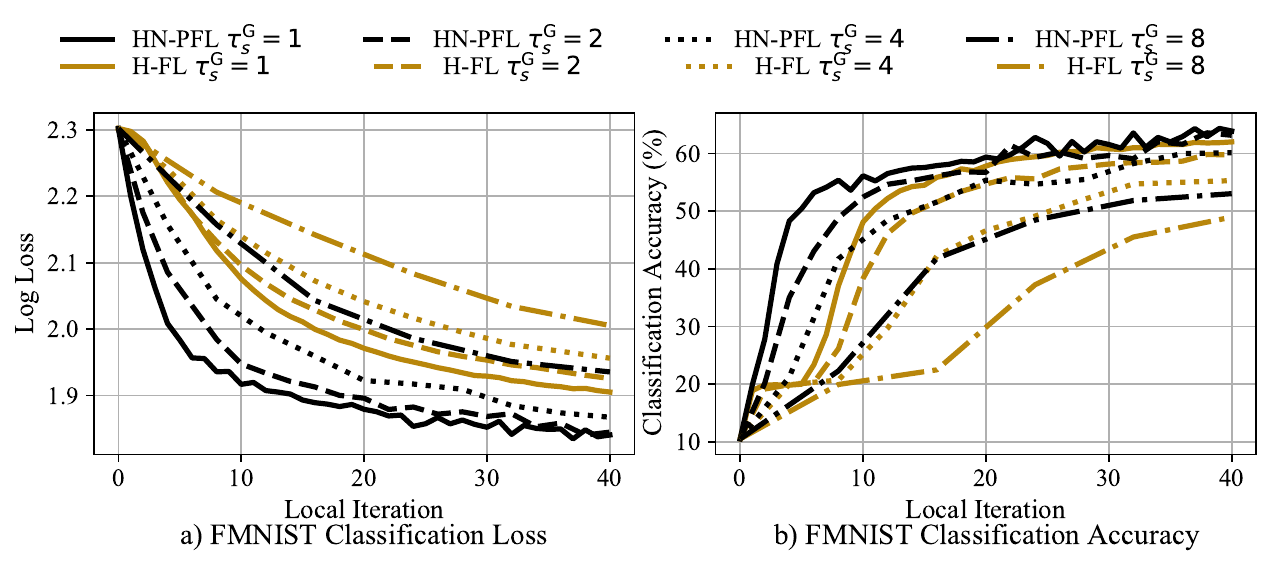}
    \caption{{\color{black}ML architecture comparisons for FMNIST with fixed $\tau_s^{\mathsf{L}} = 1$. {\tt HN-PFL} demonstrates faster convergence than {\tt H-FL} for various $\tau_s^{\mathsf{G}}$.}}
    \label{fig:fmnist_fixed_swarm_aggs}
    \end{minipage}
    \hfill
    \begin{minipage}[h]{0.48\textwidth}
    \includegraphics[width=\textwidth]{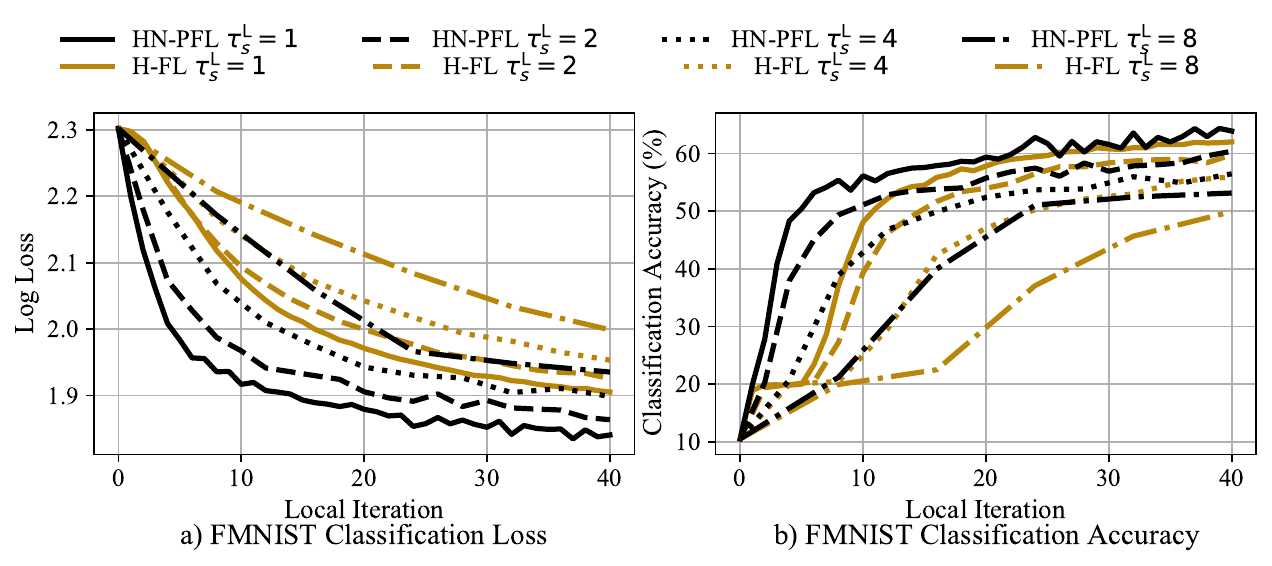}
    \caption{{\color{black}Using the same network for FMNIST as Fig.~\ref{fig:fmnist_fixed_swarm_aggs}, this experiment fixes $\tau_s^{\mathsf{G}} = 1$ to vary $\tau_s^{\mathsf{L}}$ instead. Our methodology {\tt HN-PFL} converges more rapidly than {\tt H-FL} for all test cases.}}
    \label{fig:fmnist_fixed_global_aggs}
    \end{minipage}
\end{figure}

\begin{table}[!t]
\caption{{\color{black}Energy Consumption for FMNIST to Reach $45\%$ Classification Accuracy}} 
\label{tab:personalized_fmnist} 
\begin{tabularx}{0.92\textwidth}{
p{\dimexpr.05\linewidth-2\tabcolsep-1.3333\arrayrulewidth}
p{\dimexpr.05\linewidth-2\tabcolsep-1.3333\arrayrulewidth}
p{\dimexpr.1\linewidth-2\tabcolsep-1.3333\arrayrulewidth}
p{\dimexpr.12\linewidth-2\tabcolsep-1.3333\arrayrulewidth}
p{\dimexpr.15\linewidth-2\tabcolsep-1.3333\arrayrulewidth}
| 
p{\dimexpr.05\linewidth-2\tabcolsep-1.3333\arrayrulewidth}
p{\dimexpr.05\linewidth-2\tabcolsep-1.3333\arrayrulewidth}
p{\dimexpr.1\linewidth-2\tabcolsep-1.3333\arrayrulewidth}
p{\dimexpr.12\linewidth-2\tabcolsep-1.3333\arrayrulewidth}
p{\dimexpr.15\linewidth-2\tabcolsep-1.3333\arrayrulewidth}
}
\toprule[.2em]
\multicolumn{2}{c}{\bf{Ratio}} & \multicolumn{3}{c}{\bf{FMNIST (kJ)}} & \multicolumn{2}{c}{\bf{Ratio}} & \multicolumn{3}{c}{\bf{FMNIST (kJ)}} \\
\cmidrule(lr){1-2} \cmidrule(lr){3-5} \cmidrule(lr){6-7} \cmidrule(lr){8-10} 
$\tau_s^{\mathsf{L}}$ & $\tau_s^{\mathsf{G}}$ & \bf{{\scriptsize HFL}} & \bf{{\scriptsize HNPFL}} & \bf{{\scriptsize Savings}} 
& $\tau_s^{\mathsf{L}}$ & $\tau_s^{\mathsf{G}}$ & \bf{{\scriptsize HFL}} & \bf{{\scriptsize HNPFL}} & \bf{{\scriptsize Savings}} \\
\midrule
1 & 1 & 5.22 & 2.32 & 55.6\% 
& 1 & 1 & 5.22 & 2.32 & 55.6\% \\ 
1 & 2 & 6.96 & 3.48 & 50.0\% 
& 2 & 1 & 6.96 & 4.06 & 41.7\% \\
1 & 4 & 11.03 & 6.38 & 42.2\% 
& 4 & 1 & 11.61 & 7.55 & 35.0\% \\ 
1 & 8 & 17.99 & 11.61 & 35.5\% 
& 8 & 1 & 18.57 & 11.61 & 37.5\% \\ 
\bottomrule
\end{tabularx}
\end{table}

\begin{figure}[h!]
    \centering
    \begin{minipage}[h]{0.48\textwidth}
    \includegraphics[width=\textwidth]{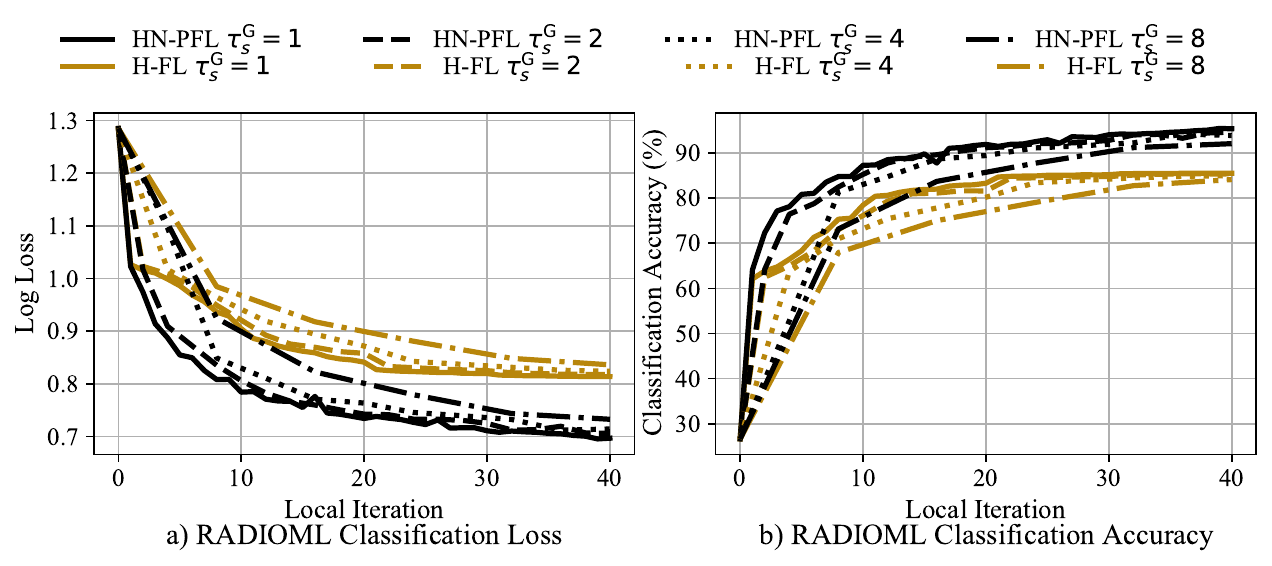}
    \caption{{\color{black}ML architecture comparisons for RADIOML with fixed $\tau_s^{\mathsf{L}} = 1$. {\tt HN-PFL} demonstrates faster convergence than {\tt H-FL} for various $\tau_s^{\mathsf{G}}$.}} 
    \label{fig:radioml_fixed_swarm_aggs}
    \end{minipage}
    \hfill
    \begin{minipage}[h]{0.48\textwidth}
    \includegraphics[width=\textwidth]{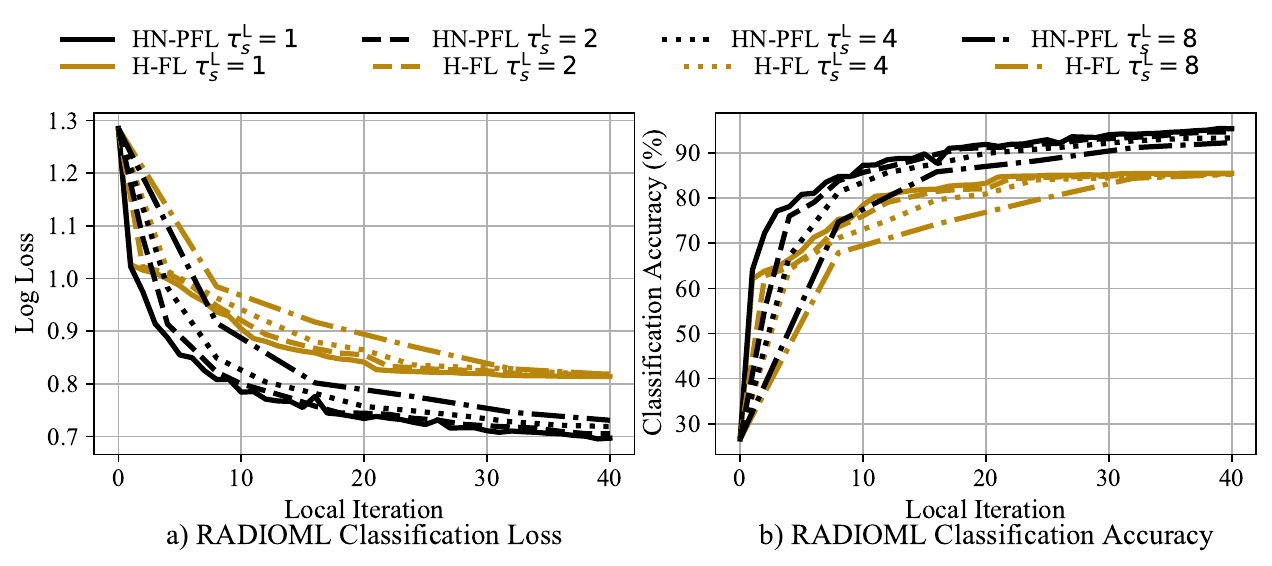}
    \caption{{\color{black}Using the same network for RADIOML as Fig.~\ref{fig:radioml_fixed_swarm_aggs}, this experiment fixes $\tau_s^{\mathsf{G}} = 1$ to vary $\tau_s^{\mathsf{L}}$ instead. Our methodology {\tt HN-PFL} converges more rapidly than {\tt H-FL} for all test cases.}}
    \label{fig:radioml_fixed_global_aggs}
    \end{minipage}
\end{figure}

\begin{table}[h!]
\caption{{\color{black}Energy Consumption for RADIOML to Reach $80\%$ Classification Accuracy}}
\label{tab:personalized_radioml} 
\begin{tabularx}{0.92\textwidth}{
p{\dimexpr.05\linewidth-2\tabcolsep-1.3333\arrayrulewidth}
p{\dimexpr.05\linewidth-2\tabcolsep-1.3333\arrayrulewidth}
p{\dimexpr.1\linewidth-2\tabcolsep-1.3333\arrayrulewidth}
p{\dimexpr.12\linewidth-2\tabcolsep-1.3333\arrayrulewidth}
p{\dimexpr.15\linewidth-2\tabcolsep-1.3333\arrayrulewidth}
| 
p{\dimexpr.05\linewidth-2\tabcolsep-1.3333\arrayrulewidth}
p{\dimexpr.05\linewidth-2\tabcolsep-1.3333\arrayrulewidth}
p{\dimexpr.1\linewidth-2\tabcolsep-1.3333\arrayrulewidth}
p{\dimexpr.12\linewidth-2\tabcolsep-1.3333\arrayrulewidth}
p{\dimexpr.15\linewidth-2\tabcolsep-1.3333\arrayrulewidth}
}
\toprule[.2em]
\multicolumn{2}{c}{\bf{Ratio}} & \multicolumn{3}{c}{\bf{RADIOML (kJ)}} & \multicolumn{2}{c}{\bf{Ratio}} & \multicolumn{3}{c}{\bf{RADIOML (kJ)}} \\
\cmidrule(lr){1-2} \cmidrule(lr){3-5} \cmidrule(lr){6-7} \cmidrule(lr){8-10} 
$\tau_s^{\mathsf{L}}$ & $\tau_s^{\mathsf{G}}$ & \bf{{\scriptsize HFL}} & \bf{{\scriptsize HNPFL}} & \bf{{\scriptsize Savings}} 
& $\tau_s^{\mathsf{L}}$ & $\tau_s^{\mathsf{G}}$ & \bf{{\scriptsize HFL}} & \bf{{\scriptsize HNPFL}} & \bf{{\scriptsize Savings}} \\
\midrule
1 & 1 & 5.80 & 1.74 & 70.0\% 
& 1 & 1 & 5.80 & 1.74 & 70.0\% \\ 
1 & 2 & 6.96 & 2.90 & 58.3\% 
& 2 & 1 & 7.45 & 2.90 & 61.1\% \\
1 & 4 & 8.71 & 3.48 & 60.0\% 
& 4 & 1 & 9.29 & 4.06 & 56.3\% \\ 
1 & 8 & 14.51 & 6.96 & 52.0\% 
& 8 & 1 & 14.51 & 7.45 & 48.7\% \\ 
\bottomrule
\end{tabularx}
\vspace{-6mm}
\end{table}

{\color{black}
\textit{RNN-based DRL Method:} 
We further consider the sensitivity of our RNN-based DRL method for swarm trajectory and temporal ML design to the values of $\epsilon$ and $\gamma$ from Sec.~\ref{sec:swarmTraj}. We first evaluate varying $\epsilon$ from among $\{0.65,0.7,0.75\}$ in Fig.~\ref{fig:DRL_performance_epsilon} with constant $\gamma=0.7$ and then varying $\gamma$ from among $\{0.65,0.7,0.75\}$ in Fig.~\ref{fig:DRL_performance_gamma} with constant $\epsilon=0.7$. Foundational work, e.g.,~\cite{paine2020hyperparameter} has previously established that deep reinforcement learning techniques are sensitive to hyperparameters, such as our $\epsilon$ and $\gamma$ terms, as we observe here.
}
\newline

{\color{black}
The choice of $\epsilon$ influences the randomness with which actions are selected in the DRL method due to exploration, with a larger $\epsilon$ denoting more randomness and a smaller $\epsilon$ denoting less randomness. Additionally, the nominal value of $\epsilon$ gradually decreases during the training process, so a smaller $\epsilon$ means that the DRL method relies on the DRL agent more frequently and earlier on in the training process for decisions. A larger $\epsilon$ correspondingly means that the DRL method relies less on the DRL agent early on in the training process. 
For the choices $\epsilon$ shown in Fig.~\ref{fig:DRL_performance_epsilon}, we can see that our RNN-based DRL methodology is able to adapt and outperform the baselines over time.}
\newline

{\color{black}
The choice of $\gamma$ determines the value of future rewards on the current reward computation. A larger $\gamma$ means that the DRL agent will place greater emphasis on the anticipated future rewards during DRL agent training, while a smaller $\gamma$ results in DRL agent training that relies more on instantaneous rewards, i.e., the specific rewards as a result of the current action. 
For the choices $\gamma$ shown in Fig.~\ref{fig:DRL_performance_gamma}, we can see that our RNN-based DRL methodology is able to outperform the baselines for $\gamma \in \{0.7,0.75\}$ and match the baselines for $\gamma =0.65$ by the end of the training process. These results lead us to choose $\epsilon = 0.7, \gamma = 0.7$ for our other experiments.} 

\begin{figure}[h!] 
    \centering
    \includegraphics[width=0.49 \textwidth]{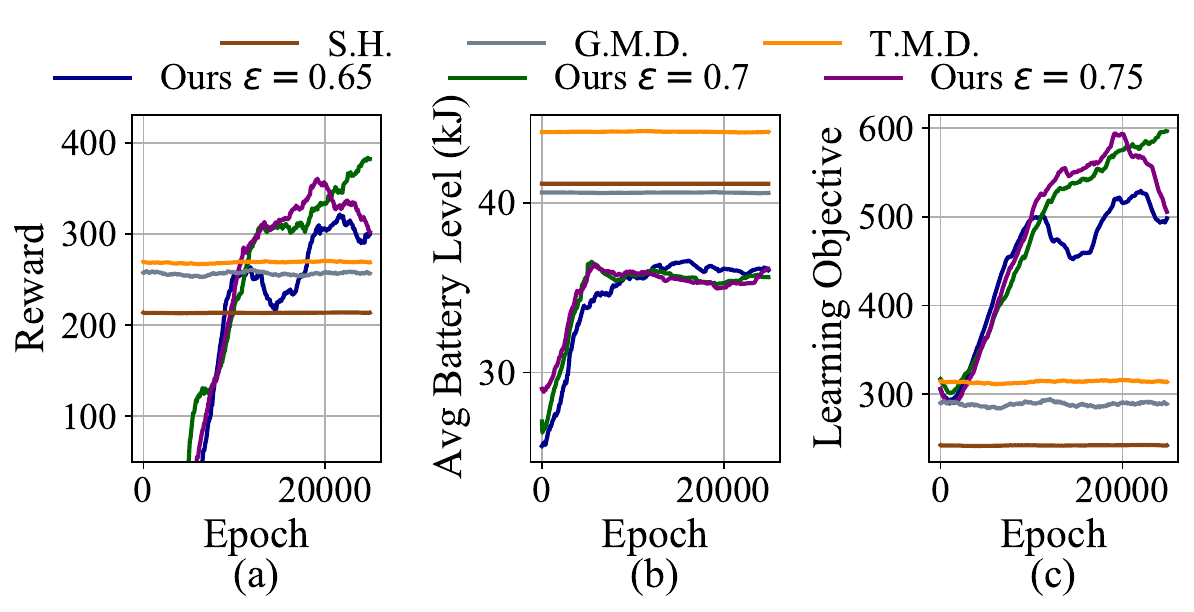}
    \caption{{\color{black}The impact of $\epsilon$ selection on our RNN-based DRL methodology for swarm trajectory and temporal ML design. 
    Our method outperforms the baselines over time for all three $\epsilon$ values.}}    
    \label{fig:DRL_performance_epsilon}
\end{figure}

\begin{figure}[h!] 
    \centering
    \includegraphics[width=0.49 \textwidth]{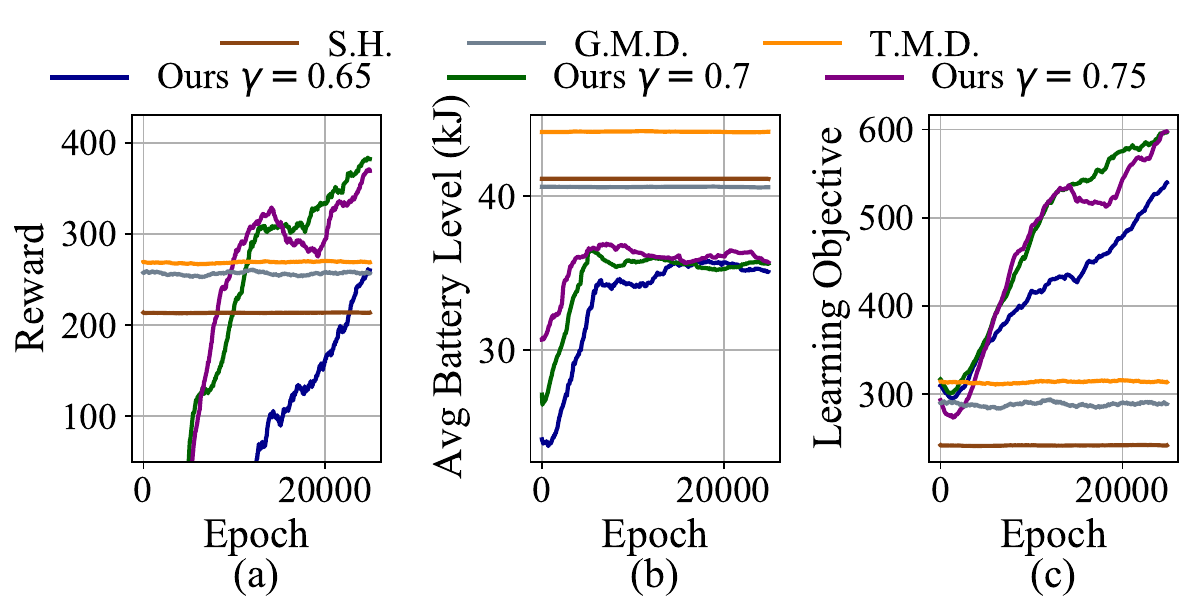}
    \caption{{\color{black}The impact of $\gamma$ selection on our RNN-based DRL methodology for swarm trajectory and temporal ML design. We see that our method either outperforms ($\gamma \in \{0.7,0.75\}$) or matches ($\gamma=0.65$) the baselines in terms of reward by the end of the training process for all three values of $\gamma$.}}
    \label{fig:DRL_performance_gamma}
\end{figure}


\clearpage
\newpage
\mbox{}
\clearpage

\section{{\color{black}A Discussion on UAV Signaling}}
\label{sec:signaling}
{\color{black}
UAV signaling, according to the highly cited paper~\cite{mozaffari2019tutorial}, is typically used to coordinate data/information transfers, though it can also be used to detect/classify the presence of a UAV~\cite{ezuma2019detection}. The specific design of a signaling layer for UAVs and devices is an important research problem, with works such as~\cite{yang2021rf} devoted to using RF signaling techniques to optimize UAV classification and~\cite{sheng2020secure,nguyen2020uav} that develop coding or modulation techniques to better determine UAV signaling. 
Below, we outline the communication sequence of {\tt HN-PFL} among devices, UAVs, and access points. 
For simplicity in the optimization formulation of $(\boldsymbol{\mathcal{P}})$, we ignore interference from simultaneous transmissions to the UAVs and devices as they are stationary and can use of orthogonal frequency bands for all communications. 
We note that existing work, e.g.,~\cite{shen2022joint}, in UAV networks for machine learning has yet to consider signaling details in-depth, which motivates a new research study with a comprehensive treatment of this topic. 
\newline

\textbf{Communications among different types of UAVs (leader, worker, and coordinator):} 
The communication between the different types of UAVs happens sequentially. 
When a UAV swarm arrives above a device cluster, the leader signals to the worker and coordinator UAVs to travel to fixed positions above the device cluster, and sends coordinator UAVs the data offloading ratios for their transmissions to worker UAVs. 
Upon arrival at their given destinations, the coordinator and worker UAVs signal to the leader UAV that they have arrived, and, once the leader UAV receives this signal from all coordinator and worker UAVs, the leader UAV signals the commencement of the training sequence and the swarm-wide aggregation frequency to the worker UAVs. 
Upon reception, the coordinator UAVs signal to relevant worker UAVs that they will initiate data transfers, and begin transferring data received from devices to the worker UAVs. 
At swarm-wide aggregation stages, the workers signal to their leader that they will begin transmitting ML model parameters, and then transfer their local ML model parameters to their leader. 
Once the leader receives all local parameter vectors, it performs a swarm-wide aggregation, synchronizes the ML model parameters across workers, and then signals initiation of the next training round. 
\newline

\textbf{UAV requests for data from devices:}
Coordinator UAVs are assigned specific devices, based on our optimization formulation in Sec.~IV, from which they collect data. 
After signaling to the leader UAV of their arrival at their fixed positions, coordinator UAVs send wake-up signals to their assigned devices to initiate data offloading. 
Devices then transmit their data to relevant coordinator UAVs. 
\newline

\textbf{Leader and access point interactions:}
At each global aggregation stage, the leader UAVs will travel to their nearest access point, and signal their arrival. 
Once the core network (via the APs) has received such a signal from all active leader UAVs, it will prompt leader UAVs to upload their swarm-wide ML model parameters. 
Once the core network has received all of these parameters, it will perform a global aggregation and synchronize the ML model parameters at leader UAVs with the result of the latest global aggregation. 
The leader UAVs will return to their respective swarms, upon synchronization completion, to commence the next training round.
\newline
}

\section{Sketch Proof of Lemma 1} \label{app:sk_l1}
\begin{skproof}
The proof uses similar techniques to those found in Lemma 4.3 in~\cite{fallah2020personalized}. We use {\color{black} our defined data processing ratios $\alpha_{j,1},\alpha_{j,2},\alpha_{j,3}$ in place for batch sizes within the data variability component of Assumption~\ref{eq:assumptions_all}.}
\end{skproof}

\section{Sketch Proof of Lemma 2} \label{app:sk_l2}
\begin{skproof}
The proof uses similar techniques to those found in Lemma 4.4 in~\cite{fallah2020personalized}. The difference is that we use the weighted aggregation definition for the meta function in~\eqref{eq:local_meta_function} before applying Assumption~\ref{eq:assumptions_all} and Jensen's inequality. 
\end{skproof}

\newpage

\section{Proof of Proposition 1} \label{app:prop2}

\begin{proof} \label{sec:app_prop1} 
To bound
\begin{equation} \label{eq:prop2_eq1}
\begin{aligned}
&\mathbb{E}\bigg[ \bigg\Vert \frac{1}{U(s)} \sum_{u \in \mathcal{U}(s)}  \sum_{j \in \mathcal{W}_u}  \frac{\Delta_j(k' \tau_s^{\mathsf{G}})}{\overline{\Delta}_u(k' \tau_s^{\mathsf{G}})}
\bigg(\mathbf{w}_{j}(t) - \overline{\mathbf{w}}(t)
\bigg) \bigg\Vert^2 \bigg],
\end{aligned}
\end{equation}

\noindent
where $t = t_s^{\mathsf{G}}(k') = t_s + k' \tau_s^{\mathsf{L}} \tau_s^{\mathsf{G}}$, we introduce $\overline{\mathbf{w}}_{u}(t)$ and omit the $t_s$ aspect of $t_s^{\mathsf{G}}(k')$, i.e., ignoring the starting time of the sequence $t_s$ in all the derivations since it is just a constant, to obtain:
\begin{equation} \label{eq:prop2_eq2}
\begin{aligned}
&\mathbb{E}\bigg[ \bigg\Vert 
\frac{1}{U(s)}\sum_{u \in \mathcal{U}(s)} \sum_{j \in \mathcal{W}_u} 
\frac{\Delta_j(k^{'} \tau_s^{\mathsf{G}})}{\overline{\Delta}_u(k^{'} \tau_s^{\mathsf{G}})}
\bigg(\mathbf{w}_{j}( \tau_s^{\mathsf{L}} \tau_s^{\mathsf{G}} k' ) - \overline{\mathbf{w}}_{u}( \tau_s^{\mathsf{L}} \tau_s^{\mathsf{G}} k' ) 
+ \overline{\mathbf{w}}_{u}(\tau_s^{\mathsf{L}} \tau_s^{\mathsf{G}} k') 
- \overline{\mathbf{w}}(\tau_s^{\mathsf{L}} \tau_s^{\mathsf{G}} k') \bigg) 
\bigg\Vert^2 \bigg] \\
&\leq 
\underbrace{2 \frac{1}{U(s)}\sum_{u \in \mathcal{U}(s)} \sum_{j \in \mathcal{W}_u} 
\frac{\Delta_j(k^{'} \tau_s^{\mathsf{G}})}{\overline{\Delta}_u(k^{'} \tau_s^{\mathsf{G}})}
\mathbb{E}[\Vert \mathbf{w}_j(\tau_s^{\mathsf{L}} \tau_s^{\mathsf{G}} k') 
- \overline{\mathbf{w}}_u(\tau_s^{\mathsf{L}} \tau_s^{\mathsf{G}} k') \Vert^2 ]  }_{(i)}
+ 
\underbrace{
2 \frac{1}{U(s)} \sum_{u \in \mathcal{U}(s)}
\mathbb{E}[\Vert \overline{\mathbf{w}}_u(\tau_s^{\mathsf{L}} \tau_s^{\mathsf{G}} k') 
- \overline{\mathbf{w}}(\tau_s^{\mathsf{L}} \tau_s^{\mathsf{G}} k')
\Vert^2] }_{(ii)},
\end{aligned}
\end{equation} 

\noindent
where the inequality is due to Jensen's inequality and $\Vert a+b \Vert^2 \leq 2(\Vert a \Vert^2 + \Vert b \Vert^2)$.
We first upper bound~\eqref{eq:prop2_eq2}(i), omitting the summations for simplicity:
\begin{equation} \label{eq:prop2_eq3}
\begin{aligned} 
& 2 \mathbb{E} [ \Vert \mathbf{w}_j(\tau_s^{\mathsf{L}} \tau_s^{\mathsf{G}} k') - \overline{\mathbf{w}}_{u}(\tau_s^{\mathsf{L}} \tau_s^{\mathsf{G}} k') \Vert^2 ]
\overset{(a)}{=} 2 \mathbb{E} \bigg[ \bigg\Vert \mathbf{w}_{j}(\tau_s^{\mathsf{L}} \tau_s^{\mathsf{G}} k') - \sum_{\hat{j} \in \mathcal{W}_u} 
\frac{ \Delta_{\hat{j}}(k^{'} \tau_s^{\mathsf{G}})}{\overline{\Delta}_u(k^{'}\tau_s^{\mathsf{G}})}
\mathbf{w}_{\hat{j}}(\tau_s^{\mathsf{L}} \tau_s^{\mathsf{G}} k') \bigg\Vert^2 \bigg] \\
& \overset{(b)}{=} 2 \mathbb{E} \bigg[ \bigg\Vert \mathbf{w}_{j}(\tau_s^{\mathsf{L}} \tau_s^{\mathsf{G}} k'-1) 
- \eta_2 \nabla \widetilde{F}_j (\mathbf{w}_{j}(\tau_s^{\mathsf{L}} \tau_s^{\mathsf{G}} k^{'} - 1))
- \sum_{\hat{j} \in \mathcal{W}_u} \frac{\Delta_{\hat{j}} (k^{'} \tau_s^{\mathsf{G}} ) }{\overline{\Delta}_u(k^{'}  \tau_s^{\mathsf{G}})} 
\bigg(\mathbf{w}_{\hat{j}}(\tau_s^{\mathsf{L}} \tau_s^{\mathsf{G}} k' -1) - \eta_2 \nabla \widetilde{ F}_{\hat{j}} (\mathbf{w}_{\hat{j}}(\tau_s^{\mathsf{L}} \tau_s^{\mathsf{G}} k' - 1)) \bigg)
\bigg\Vert^2 \bigg] \\
& \overset{(c)}{\leq} 
\underbrace{8 \mathbb{E} [ \Vert \mathbf{w}_{j}(\tau_s^{\mathsf{L}} \tau_s^{\mathsf{G}} k' - 1) - \overline{\mathbf{w}}_{u} (\tau_s^{\mathsf{L}} \tau_s^{\mathsf{G}} k' - 1) \Vert^2 ]}_{(i)}
+ 
\underbrace{8 \eta_2^2 
\mathbb{E} [ \Vert 
\nabla F_j(\mathbf{w}_{i}(\tau_s^{\mathsf{L}} \tau_s^{\mathsf{G}} k' - 1)) - \nabla \widetilde{F}_j (\mathbf{w}_{j}(\tau_s^{\mathsf{L}} \tau_s^{\mathsf{G}} k' - 1)) 
\Vert^2 ]}_{(ii)} \\
& + 
\underbrace{8 \eta_2^2 
\mathbb{E} \bigg[ \bigg\Vert 
\sum_{\hat{j} \in \mathcal{W}_u} \frac{\Delta_{\hat{j}}( k' \tau_s^{\mathsf{G}} ) }{\overline{\Delta}_u( k' \tau_s^{\mathsf{G}} )} \bigg( \nabla \widetilde{F}_{\hat{j}} (\mathbf{w}_{\hat{j}}(\tau_s^{\mathsf{L}} \tau_s^{\mathsf{G}} k' - 1)) - \nabla F_{\hat{j}}(\mathbf{w}_{\hat{j}}(\tau_s^{\mathsf{L}} \tau_s^{\mathsf{G}} k'-1)) \bigg) 
\bigg\Vert^2 \bigg]}_{(iii)} \\
& + 
\underbrace{8 \eta_2^2  
\mathbb{E} \bigg[ \bigg\Vert 
\nabla F_j(\mathbf{w}_{j}(\tau_s^{\mathsf{L}} \tau_s^{\mathsf{G}} k' -1)) - \sum_{\hat{j} \in \mathcal{W}_u} \frac{\Delta_{\hat{j}}(k' \tau_s^{\mathsf{G}}) }{\overline{\Delta}_{u}(k' \tau_s^{\mathsf{G}})} \nabla F_{\hat{j}}(\mathbf{w}_{\hat{j}}(\tau_s^{\mathsf{L}} \tau_s^{\mathsf{G}} k' - 1))
\bigg\Vert^2 \bigg]}_{(iv)},
\end{aligned}
\end{equation}
where $(a)$ is the aggregation rule of $\overline{\mathbf{w}}_u$, $(b)$ uses the gradient update from~\eqref{eq:pfl_meta_gradient}, and $(c)$ introduces $\eta_2 \nabla F_j$ and $\eta_2 \nabla \overline{F}_u$ terms and then applies Cauchy-Schwarz, i.e., $(\sum_i^n a_i)^2 \leq n(\sum_i^n a_i^2) $. 
The bounds for~\eqref{eq:prop2_eq3}(ii) and~\eqref{eq:prop2_eq3}(iii) follow immediately from Lemma~\ref{theory:lemma_sigma}, and we analyze~\eqref{eq:prop2_eq3}(iv), while recalling the summations from~\eqref{eq:prop2_eq2} as follows:
\begin{align} \label{eq:prop2_eq4}
& 8 \eta_2^2 
\frac{1}{U(s)}\sum_{u \in \mathcal{U}(s)} \sum_{j \in \mathcal{W}_u} \frac{\Delta_j(k' \tau_s^{\mathsf{G}} )}{\overline{\Delta}_u( k' \tau_s^{\mathsf{G}} )} 
\mathbb{E} \bigg[ \bigg\Vert  
\nabla F_j(\mathbf{w}_j( \tau_s^{\mathsf{L}} \tau_s^{\mathsf{G}} k' - 1)) 
- \sum_{\hat{j} \in \mathcal{W}_u} 
\frac{\Delta_{\hat{j}}(k'\tau_s^{\mathsf{G}})}{\overline{\Delta}_u(k' \tau_s^{\mathsf{G}})}
\nabla F_{\hat{j}}(\mathbf{w}_{\hat{j}}(\tau_s^{\mathsf{L}} \tau_s^{\mathsf{G}} k' - 1))
\bigg\Vert^2 \bigg] \\ \nonumber
& \overset{(a)}{=}
8 \eta_2^2 
\frac{1}{U(s)} \sum_{u \in \mathcal{U}(s) } \sum_{j \in \mathcal{W}_u} \frac{\Delta_j(k' \tau_s^{\mathsf{G}})}{\overline{\Delta}_u(k' \tau_s^{\mathsf{G}})} 
\mathbb{E} \bigg[ \bigg\Vert  
\nabla F_j(\mathbf{w}_{j}(\tau_s^{\mathsf{L}} \tau_s^{\mathsf{G}} k' -1)) 
- \nabla F_j(\overline{\mathbf{w}}_{u}(\tau_s^{\mathsf{L}} \tau_s^{\mathsf{G}} k' -1))
+ \nabla F_j(\overline{\mathbf{w}}_{u}(\tau_s^{\mathsf{L}} \tau_s^{\mathsf{G}} k' -1))
\\ \nonumber
& 
- \sum_{\hat{j} \in \mathcal{W}_u} \frac{\Delta_{\hat{j}}(k' \tau_s^{\mathsf{G}})}{\overline{\Delta}_{u}(k' \tau_s^{\mathsf{G}})} 
\bigg( 
\nabla F_{\hat{j}}(\mathbf{w}_{\hat{j}}(\tau_s^{\mathsf{L}} \tau_s^{\mathsf{G}} k' - 1)) 
+ \nabla F_{\hat{j}}(\overline{\mathbf{w}}_{u}(\tau_s^{\mathsf{L}} \tau_s^{\mathsf{G}} k' - 1))
- \nabla F_{\hat{j}}(\overline{\mathbf{w}}_{u}(\tau_s^{\mathsf{L}} \tau_s^{\mathsf{G}} k' - 1))
\bigg)
\bigg\Vert^2 \bigg] \\ \nonumber
& \overset{(b)}{\leq} 
24 \eta_2^2  
\frac{1}{U(s)}\sum_{u \in \mathcal{U}(s)} \sum_{j \in \mathcal{W}_u} \frac{\Delta_j(k' \tau_s^{\mathsf{G}})}{\overline{\Delta}_u(k' \tau_s^{\mathsf{G}})}
\mathbb{E} [ \Vert  
\nabla F_j(\mathbf{w}_{j}(\tau_s^{\mathsf{L}} \tau_s^{\mathsf{G}} k' -1)) 
- \nabla F_j(\overline{\mathbf{w}}_{u}(\tau_s^{\mathsf{L}} \tau_s^{\mathsf{G}} k' -1)) 
\Vert^2 ]
\\ \nonumber
& + 24 \eta_2^2 
\frac{1}{U(s)}\sum_{u \in \mathcal{U}(s)} \sum_{j \in \mathcal{W}_u} \frac{\Delta_j(k' \tau_s^{\mathsf{G}})}{\overline{\Delta}_u(k' \tau_s^{\mathsf{G}})}
\mathbb{E} \bigg[ \bigg\Vert  \sum_{\hat{j} \in \mathcal{W}_u} 
\frac{\Delta_{\hat{j}}(k' \tau_s^{\mathsf{G}})}{ \overline{\Delta}_{u}(k' \tau_s^{\mathsf{G}})}
\bigg(
\nabla F_{\hat{j}}(\overline{\mathbf{w}}_{u}(\tau_s^{\mathsf{L}} \tau_s^{\mathsf{G}} k' - 1))
- \nabla F_{\hat{j}}(\mathbf{w}_{\hat{j}}(\tau_s^{\mathsf{L}} \tau_s^{\mathsf{G}} k' - 1)) 
\bigg)
\bigg\Vert^2 \bigg] \\ \nonumber
& + 24 \eta_2^2 
\frac{1}{U(s)}\sum_{u \in \mathcal{U}(s)}  \sum_{j  \in \mathcal{W}_u}\frac{\Delta_j(k' \tau_s^{\mathsf{G}})}{\overline{\Delta}_u(k' \tau_s^{\mathsf{G}})} 
\mathbb{E} \bigg[ \bigg\Vert \nabla F_j(\overline{\mathbf{w}}_{u}(\tau_s^{\mathsf{L}} \tau_s^{\mathsf{G}} k' -1)) 
- \sum_{\hat{j} \in \mathcal{W}_u} \frac{\Delta_{\hat{j}}(k' \tau_s^{\mathsf{G}}) }{\overline{\Delta}_{u}(k' \tau_s^{\mathsf{G}})} 
\nabla F_{\hat{j}}(\overline{\mathbf{w}}_u(\tau_s^{\mathsf{L}} \tau_s^{\mathsf{G}} k' - 1))
\bigg \Vert^2 \bigg] \\ \nonumber
& \overset{(c)}{\leq} 
24 \eta_2^2  
\frac{1}{U(s)}\sum_{u \in \mathcal{U}(s)} \sum_{j  \in \mathcal{W}_u}\frac{\Delta_j(k' \tau_s^{\mathsf{G}})}{\overline{\Delta}_u(k' \tau_s^{\mathsf{G}})}
\mathbb{E} [ \Vert  
\nabla F_j(\mathbf{w}_{j}(\tau_s^{\mathsf{L}} \tau_s^{\mathsf{G}} k' -1)) 
- \nabla F_j(\overline{\mathbf{w}}_{u}(\tau_s^{\mathsf{L}} \tau_s^{\mathsf{G}} k' -1)) 
\Vert^2 ]
\\ \nonumber
& + 24 \eta_2^2 
\frac{1}{U(s)}\sum_{u \in \mathcal{U}(s)} \sum_{j  \in \mathcal{W}_u} \frac{\Delta_j(k' \tau_s^{\mathsf{G}})}{\overline{\Delta}_u(k' \tau_s^{\mathsf{G}})}
\sum_{\hat{j} \in \mathcal{W}_u} \frac{\Delta_{\hat{j}}(k' \tau_s^{\mathsf{G}})}{\overline{\Delta}_{u}(k' \tau_s^{\mathsf{G}})}
\mathbb{E} \bigg[ \bigg\Vert 
\nabla F_{\hat{j}}(\overline{\mathbf{w}}_u(\tau_s^{\mathsf{L}} \tau_s^{\mathsf{G}} k' - 1))
- \nabla F_{\hat{j}}(\mathbf{w}_{\hat{j}}(\tau_s^{\mathsf{L}} \tau_s^{\mathsf{G}} k' - 1)) 
\bigg\Vert^2 \bigg]
\\ \nonumber
& 
+ 24 \eta_2^2 
\frac{1}{U(s)}\sum_{u \in \mathcal{U}(s)}  \sum_{j  \in \mathcal{W}_u}\frac{\Delta_j(k' \tau_s^{\mathsf{G}})}{\overline{\Delta}_u(k' \tau_s^{\mathsf{G}})} 
\mathbb{E} \bigg[ \bigg\Vert \nabla F_j(\overline{\mathbf{w}}_{u}(\tau_s^{\mathsf{L}} \tau_s^{\mathsf{G}} k' -1)) 
- \nabla \overline{F}_{u}(\overline{\mathbf{w}}_u(\tau_s^{\mathsf{L}} \tau_s^{\mathsf{G}} k' - 1))
\bigg \Vert^2 \bigg] \\ \nonumber
& \overset{(d)}{\leq} 
24 \eta_2^2 \frac{1}{U(s)}\sum_{u \in \mathcal{U}(s)} \sum_{j  \in \mathcal{W}_u}\frac{\Delta_j(k' \tau_s^{\mathsf{G}})}{\overline{\Delta}_u(k' \tau_s^{\mathsf{G}})} \mu_F^2 
\mathbb{E} [ \Vert \mathbf{w}_{j}(\tau_s^{\mathsf{L}} \tau_s^{\mathsf{G}} k' -1) - \overline{\mathbf{w}}_u (\tau_s^{\mathsf{L}} \tau_s^{\mathsf{G}} k' -1) \Vert^2 ]
\\ \nonumber
& + 24 \eta_2^2 
\frac{1}{U(s)}\sum_{u \in \mathcal{U}(s)} 
\sum_{\hat{j} \in \mathcal{W}_u} \frac{\Delta_{\hat{j}}(k' \tau_s^{\mathsf{G}})}{\overline{\Delta}_{u}(k' \tau_s^{\mathsf{G}})}
\mu_F^2 
\mathbb{E} [ \Vert \overline{\mathbf{w}}_u(\tau_s^{\mathsf{L}} \tau_s^{\mathsf{G}} k' -1) - \mathbf{w}_{\hat{j}}(\tau_s^{\mathsf{L}} \tau_s^{\mathsf{G}} k' -1) \Vert^2 ] 
\\ \nonumber
& 
+ 24 \eta_2^2 
\frac{1}{U(s)}\sum_{u \in \mathcal{U}(s)}  \sum_{j  \in \mathcal{W}_u}\frac{\Delta_j(k' \tau_s^{\mathsf{G}})}{\overline{\Delta}_u(k' \tau_s^{\mathsf{G}})} 
\mathbb{E} \bigg[ \bigg \Vert \nabla F_j(\overline{\mathbf{w}}_{u}(\tau_s^{\mathsf{L}} \tau_s^{\mathsf{G}} k' -1)) 
- \nabla \overline{F}_{u}(\overline{\mathbf{w}}_u(\tau_s^{\mathsf{L}} \tau_s^{\mathsf{G}} k' - 1))
\bigg \Vert^2 \bigg]  \\ \nonumber
& \overset{(e)}{\leq} 
48 \mu_F^2 \eta_2^2  \frac{1}{U(s)}\sum_{u \in \mathcal{U}(s)} \sum_{j  \in \mathcal{W}_u}
\frac{\Delta_j(k' \tau_s^{\mathsf{G}})}{\overline{\Delta}_u(k' \tau_s^{\mathsf{G}})}
\mathbb{E} [ \Vert \mathbf{w}_{j}(\tau_s^{\mathsf{L}} \tau_s^{\mathsf{G}} k' -1) - \overline{\mathbf{w}}_u (\tau_s^{\mathsf{L}} \tau_s^{\mathsf{G}} k' -1) \Vert^2 ]
+
24 \eta_2^2 \frac{1}{U(s)} \sum_{u \in \mathcal{U}(s)} (\gamma_{F}^{u})^2,
\end{align}
where (a) introduces $\nabla F_j(\overline{\mathbf{w}}_u)$ and $\nabla F_{\hat{j}}(\overline{\mathbf{w}}_u)$,
(b) applies Jensen's inequality, 
(c) recalls the aggregation rule of $\nabla \overline{F}_u$ and applies Jensen's inequality, 
(d) uses $\mu_F$-smoothness (i.e., $\Vert \nabla F_j(\mathbf{w}) - \nabla F_j (\mathbf{w}') \Vert \leq \mu_F \Vert \mathbf{w} - \mathbf{w}' \Vert$ $\forall i \in \cup_{u \in \mathcal{U}(s)} \mathcal{W}_u$), 
and (e) leverages Lemma~\ref{theory:lemma_gamma}.
Combining~\eqref{eq:prop2_eq3} and~\eqref{eq:prop2_eq4} yields: 
\begingroup
\allowdisplaybreaks
\begin{align}
& 2 \frac{1}{U(s)}\sum_{u \in \mathcal{U}(s)}  \sum_{j  \in \mathcal{W}_u} \frac{\Delta_j(k' \tau_s^{\mathsf{G}})} {\overline{\Delta}_u(k' \tau_s^{\mathsf{G}})}
\mathbb{E} [ \Vert \mathbf{w}_j(\tau_s^{\mathsf{L}} \tau_s^{\mathsf{G}} k') - \overline{\mathbf{w}}_u(\tau_s^{\mathsf{L}} \tau_s^{\mathsf{G}} k') \Vert^2 ] \label{eq:prop2_eq5} \\
& \overset{(a)}{\leq} 
8\frac{1}{U(s)}\sum_{u \in \mathcal{U}(s)}  \sum_{j  \in \mathcal{W}_u}\frac{\Delta_j(k' \tau_s^{\mathsf{G}})}{\overline{\Delta}_u(k' \tau_s^{\mathsf{G}})}
\mathbb{E} [ \Vert \mathbf{w}_j(\tau_s^{\mathsf{L}} \tau_s^{\mathsf{G}} k' - 1) - \overline{\mathbf{w}}_u (\tau_s^{\mathsf{L}} \tau_s^{\mathsf{G}} k' - 1) \Vert^2 ] \nonumber\\
& + 8 \eta_2^2
\frac{1}{U(s)}\sum_{u \in \mathcal{U}(s)}  \sum_{j  \in \mathcal{W}_u}\frac{\Delta_j(k' \tau_s^{\mathsf{G}})}{\overline{\Delta}_u(k' \tau_s^{\mathsf{G}})}
\mathbb{E} [ \Vert 
\nabla F_j(\mathbf{w}_j(\tau_s^{\mathsf{L}} \tau_s^{\mathsf{G}} k' - 1)) - \nabla \widetilde{ F}_j(\mathbf{w}_j(\tau_s^{\mathsf{L}} \tau_s^{\mathsf{G}} k' - 1)) 
\Vert^2 ] \nonumber\\
& + 8 \eta_2^2
\frac{1}{U(s)}\sum_{u \in \mathcal{U}(s)}  \sum_{j  \in \mathcal{W}_u}\frac{\Delta_j(k' \tau_s^{\mathsf{G}})}{\overline{\Delta}_u(k' \tau_s^{\mathsf{G}})}
\mathbb{E} \bigg[ \bigg\Vert 
\sum_{\hat{j} \in \mathcal{W}_u} \frac{\Delta_{\hat{j}}(k' \tau_s^{\mathsf{G}})}{\overline{\Delta}_u(k' \tau_s^{\mathsf{G}})} 
\bigg( \nabla \widetilde{ F}_{\hat{j}}(\mathbf{w}_{\hat{j}}(\tau_s^{\mathsf{L}} \tau_s^{\mathsf{G}} k'-1)) - \nabla F_{\hat{j}}(\mathbf{w}_{\hat{j}}(\tau_s^{\mathsf{L}} \tau_s^{\mathsf{G}} k'-1)) \bigg) 
\bigg\Vert^2 \bigg] \nonumber\\
&+ 8 \eta_2^2
\frac{1}{U(s)}\sum_{u \in \mathcal{U}(s)}  \sum_{j  \in \mathcal{W}_u}\frac{\Delta_j(k' \tau_s^{\mathsf{G}})}{\overline{\Delta}_u(k' \tau_s^{\mathsf{G}})}
\mathbb{E} \bigg[ \bigg\Vert  
\nabla F_j(\mathbf{w}_{j}(\tau_s^{\mathsf{L}} \tau_s^{\mathsf{G}} k' -1)) - \sum_{\hat{j} \in \mathcal{W}_u} \frac{\Delta_{\hat{j}}(k' \tau_s^{\mathsf{G}})}{\overline{\Delta}_{u}(k' \tau_s^{\mathsf{G}})} \nabla F_{\hat{j}}(\mathbf{w}_{\hat{j}}(\tau_s^{\mathsf{L}} \tau_s^{\mathsf{G}} k' - 1))
\bigg\Vert^2 \bigg] \nonumber\\
& \overset{(b)}{\leq}
8 \frac{1}{U(s)}\sum_{u \in \mathcal{U}(s)}  \sum_{j  \in \mathcal{W}_u}\frac{\Delta_j(k' \tau_s^{\mathsf{G}})}{\overline{\Delta}_u(k' \tau_s^{\mathsf{G}})}
\mathbb{E} [ \Vert \mathbf{w}_j(\tau_s^{\mathsf{L}} \tau_s^{\mathsf{G}} k' - 1) - \overline{\mathbf{w}}_u (\tau_s^{\mathsf{L}} \tau_s^{\mathsf{G}} k' - 1) \Vert^2 ] 
+ 
16 \eta_2^2 \frac{1}{U(s)}\sum_{u \in \mathcal{U}(s)}  \sum_{j  \in \mathcal{W}_u}\frac{\Delta_j(k' \tau_s^{\mathsf{G}})}{\overline{\Delta}_u(k' \tau_s^{\mathsf{G}})}
(\sigma_{j}^{\mathsf{F}}(\tau_s^{\mathsf{L}} \tau_s^{\mathsf{G}} k' -1))^2
\nonumber\\
& + 8 \eta_2^2
\frac{1}{U(s)}\sum_{u \in \mathcal{U}(s)}  \sum_{j  \in \mathcal{W}_u}\frac{\Delta_j(k' \tau_s^{\mathsf{G}})}{\overline{\Delta}_u(k' \tau_s^{\mathsf{G}})}
\mathbb{E} \bigg[ \bigg\Vert  
\nabla F_j(\mathbf{w}_{j}(\tau_s^{\mathsf{L}} \tau_s^{\mathsf{G}} k' -1)) - \sum_{\hat{j} \in \mathcal{W}_u} \frac{\Delta_{\hat{j}}(k' \tau_s^{\mathsf{G}}) }{\overline{\Delta}_{u}(k' \tau_s^{\mathsf{G}})} \nabla F_{\hat{j}}(\mathbf{w}_{\hat{j}}(\tau_s^{\mathsf{L}} \tau_s^{\mathsf{G}} k' - 1))
\bigg\Vert^2 \bigg] \nonumber \\
& \overset{(c)}{\leq}
(8+ 48 \eta_2^2 \mu_F^2)
\frac{1}{U(s)}\sum_{u \in \mathcal{U}(s)}  \sum_{j  \in \mathcal{W}_u}\frac{\Delta_j(k' \tau_s^{\mathsf{G}})}{\overline{\Delta}_u(k' \tau_s^{\mathsf{G}})}
\mathbb{E} [ \Vert \mathbf{w}_j(\tau_s^{\mathsf{L}} \tau_s^{\mathsf{G}} k' - 1) - \overline{\mathbf{w}}_u (\tau_s^{\mathsf{L}} \tau_s^{\mathsf{G}} k' - 1) \Vert^2 ] 
\nonumber\\
& + 16 \eta_2^2  \frac{1}{U(s)}\sum_{u \in \mathcal{U}(s)} \sum_{j  \in \mathcal{W}_u}\frac{\Delta_j(k' \tau_s^{\mathsf{G}})}{\overline{\Delta}_u(k' \tau_s^{\mathsf{G}})}  
(\sigma_{j}^{\mathsf{F}}(\tau_s^{\mathsf{L}} \tau_s^{\mathsf{G}} k' -1))^2
+
24 \eta_2^2
\frac{1}{U(s)} \sum_{u \in \mathcal{U}(s)} (\gamma_{F}^{u})^2,
\nonumber
\end{align}
\endgroup
where $(a)$ restates~\eqref{eq:prop2_eq3} with the previously omitted summations, $(b)$ applies Jensen's inequality and Lemma~\ref{theory:lemma_sigma}, and $(c)$ uses the result in~\eqref{eq:prop2_eq4}. Solving~\eqref{eq:prop2_eq5} recursively yields:
\begin{align} \label{eq:prop2_eq6}
& 2 \frac{1}{U(s)}\sum_{u \in \mathcal{U}(s)}  \sum_{j  \in \mathcal{W}_u}\frac{\Delta_j(k' \tau_s^{\mathsf{G}}) }{\overline{\Delta}_u(k' \tau_s^{\mathsf{G}})}  \mathbb{E} [ \Vert \mathbf{w}_j(\tau_s^{\mathsf{L}} \tau_s^{\mathsf{G}} k') - \overline{\mathbf{w}}_u(\tau_s^{\mathsf{L}} \tau_s^{\mathsf{G}} k') \Vert^2 ] \\
& 
\overset{(a)}{\leq }
16 \eta_2^2 \frac{1}{U(s)}\sum_{u \in \mathcal{U}(s)}  \sum_{j  \in \mathcal{W}_u} \frac{\Delta_j(k' \tau_s^{\mathsf{G}})}{\overline{\Delta}_u(k' \tau_s^{\mathsf{G}})} 
\bigg(
\sum_{y=0}^{\tau_s^{\mathsf{L}} -1}
(\sigma_{j}^{\mathsf{F}}(\tau_s^{\mathsf{L}} \tau_s^{\mathsf{G}} k' - 1 -y))^2 (8+48 \eta_2^2 \mu_F^2)^{y}
\bigg)
+ 24 \eta_2^2
\frac{1}{U(s)} \sum_{u \in \mathcal{U}(s)} (\gamma_{F}^{u})^2
\frac{1 - (8 + 48 \eta_2^2 \mu_F^2)^{\tau_s^{\mathsf{L}}}}{1- (8 + 48 \eta_2^2 \mu_F^2)}\\
&
\overset{(b)}{\leq}
\bigg(
16 \eta_2^2 \frac{1}{U(s)}\sum_{u \in \mathcal{U}(s)} \sum_{j  \in \mathcal{W}_u} \frac{\Delta_j(k' \tau_s^{\mathsf{G}})}{\overline{\Delta}_u(k' \tau_s^{\mathsf{G}})}
\sum_{y=0}^{\tau_s^{\mathsf{L}} -1}
(\sigma_{j}^{\mathsf{F}}(\tau_s^{\mathsf{L}} \tau_s^{\mathsf{G}} k' - 1 -y))^2
+
24 \eta_2^2 \frac{1}{U(s)} \sum_{u \in \mathcal{U}(s)} (\gamma_{F}^{u})^2
\bigg)
\frac{1 - (8+ 48 \eta_2^2 \mu_F^2)^{\tau_s^{\mathsf{L}}}}{1- (8 + 48 \eta_2^2 \mu_F^2)},
\end{align}
where $(a)$ is the result of recursion and the finite sum of geometric series, and $(b)$ uses $\sum_i a_i b_i \leq \sum_i a_i \sum_i b_i$ $\forall a_i, b_i \geq 0$. 
Next, we bound~\eqref{eq:prop2_eq2}(ii) as follows:
\begin{equation} \label{eq:prop2_eq7}
\begin{aligned}
& 2 \frac{1}{U(s)} \sum_{u \in \mathcal{U}(s)} 
\mathbb{E} [ \Vert \overline{\mathbf{w}}_u(\tau_s^{\mathsf{L}} \tau_s^{\mathsf{G}} k') - \overline{\mathbf{w}}(\tau_s^{\mathsf{L}} \tau_s^{\mathsf{G}} k') \Vert^2 ] 
\overset{(a)}{=} 
2 \frac{1}{U(s)} \sum_{u \in \mathcal{U}(s)} 
\mathbb{E} \bigg[ \bigg\Vert  \overline{\mathbf{w}}_u(\tau_s^{\mathsf{L}} \tau_s^{\mathsf{G}} k')
- \frac{1}{U(s)} \sum_{\hat{u} \in \mathcal{U}(s)}
\overline{\mathbf{w}}_{\hat{u}}(\tau_s^{\mathsf{L}} \tau_s^{\mathsf{G}} k') \bigg\Vert^2 \bigg] \\
& \overset{(b)}{=}
2 \frac{1}{U(s)} \sum_{u \in \mathcal{U}(s)} 
\mathbb{E} \bigg[ \bigg\Vert  \overline{\mathbf{w}}_u(\tau_s^{\mathsf{L}} \tau_s^{\mathsf{G}} k'-1)
- \eta_2 \nabla  \widetilde{\overline{F}}_u(\overline{\mathbf{w}}_u(\tau_s^{\mathsf{L}} \tau_s^{\mathsf{G}} k'-1))
- \frac{1}{U(s)} \sum_{\hat{u} \in \mathcal{U}(s)}\bigg(
\overline{\mathbf{w}}_{\hat{u}}(\tau_s^{\mathsf{L}} \tau_s^{\mathsf{G}} k'-1) 
- \eta_2 \nabla \widetilde{ \overline{F}}_{\hat{u}}(\overline{\mathbf{w}}_{\hat{u}}(\tau_s^{\mathsf{L}} \tau_s^{\mathsf{G}} k'-1))
\bigg)
\bigg\Vert^2 \bigg] \\
& \overset{(c)}{=}
2 \frac{1}{U(s)} \sum_{u \in \mathcal{U}(s)} 
\mathbb{E} \bigg[ \bigg\Vert  \overline{\mathbf{w}}_u(\tau_s^{\mathsf{L}} \tau_s^{\mathsf{G}} k'-1)
- \eta_2 \nabla  \widetilde{\overline{F}}_u(\overline{\mathbf{w}}_u(\tau_s^{\mathsf{L}} \tau_s^{\mathsf{G}} k'-1)) 
+ \eta_2 \nabla \overline{F}_u (\overline{\mathbf{w}}_u(\tau_s^{\mathsf{L}} \tau_s^{\mathsf{G}} k' -1))
- \eta_2 \nabla \overline{F}_u (\overline{\mathbf{w}}_u(\tau_s^{\mathsf{L}} \tau_s^{\mathsf{G}} k' -1))
\\
&
- \frac{1}{U(s)} \sum_{\hat{u} \in \mathcal{U}(s)}\bigg(
\overline{\mathbf{w}}_{\hat{u}}(\tau_s^{\mathsf{L}} \tau_s^{\mathsf{G}} k'-1) 
- \eta_2 \nabla \widetilde{ \overline{F}}_{\hat{u}}(\overline{\mathbf{w}}_{\hat{u}}(\tau_s^{\mathsf{L}} \tau_s^{\mathsf{G}} k'-1))
+ \eta_2 \nabla \overline{F}_{\hat{u}} (\overline{\mathbf{w}}_{\hat{u}}(\tau_s^{\mathsf{L}} \tau_s^{\mathsf{G}} k'-1))
- \eta_2 \nabla \overline{F}_{\hat{u}} (\overline{\mathbf{w}}_{\hat{u}}(\tau_s^{\mathsf{L}} \tau_s^{\mathsf{G}} k'-1))
\bigg)
\bigg\Vert^2 \bigg] \\
& \overset{(d)}{\leq} 
\underbrace{
8 \frac{1}{U(s)} \sum_{u \in \mathcal{U}(s)} 
\mathbb{E} \bigg[ \bigg\Vert 
\overline{\mathbf{w}}_u(\tau_s^{\mathsf{L}} \tau_s^{\mathsf{G}} k' - 1) - \frac{1}{U(s)} 
\sum_{\hat{u} \in \mathcal{U}(s)} \overline{\mathbf{w}}_{\hat{u}}(\tau_s^{\mathsf{L}} \tau_s^{\mathsf{G}} k' -1)
\bigg\Vert^2 \bigg]}_{(i)} \\
& +
\underbrace{
8 \eta_2^2 \frac{1}{U(s)} \sum_{u \in \mathcal{U}(s)} 
\mathbb{E} [ \Vert \nabla \overline{F}_u (\overline{\mathbf{w}}_u(\tau_s^{\mathsf{L}} \tau_s^{\mathsf{G}} k' -1)) 
- \nabla  \widetilde{\overline{F}}_u(\overline{\mathbf{w}}_u(\tau_s^{\mathsf{L}} \tau_s^{\mathsf{G}} k'-1)) \Vert^2 ]
}_{(ii)}
\\
& + 
\underbrace{
8 \eta_2^2 \frac{1}{U(s)} \sum_{u \in \mathcal{U}(s)} 
\mathbb{E} \bigg[ \bigg\Vert 
\frac{1}{U(s)} \sum_{\hat{u} \in \mathcal{U}(s)}\bigg(
\nabla \widetilde{ \overline{F}}_{\hat{u}}(\overline{\mathbf{w}}_{\hat{u}}(\tau_s^{\mathsf{L}} \tau_s^{\mathsf{G}} k'-1))
- \nabla \overline{F}_{\hat{u}} (\overline{\mathbf{w}}_{\hat{u}}(\tau_s^{\mathsf{L}} \tau_s^{\mathsf{G}} k'-1)) 
\bigg)
\bigg\Vert^2 \bigg]}_{(iii)} \\
& + 
\underbrace{
8 \eta_2^2 \frac{1}{U(s)} \sum_{u \in \mathcal{U}(s)} 
\mathbb{E} \bigg[ \bigg\Vert 
\frac{1}{U(s)} \sum_{\hat{u} \in \mathcal{U}(s)}\nabla \overline{F}_{\hat{u}} (\overline{\mathbf{w}}_{\hat{u}}(\tau_s^{\mathsf{L}} \tau_s^{\mathsf{G}} k'-1))
- \nabla \overline{F}_u (\overline{\mathbf{w}}_u(\tau_s^{\mathsf{L}} \tau_s^{\mathsf{G}} k' -1)) \bigg\Vert^2 \bigg]}_{(iv)},
\end{aligned}
\end{equation}

\noindent
where $(a)$ uses the definition of $\overline{\mathbf{w}}$ from~\eqref{eq:global_aggregation_rule}, $(b)$ is the global parameter update rule with $\widetilde{\overline{F}}_u$ using an approximation $\widetilde{F}$ of $F$ in~\eqref{eq:local_meta_function} (the same applies for $\widetilde{\overline{F}}$), $(c)$ introduces $\eta_2 \nabla \overline{F}_u(\overline{\mathbf{w}}_u)$ and $\eta_2 \nabla \overline{F}_{\hat{u}}(\overline{\mathbf{w}}_{\hat{u}})$, and $(d)$ follows from Cauchy-Schwarz inequality. Upper bounds for~\eqref{eq:prop2_eq7}(ii) and~\eqref{eq:prop2_eq7}(iii) follow immediately from Lemma~\ref{theory:lemma_sigma},  
and we bound~\eqref{eq:prop2_eq7}(iv) as follows:
\begin{align} \label{eq:prop2_eq8}
& 
8 \eta_2^2 \frac{1}{U(s)} \sum_{u \in \mathcal{U}(s)} 
\mathbb{E} \bigg[ \bigg\Vert 
\frac{1}{U(s)} \sum_{\hat{u} \in \mathcal{U}(s)}\nabla \overline{F}_{\hat{u}} (\overline{\mathbf{w}}_{\hat{u}}(\tau_s^{\mathsf{L}} \tau_s^{\mathsf{G}} k'-1))
- \nabla \overline{F}_u (\overline{\mathbf{w}}_u(\tau_s^{\mathsf{L}} \tau_s^{\mathsf{G}} k' -1)) 
\bigg\Vert^2 \bigg] \nonumber \\
& \overset{(a)}{=} 
8 \eta_2^2 \frac{1}{U(s)} \sum_{u \in \mathcal{U}(s)} 
\mathbb{E} \bigg[ \bigg\Vert 
\frac{1}{U(s)} \sum_{\hat{u} \in \mathcal{U}(s)}\bigg(
\nabla \overline{F}_{\hat{u}} (\overline{\mathbf{w}}_{\hat{u}}(\tau_s^{\mathsf{L}} \tau_s^{\mathsf{G}} k'-1))
- \nabla \overline{F}_{\hat{u}} (\overline{\mathbf{w}}(\tau_s^{\mathsf{L}} \tau_s^{\mathsf{G}} k'-1))
+ \nabla \overline{F}_{\hat{u}} (\overline{\mathbf{w}}(\tau_s^{\mathsf{L}} \tau_s^{\mathsf{G}} k'-1))
\bigg)
\nonumber \\
& 
- \nabla \overline{F}_u (\overline{\mathbf{w}}_u(\tau_s^{\mathsf{L}} \tau_s^{\mathsf{G}} k' -1))
+ \nabla \overline{F}_u (\overline{\mathbf{w}}(\tau_s^{\mathsf{L}} \tau_s^{\mathsf{G}} k' -1))
- \nabla \overline{F}_u (\overline{\mathbf{w}}(\tau_s^{\mathsf{L}} \tau_s^{\mathsf{G}} k' -1))
\bigg\Vert^2 \bigg] \nonumber \\
& \overset{(b)}{\leq}
24 \eta_2^2 \frac{1}{U(s)} \sum_{u \in \mathcal{U}(s)} 
\mathbb{E} \bigg[\bigg\Vert 
\frac{1}{U(s)} \sum_{\hat{u} \in \mathcal{U}(s)} \sum_{j \in \mathcal{W}_{\hat{u}}} \frac{\Delta_j(k' \tau_s^{\mathsf{G}})}{\Delta_{\hat{u}}(k' \tau_s^{\mathsf{G}})}
\bigg( \nabla F_j(\overline{\mathbf{w}}_{\hat{u}}(\tau_s^{\mathsf{L}}\tau_s^{\mathsf{G}}k'-1)) - \nabla F_j(\overline{\mathbf{w}}(\tau_s^{\mathsf{L}}\tau_s^{\mathsf{G}}k'-1)) \bigg)
\bigg\Vert^2 \bigg]
\nonumber \\
& +
24 \eta_2^2 \frac{1}{U(s)} \sum_{u \in \mathcal{U}(s)} 
\mathbb{E} \bigg[\bigg\Vert 
\sum_{j \in \mathcal{W}_{\hat{u}}} \frac{\Delta_j(k' \tau_s^{\mathsf{G}})}{\Delta_{\hat{u}}(k' \tau_s^{\mathsf{G}})} 
\bigg( \nabla F_j(\overline{\mathbf{w}}(\tau_s^{\mathsf{L}}\tau_s^{\mathsf{G}}k'-1)) - \nabla F_j(\overline{\mathbf{w}}_u(\tau_s^{\mathsf{L}}\tau_s^{\mathsf{G}}k'-1)) \bigg)
\bigg\Vert^2 \bigg] \nonumber \\
& +
24 \eta_2^2 \frac{1}{U(s)} \sum_{u \in \mathcal{U}(s)} 
\mathbb{E} [\Vert 
\nabla \overline{F} (\overline{\mathbf{w}}(\tau_s^{\mathsf{L}} \tau_s^{\mathsf{G}} k'-1)) - \nabla \overline{F}_u (\overline{\mathbf{w}}(\tau_s^{\mathsf{L}} \tau_s^{\mathsf{G}} k' -1))
\Vert^2 ] \nonumber \\
& \overset{(c)}{\leq}
24 \eta_2^2 \frac{1}{U(s)} \sum_{u \in \mathcal{U}(s)} 
\frac{1}{U(s)} \sum_{\hat{u} \in \mathcal{U}(s)}\sum_{j \in \mathcal{W}_{\hat{u}}} \frac{\Delta_j(k' \tau_s^{\mathsf{G}})}{\Delta_{\hat{u}}(k' \tau_s^{\mathsf{G}})}
\mathbb{E} \bigg[\bigg\Vert 
\nabla F_j(\overline{\mathbf{w}}_{\hat{u}}(\tau_s^{\mathsf{L}}\tau_s^{\mathsf{G}}k'-1)) - \nabla F_j(\overline{\mathbf{w}}(\tau_s^{\mathsf{L}}\tau_s^{\mathsf{G}}k'-1))
\bigg\Vert^2 \bigg]
\nonumber \\
& +
24 \eta_2^2 \frac{1}{U(s)} \sum_{u \in \mathcal{U}(s)} 
\sum_{j \in \mathcal{W}_{\hat{u}}} \frac{\Delta_j(k' \tau_s^{\mathsf{G}})}{\Delta_{\hat{u}}(k' \tau_s^{\mathsf{G}})} 
\mathbb{E} \bigg[\bigg\Vert 
\nabla F_j(\overline{\mathbf{w}}(\tau_s^{\mathsf{L}}\tau_s^{\mathsf{G}}k'-1)) - \nabla F_j(\overline{\mathbf{w}}_u(\tau_s^{\mathsf{L}}\tau_s^{\mathsf{G}}k'-1))
\bigg\Vert^2 \bigg] \nonumber \\
& +
24 \eta_2^2 \frac{1}{U(s)} \sum_{u \in \mathcal{U}(s)} 
\mathbb{E} [\Vert 
\nabla \overline{F} (\overline{\mathbf{w}}(\tau_s^{\mathsf{L}} \tau_s^{\mathsf{G}} k'-1)) - \nabla \overline{F}_u (\overline{\mathbf{w}}(\tau_s^{\mathsf{L}} \tau_s^{\mathsf{G}} k' -1))
\Vert^2 ]  \nonumber \\
& \overset{(d)}{\leq}
24 \eta_2^2 \mu_F^2 \frac{1}{U(s)} \sum_{u \in \mathcal{U}(s)} 
\frac{1}{U(s)} \sum_{\hat{u} \in \mathcal{U}(s)}\sum_{j \in \mathcal{W}_{\hat{u}}} \frac{\Delta_j(k' \tau_s^{\mathsf{G}})}{\Delta_{\hat{u}}(k' \tau_s^{\mathsf{G}})}
\mathbb{E} [\Vert 
\overline{\mathbf{w}}_{\hat{u}}(\tau_s^{\mathsf{L}}\tau_s^{\mathsf{G}}k'-1) - \overline{\mathbf{w}}(\tau_s^{\mathsf{L}}\tau_s^{\mathsf{G}}k'-1)
\Vert^2 ]
\nonumber \\
& +
24 \eta_2^2 \mu_F^2 \frac{1}{U(s)} \sum_{u \in \mathcal{U}(s)} 
\sum_{j \in \mathcal{W}_{\hat{u}}} \frac{\Delta_j(k' \tau_s^{\mathsf{G}})}{\Delta_{\hat{u}}(k' \tau_s^{\mathsf{G}})} 
\mathbb{E} [\Vert 
\overline{\mathbf{w}}(\tau_s^{\mathsf{L}}\tau_s^{\mathsf{G}}k'-1) - \overline{\mathbf{w}}_u(\tau_s^{\mathsf{L}}\tau_s^{\mathsf{G}}k'-1)
\Vert^2 ] \nonumber \\
& +
24 \eta_2^2 \frac{1}{U(s)} \sum_{u \in \mathcal{U}(s)} 
\mathbb{E} [\Vert 
\nabla \overline{F} (\overline{\mathbf{w}}(\tau_s^{\mathsf{L}} \tau_s^{\mathsf{G}} k'-1)) - \nabla \overline{F}_u (\overline{\mathbf{w}}(\tau_s^{\mathsf{L}} \tau_s^{\mathsf{G}} k' -1))
\Vert^2 ]  \nonumber \\
& \overset{(e)}{\leq}
48 \eta_2^2 \mu_F^2 \frac{1}{U(s)} \sum_{u \in \mathcal{U}(s)} 
\mathbb{E} [\Vert 
\overline{\mathbf{w}}_{\hat{u}}(\tau_s^{\mathsf{L}}\tau_s^{\mathsf{G}}k'-1) - \overline{\mathbf{w}}(\tau_s^{\mathsf{L}}\tau_s^{\mathsf{G}}k'-1)
\Vert^2 ]
+
24 \eta_2^2 \gamma_F^2 ,
\end{align}

\noindent
where $(a)$ introduces $\nabla \overline{F}_{\hat{u}}(\mathbf{w})$ and $\nabla \overline{F}_{u}(\mathbf{w})$ terms, $(b)$ uses the Cauchy-Schwarz followed by the swarm-wide meta-gradient definition from~\eqref{eq:local_meta_function}, $(c)$ applies Jensen's inequality to the expectations, $(d)$ recalls the $\mu_F$-smoothness property of $\nabla F_j$, and $(e)$ uses the result of Lemma~\ref{theory:lemma_gamma}. Combining~\eqref{eq:prop2_eq7} and~\eqref{eq:prop2_eq8} yields:
\begin{align} \label{eq:prop2_eq9}
& 2 \frac{1}{U(s)}\sum_{u \in \mathcal{U}(s)}
\mathbb{E}[\Vert \overline{\mathbf{w}}_u(\tau_s^{\mathsf{L}} \tau_s^{\mathsf{G}} k') - \overline{\mathbf{w}}(\tau_s^{\mathsf{L}} \tau_s^{\mathsf{G}} k') \Vert^2 ] \nonumber\\
& \overset{(a)}{\leq} 
8 \frac{1}{U(s)} \sum_{u \in \mathcal{U}(s)} 
\mathbb{E} \bigg[ \bigg\Vert 
\overline{\mathbf{w}}_u(\tau_s^{\mathsf{L}} \tau_s^{\mathsf{G}} k' - 1) - \frac{1}{U(s)} \sum_{\hat{u} \in \mathcal{U}(s)} \overline{\mathbf{w}}_{\hat{u}}(\tau_s^{\mathsf{L}} \tau_s^{\mathsf{G}} k' -1)
\bigg\Vert^2 \bigg] \nonumber \\
& +
8 \eta_2^2 \frac{1}{U(s)} \sum_{u \in \mathcal{U}(s)} 
\mathbb{E} [ \Vert \nabla \overline{F}_u (\overline{\mathbf{w}}_u(\tau_s^{\mathsf{L}} \tau_s^{\mathsf{G}} k' -1)) 
- \nabla  \widetilde{\overline{F}}_u(\overline{\mathbf{w}}_u(\tau_s^{\mathsf{L}} \tau_s^{\mathsf{G}} k'-1)) \Vert^2 ]
\nonumber \\
& + 
8 \eta_2^2 \frac{1}{U(s)} \sum_{u \in \mathcal{U}(s)} 
\mathbb{E} \bigg[ \bigg\Vert 
\frac{1}{U(s)} \sum_{\hat{u} \in \mathcal{U}(s)}\bigg(
\nabla \widetilde{ \overline{F}}_{\hat{u}}(\overline{\mathbf{w}}_{\hat{u}}(\tau_s^{\mathsf{L}} \tau_s^{\mathsf{G}} k'-1))
- \nabla \overline{F}_{\hat{u}} (\overline{\mathbf{w}}_{\hat{u}}(\tau_s^{\mathsf{L}} \tau_s^{\mathsf{G}} k'-1)) 
\bigg)
\bigg\Vert^2 \bigg] 
\nonumber \\
& + 
48 \eta_2^2 \mu_F^2 \frac{1}{U(s)} \sum_{u \in \mathcal{U}(s)} 
\mathbb{E} [\Vert 
\overline{\mathbf{w}}_{\hat{u}}(\tau_s^{\mathsf{L}}\tau_s^{\mathsf{G}}k'-1) - \overline{\mathbf{w}}(\tau_s^{\mathsf{L}}\tau_s^{\mathsf{G}}k'-1)
\Vert^2 ]
+
24 \eta_2^2 \gamma_F^2
\nonumber \\
& \overset{(b)}{\leq} 
(8+ 48 \eta_2^2 \mu_F^2 )
\frac{1}{U(s)} \sum_{u \in \mathcal{U}(s)} 
\mathbb{E} \bigg[ \bigg\Vert 
\overline{\mathbf{w}}_u(\tau_s^{\mathsf{L}} \tau_s^{\mathsf{G}} k' - 1) - \frac{1}{U(s)} \sum_{\hat{u} \in \mathcal{U}(s)} \overline{\mathbf{w}}_{\hat{u}}(\tau_s^{\mathsf{L}} \tau_s^{\mathsf{G}} k' -1)
\bigg\Vert^2 \bigg] 
+ 24 \eta_2^2 \gamma_F^2
\nonumber \\
& +
8 \eta_2^2 \frac{1}{U(s)} \sum_{u \in \mathcal{U}(s)} 
\mathbb{E} \bigg[ \bigg\Vert 
\sum_{j  \in \mathcal{W}_u}\frac{\Delta_j(k' \tau_s^{\mathsf{G}})}{\overline{\Delta}_u(k' \tau_s^{\mathsf{G}})} 
\bigg(
\nabla F_j (\overline{\mathbf{w}}_u(\tau_s^{\mathsf{L}} \tau_s^{\mathsf{G}} k' -1)) 
- \nabla \widetilde{ F}_j(\overline{\mathbf{w}}_u(\tau_s^{\mathsf{L}} \tau_s^{\mathsf{G}} k'-1)) 
\bigg) \bigg\Vert^2 \bigg] \nonumber \\
& + 
8 \eta_2^2 \frac{1}{U(s)} \sum_{u \in \mathcal{U}(s)} 
\mathbb{E} \bigg[ \bigg\Vert 
\frac{1}{U(s)} \sum_{\hat{u} \in \mathcal{U}(s)}\sum_{j  \in \mathcal{W}_u}\frac{\Delta_j(k' \tau_s^{\mathsf{G}})}{\overline{\Delta}_u(k' \tau_s^{\mathsf{G}})} \bigg(
\nabla \widetilde{F}_{j}(\overline{\mathbf{w}}_{\hat{u}}(\tau_s^{\mathsf{L}} \tau_s^{\mathsf{G}} k'-1))
- \nabla F_{j} (\overline{\mathbf{w}}_{\hat{u}}(\tau_s^{\mathsf{L}} \tau_s^{\mathsf{G}} k'-1)) 
\bigg)
\bigg\Vert^2 \bigg] 
\nonumber \\
& \overset{(c)}{\leq} 
(8 + 48 \eta_2^2 \mu_F^2 )
\frac{1}{U(s)} \sum_{u \in \mathcal{U}(s)} 
\mathbb{E} [ \Vert 
\overline{\mathbf{w}}_u(\tau_s^{\mathsf{L}} \tau_s^{\mathsf{G}} k' - 1) - \overline{\mathbf{w}}(\tau_s^{\mathsf{L}} \tau_s^{\mathsf{G}} k' -1)
\Vert^2 ] 
+ 24 \eta_2^2 \gamma_F^2
\nonumber \\
& +
16 \eta_2^2 \frac{1}{U(s)} \sum_{u \in \mathcal{U}(s)} \sum_{j  \in \mathcal{W}_u}\frac{\Delta_j(k' \tau_s^{\mathsf{G}})}{\overline{\Delta}_u(k' \tau_s^{\mathsf{G}})} 
(\sigma_F^j(\tau_s^{\mathsf{L}} \tau_s^{\mathsf{G}} k'-1))^2,
\end{align}

\noindent
where $(a)$ combines~\eqref{eq:prop2_eq7} and~\eqref{eq:prop2_eq8}, $(b)$ applies definition of $\nabla \overline{F}_u$ from~\eqref{eq:local_meta_function}, $(c)$ uses Jensen's inequality followed by Lemma~\ref{theory:lemma_sigma} and the definition of $\overline{\mathbf{w}}$ from~\eqref{eq:global_aggregation_rule}.
Solving~\eqref{eq:prop2_eq9} recursively yields:
\begin{align}
\label{eq:prop2_eq10}
& 2 \frac{1}{U(s)}\sum_{u \in \mathcal{U}(s)} \mathbb{E} [ \Vert \overline{\mathbf{w}}_u(\tau_s^{\mathsf{L}} \tau_s^{\mathsf{G}} k') - \overline{\mathbf{w}}(\tau_s^{\mathsf{L}} \tau_s^{\mathsf{G}} k') \Vert^2 ] 
\\
& \overset{(a)}{\leq}
16 \eta_2^2 \frac{1}{U(s)}\sum_{u \in \mathcal{U}(s)} \sum_{j  \in \mathcal{W}_u}\frac{\Delta_j(k' \tau_s^{\mathsf{G}}) }{\overline{\Delta}_u(k' \tau_s^{\mathsf{G}})}
\bigg(
\sum_{y=0}^{\tau_s^{\mathsf{L}}\tau_s^{\mathsf{G}}-1} (\sigma_{j}^{\mathsf{F}}(\tau_s^{\mathsf{L}} \tau_s^{\mathsf{G}} k'-1-y))^2 (8+ 48 \eta_2^2 \mu_F^2)^{y}
\bigg)
+
24 \eta_2^2 \gamma_F^2
\frac{1 - (8 + 48 \eta_2^2 \mu_F^2)^{\tau_s^{\mathsf{L}} \tau_s^{\mathsf{G}}}}{1- (8 + 48 \eta_2^2 \mu_F^2 )} \\
& \overset{(b)}{\leq}
\bigg(
16 \eta_2^2 \frac{1}{U(s)} \sum_{u \in \mathcal{U}(s)}  \sum_{j  \in \mathcal{W}_u}\frac{\Delta_j(k' \tau_s^{\mathsf{G}}) }{\overline{\Delta}_u(k' \tau_s^{\mathsf{G}})}
\sum_{y=0}^{\tau_s^{\mathsf{L}}\tau_s^{\mathsf{G}}-1}(\sigma_{j}^{\mathsf{F}}(\tau_s^{\mathsf{L}} \tau_s^{\mathsf{G}} k'-1-y))^2
+
24 \eta_2^2 \gamma_F^2
\bigg)
\frac{1 - (8 + 48 \eta_2^2 \mu_F^2)^{\tau_s^{\mathsf{L}} \tau_s^{\mathsf{G}}}}{1- (8 + 48 \eta_2^2 \mu_F^2)},
\end{align}
where $(a)$ is the result of recursion and the finite sum of geometric series, and $(b)$ uses $\sum_{i} a_i b_i \leq \sum_{i} a_i \sum_{i} b_i$ $\forall a_i, b_i \geq 0$. Combining the results of~\eqref{eq:prop2_eq6} and~\eqref{eq:prop2_eq10} in~\eqref{eq:prop2_eq2} yields:
\begin{equation}
\label{eq:prop2_eq14}
\begin{aligned}
& 2 \frac{1}{U(s)}\sum_{u \in \mathcal{U}(s)} \sum_{j  \in \mathcal{W}_u}\frac{\Delta_j(k' \tau_s^{\mathsf{G}}) }{\overline{\Delta}_u(k' \tau_s^{\mathsf{G}})}
\bigg( \mathbb{E}[\Vert \mathbf{w}_j(\tau_s^{\mathsf{L}} \tau_s^{\mathsf{G}} k') - \overline{\mathbf{w}}_u(\tau_s^{\mathsf{L}} \tau_s^{\mathsf{G}} k') \Vert^2 ] + \mathbb{E}[\Vert \overline{\mathbf{w}}_u(\tau_s^{\mathsf{L}} \tau_s^{\mathsf{G}} k') - \overline{\mathbf{w}}(\tau_s^{\mathsf{L}} \tau_s^{\mathsf{G}} k')
\Vert^2 ] \bigg) \\
& \leq 
\bigg(
16 \eta_2^2 \frac{1}{U(s)}\sum_{u \in \mathcal{U}(s)} \sum_{j  \in \mathcal{W}_u}\frac{\Delta_j(k' \tau_s^{\mathsf{G}}) }{\overline{\Delta}_u(k' \tau_s^{\mathsf{G}})}
\sum_{y=1}^{\tau_s^{\mathsf{L}}}
(\sigma_{j}^{\mathsf{F}}(\tau_s^{\mathsf{L}} \tau_s^{\mathsf{G}} k' - y))^2
+
24 \eta_2^2
\frac{1}{U(s)} \sum_{u \in \mathcal{U}(s)}(\gamma_{F}^{u})^2
\bigg)
\frac{1 - (8 + 48 \eta_2^2 \mu_F^2)^{\tau_s^{\mathsf{L}}}}{1- (8 + 48 \eta_2^2 \mu_F^2)}\\
& +
\bigg(
16 \eta_2^2 \frac{1}{U(s)}\sum_{u \in \mathcal{U}(s)} \sum_{j  \in \mathcal{W}_u}\frac{\Delta_j(k' \tau_s^{\mathsf{G}}) }{\overline{\Delta}_u(k' \tau_s^{\mathsf{G}})}
\sum_{y=1}^{\tau_s^{\mathsf{L}}\tau_s^{\mathsf{G}}} (\sigma_{j}^{\mathsf{F}}(\tau_s^{\mathsf{L}} \tau_s^{\mathsf{G}} k'-y))^2
+
24 \eta_2^2 \gamma_F^2
\bigg)
\frac{1 - (8 + 48 \eta_2^2 \mu_F^2)^{\tau_s^{\mathsf{L}} \tau_s^{\mathsf{G}}}}{1- (8 + 48 \eta_2^2 \mu_F^2)},
\end{aligned}
\end{equation}
with the summation shifted from $y=0$ to $y=1$.
\end{proof}

\newpage
\section{Proof of Theorem 1} \label{app:thm1}
\begin{proof}
First, from Assumption~\ref{eq:assumptions_all}, we see that $F_j(t)$, $F_u(t)$, and $F(t)$ are all $\mu_F$-smooth where $\mu_F = 4 \mu^{\mathsf{G}} + \eta_1 \mu^{\mathsf{H}} B$ (result follows immediately from~\cite{fallah2020personalized} upon using triangle inequality). 
In order to bound the first-order stationary point, $\mathbb{E}[ \Vert \nabla \overline{F}(\overline{\mathbf{w}}) \Vert^2]$, we start with 
\begin{equation} \label{eq:theorem1_eq1}
\begin{aligned}
& \overline{F}( \overline{\mathbf{w}}(\tau_s^{\mathsf{L}}\tau_s^{\mathsf{G}} k')) 
\overset{(a)}{\leq}
\overline{F}(\overline{\mathbf{w}}(\tau_s^{\mathsf{L}}\tau_s^{\mathsf{G}} k'-1)) + \nabla \overline{F}(\overline{\mathbf{w}}(\tau_s^{\mathsf{L}}\tau_s^{\mathsf{G}} k'-1))^{T}(\overline{\mathbf{w}}(\tau_s^{\mathsf{L}}\tau_s^{\mathsf{G}} k') - \overline{\mathbf{w}}(\tau_s^{\mathsf{L}}\tau_s^{\mathsf{G}} k'-1)) + \frac{\mu_F}{2} \Vert \overline{\mathbf{w}}(\tau_s^{\mathsf{L}}\tau_s^{\mathsf{G}} k') - \overline{\mathbf{w}}(\tau_s^{\mathsf{L}}\tau_s^{\mathsf{G}} k'-1)\Vert^2 \\
& \overset{(b)}{=}
\overline{F}(\overline{\mathbf{w}}(\tau_s^{\mathsf{L}}\tau_s^{\mathsf{G}} k'-1)) -\eta_2 \nabla \overline{F}(\overline{\mathbf{w}}(\tau_s^{\mathsf{L}}\tau_s^{\mathsf{G}} k'-1))^{T}
\bigg( \frac{1}{U(s)} \sum_{u \in \mathcal{U}(s)} \sum_{j \in \mathcal{W}_u} \frac{\Delta_j(k' \tau_s^{\mathsf{G}}) }{\overline{\Delta}_u(k' \tau_s^{\mathsf{G}})} 
\nabla \widetilde{F}_j(\mathbf{w}_j(\tau_s^{\mathsf{L}}\tau_s^{\mathsf{G}} k'-1))  \bigg) 
\\
& + \frac{\mu_F}{2} \bigg\Vert 
- \eta_2 \frac{1}{U(s)} \sum_{u \in \mathcal{U}(s)} \sum_{j \in \mathcal{W}_u} \frac{\Delta_j(k' \tau_s^{\mathsf{G}}) }{\overline{\Delta}_u(k' \tau_s^{\mathsf{G}})}
\nabla \widetilde{F}_j(\mathbf{w}_j(\tau_s^{\mathsf{L}}\tau_s^{\mathsf{G}} k'-1)) \bigg\Vert^2\\
& \overset{(c)}{= }
\overline{F}(\overline{\mathbf{w}}(\tau_s^{\mathsf{L}}\tau_s^{\mathsf{G}} k'-1))
- \eta_2 \nabla \overline{F}(\overline{\mathbf{w}}(\tau_s^{\mathsf{L}}\tau_s^{\mathsf{G}} k'-1))^{T}
\bigg( \frac{1}{U(s)} \sum_{u \in \mathcal{U}(s)} \sum_{j \in \mathcal{W}_u} \frac{\Delta_j(k' \tau_s^{\mathsf{G}}) }{\overline{\Delta}_u(k' \tau_s^{\mathsf{G}})}
\nabla \widetilde{F}_j(\mathbf{w}_j(\tau_s^{\mathsf{L}}\tau_s^{\mathsf{G}} k'-1))\bigg) \\
&
+ \eta_2^2 \frac{\mu_F}{2} \bigg\Vert \frac{1}{U(s)} \sum_{u \in \mathcal{U}(s)} \sum_{j \in \mathcal{W}_u} \frac{\Delta_j(k' \tau_s^{\mathsf{G}}) }{\overline{\Delta}_u(k' \tau_s^{\mathsf{G}})}
\nabla \widetilde{F}_j(\mathbf{w}_j(\tau_s^{\mathsf{L}}\tau_s^{\mathsf{G}} k'-1)) \bigg\Vert^2,
\end{aligned}
\end{equation}
where $(a)$ follows from the $\mu_F$-smoothness property of $F$, $F_u$, and $F_j$, $(b)$ uses the meta-gradient update procedure from~\eqref{eq:pfl_meta_gradient} and the definition of $\overline{\mathbf{w}}$ from~\eqref{eq:global_aggregation_rule}, and $(c)$ simplifies the algebra.
Taking the expectation of both sides of~\eqref{eq:theorem1_eq1} yields:
\begin{equation} \label{eq:theorem1_eq2}
\begin{aligned}
&\mathbb{E}[\overline{F}(\overline{\mathbf{w}}(\tau_s^{\mathsf{L}}\tau_s^{\mathsf{G}} k'))] - \mathbb{E}[\overline{F}(\overline{\mathbf{w}}(\tau_s^{\mathsf{L}}\tau_s^{\mathsf{G}} k'-1))] 
\leq \\
& \underbrace{- \eta_2 \mathbb{E} \bigg[ 
\nabla \overline{F}(\overline{\mathbf{w}}(\tau_s^{\mathsf{L}}\tau_s^{\mathsf{G}} k'-1))^{T}
\bigg( \frac{1}{U(s)} \sum_{u \in \mathcal{U}(s)} \sum_{j \in \mathcal{W}_u} \frac{\Delta_j(k' \tau_s^{\mathsf{G}})}{\overline{\Delta}_u(k' \tau_s^{\mathsf{G}})}  \nabla \widetilde{F}_j(\mathbf{w}_j(\tau_s^{\mathsf{L}}\tau_s^{\mathsf{G}} k'-1))\bigg) \bigg]}_{(i)} \\
& + \underbrace{ \eta_2^2 \frac{\mu_F}{2} \mathbb{E}\bigg[ \bigg\Vert 
\frac{1}{U(s)} \sum_{u \in \mathcal{U}(s)} \sum_{j \in \mathcal{W}_u} \frac{\Delta_j(k' \tau_s^{\mathsf{G}})}{\overline{\Delta}_u(k' \tau_s^{\mathsf{G}})} \nabla \widetilde{F}_j(\mathbf{w}_j(\tau_s^{\mathsf{L}}\tau_s^{\mathsf{G}} k'-1))
\bigg\Vert^2 \bigg]}_{(ii)} .
\end{aligned}
\end{equation}

\noindent
Before we bound~\eqref{eq:theorem1_eq2}(i) and~\eqref{eq:theorem1_eq2}(ii), we first analyze
\begin{equation}\label{eq:theorem1_eq3}
\begin{aligned}
& \frac{1}{U(s)} \sum_{u \in \mathcal{U}(s)} \sum_{j \in \mathcal{W}_u} \frac{\Delta_j(k' \tau_s^{\mathsf{G}})}{\overline{\Delta}_u(k' \tau_s^{\mathsf{G}})}  \nabla \widetilde{F}_j(\mathbf{w}_j(\tau_s^{\mathsf{L}}\tau_s^{\mathsf{G}} k'-1)) \\
& = 
X + Y + \frac{1}{U(s)} \sum_{u \in \mathcal{U}(s)}  \sum_{j \in \mathcal{W}_u} \frac{\Delta_j(k' \tau_s^{\mathsf{G}})}{\overline{\Delta}_u(k' \tau_s^{\mathsf{G}})} \nabla F_j(\overline{\mathbf{w}}(\tau_s^{\mathsf{L}}\tau_s^{\mathsf{G}} k'-1)),
\end{aligned}
\end{equation}
where:
\begin{equation}
\begin{aligned}
& X \triangleq \frac{1}{U(s)} \sum_{u \in \mathcal{U}(s)} \sum_{j \in \mathcal{W}_u} \frac{\Delta_j(k' \tau_s^{\mathsf{G}})}{\overline{\Delta}_u(k' \tau_s^{\mathsf{G}})} 
\bigg[\nabla \widetilde{F}_j(\mathbf{w}_j(\tau_s^{\mathsf{L}}\tau_s^{\mathsf{G}} k'-1)) - \nabla F_j(\mathbf{w}_j(\tau_s^{\mathsf{L}}\tau_s^{\mathsf{G}} k'-1)) \bigg]\\
& Y \triangleq \frac{1}{U(s)} \sum_{u \in \mathcal{U}(s)} \sum_{j \in \mathcal{W}_u} \frac{\Delta_j(k' \tau_s^{\mathsf{G}})}{\overline{\Delta}_u(k' \tau_s^{\mathsf{G}})} 
\bigg[\nabla F_j(\mathbf{w}_j(\tau_s^{\mathsf{L}}\tau_s^{\mathsf{G}} k'-1)) - \nabla F_j(\overline{\mathbf{w}}(\tau_s^{\mathsf{L}}\tau_s^{\mathsf{G}} k'-1)) \bigg].
\end{aligned}
\end{equation}

\noindent
We now bound $\Vert X \Vert^2$ as follows:
\begin{equation} \label{eq:theorem1_eq4}
\begin{aligned}
&\mathbb{E}[\Vert X \Vert^2] = \mathbb{E}\bigg[ \bigg\Vert
\frac{1}{U(s)} \sum_{u \in \mathcal{U}(s)} \sum_{j \in \mathcal{W}_u} \frac{\Delta_j(k' \tau_s^{\mathsf{G}})}{\overline{\Delta}_u(k' \tau_s^{\mathsf{G}})} 
\bigg( \nabla \widetilde{F}_j(\mathbf{w}_j(\tau_s^{\mathsf{L}}\tau_s^{\mathsf{G}} k'-1)) - \nabla F_j(\mathbf{w}_j(\tau_s^{\mathsf{L}}\tau_s^{\mathsf{G}} k'-1)) \bigg) \bigg\Vert^2 \bigg] \\
&\overset{(a)}{\leq}
\frac{1}{U(s)} \sum_{u \in \mathcal{U}(s)} \sum_{j \in \mathcal{W}_u} \frac{\Delta_j(k' \tau_s^{\mathsf{G}})}{\overline{\Delta}_u(k' \tau_s^{\mathsf{G}})} 
\mathbb{E}[ \Vert 
\nabla \widetilde{F}_j(\mathbf{w}_j(\tau_s^{\mathsf{L}}\tau_s^{\mathsf{G}} k'-1)) - \nabla F_j(\mathbf{w}_j(\tau_s^{\mathsf{L}}\tau_s^{\mathsf{G}} k'-1))
\Vert^2 ] \\
&\overset{(b)}{\leq}
\frac{1}{U(s)} \sum_{u \in \mathcal{U}(s)} \sum_{j \in \mathcal{W}_u} \frac{\Delta_j(k' \tau_s^{\mathsf{G}})}{\overline{\Delta}_u(k' \tau_s^{\mathsf{G}})} 
(\sigma_{j}^{\mathsf{F}}(\tau_s^{\mathsf{L}}\tau_s^{\mathsf{G}} k' -1))^2,
\end{aligned}
\end{equation}
using (a) Jensen's inequality, and (b) Lemma~\ref{theory:lemma_sigma}. Next, we bound $\Vert Y \Vert^2$ as follows:
\begin{equation} \label{eq:theorem1_eq5}
\begin{aligned}
&\mathbb{E}[\Vert Y \Vert^2] = \mathbb{E}\bigg[ \bigg\Vert
\frac{1}{U(s)} \sum_{u \in \mathcal{U}(s)} \sum_{j \in \mathcal{W}_u} \frac{\Delta_j(k' \tau_s^{\mathsf{G}})}{\overline{\Delta}_u(k' \tau_s^{\mathsf{G}})} 
\bigg( \nabla F_j(\mathbf{w}_j(\tau_s^{\mathsf{L}}\tau_s^{\mathsf{G}}k'-1)) - \nabla F_j(\overline{\mathbf{w}}(\tau_s^{\mathsf{L}}\tau_s^{\mathsf{G}}k'-1)) \bigg) 
\bigg\Vert^2 \bigg] \\
& \overset{(a)}{\leq}
\mu_F^2 \mathbb{E} \bigg[ \bigg\Vert 
\frac{1}{U(s)} \sum_{u \in \mathcal{U}(s)} \sum_{j \in \mathcal{W}_u} \frac{\Delta_j(k' \tau_s^{\mathsf{G}})}{\overline{\Delta}_u(k' \tau_s^{\mathsf{G}})} 
\bigg(
\mathbf{w}_j(\tau_s^{\mathsf{L}} \tau_s^{\mathsf{G}} k'-1) - \overline{\mathbf{w}}(\tau_s^{\mathsf{L}} \tau_s^{\mathsf{G}} k'-1)
\bigg)
\bigg\Vert^2 \bigg] 
\\
& \overset{(b)}{\leq}
\mu_F^2 \Upsilon(k'\tau_s^{\mathsf{L}}\tau_s^{\mathsf{G}}),
\end{aligned}
\end{equation}
where $(a)$ uses $\mu_F$-Lipschitz gradient property of $F_j$, and $(b)$ follows from Proposition~\ref{theory:prop_l2}, note that the recursion step which bounds $\mathbf{w}_j(t)-\overline{\mathbf{w}}(t)$ in Proposition~\ref{theory:prop_l2} also bounds $\mathbf{w}_j(t-1) - \overline{\mathbf{w}}(t-1)$. Using these results, we first analyze~\eqref{eq:theorem1_eq2}(i) as follows:
\begin{equation}
\label{eq:theorem1_eq6}
\begin{aligned}
& - \eta_2 \mathbb{E} \bigg[
\nabla \overline{F}(\overline{\mathbf{w}}(\tau_s^{\mathsf{L}}\tau_s^{\mathsf{G}}k'-1))^T 
\bigg(\frac{1}{U(s)}\sum_{u \in \mathcal{U}(s)} \sum_{j \in \mathcal{W}_u} \frac{\Delta_j(k' \tau_s^{\mathsf{G}})}{\overline{\Delta}_u(k' \tau_s^{\mathsf{G}})} 
\nabla \widetilde{F}_j(\mathbf{w}_j(\tau_s^{\mathsf{L}}\tau_s^{\mathsf{G}}k'-1)) 
\bigg) \bigg] \\
& \overset{(a)}{=}
- \eta_2 \mathbb{E} \bigg[\nabla \overline{F}(\overline{\mathbf{w}}(\tau_s^{\mathsf{L}}\tau_s^{\mathsf{G}}k'-1))^T 
\bigg( X + Y + \frac{1}{U(s)}\sum_{u \in \mathcal{U}(s)} \sum_{j \in \mathcal{W}_u} \frac{\Delta_j(k' \tau_s^{\mathsf{G}})}{\overline{\Delta}_u(k' \tau_s^{\mathsf{G}})} 
\nabla F_j(\overline{\mathbf{w}}(\tau_s^{\mathsf{L}}\tau_s^{\mathsf{G}}k' - 1))
\bigg) \bigg] \\
& \overset{(b)}{=}
- \eta_2 \mathbb{E} \bigg[\nabla \overline{F}(\overline{\mathbf{w}}(\tau_s^{\mathsf{L}}\tau_s^{\mathsf{G}}k'-1))^T 
\bigg( X + Y +
\nabla \overline{F}(\overline{\mathbf{w}}(\tau_s^{\mathsf{L}}\tau_s^{\mathsf{G}}k' - 1))
\bigg) \bigg] \\
& \overset{(c)}{=}
- \eta_2 
\bigg(
\mathbb{E} [\nabla \overline{F}(\overline{\mathbf{w}}(\tau_s^{\mathsf{L}}\tau_s^{\mathsf{G}}k'-1))^T 
(X + Y) ] + \mathbb{E} [\Vert \nabla \overline{F}(\overline{\mathbf{w}}(\tau_s^{\mathsf{L}}\tau_s^{\mathsf{G}}k'-1)) \Vert^2] 
\bigg)
\\
& \overset{(d)}{\leq}
\frac{-\eta_2 }{2} \mathbb{E}[ \Vert \nabla \overline{F}(\overline{\mathbf{w}}(\tau_s^{\mathsf{L}}\tau_s^{\mathsf{G}}k'-1)) \Vert^2 ] 
+ \eta_2 \bigg( \mathbb{E} [\Vert X \Vert^2] + \mathbb{E}[\Vert Y \Vert^2]
\bigg)
\\
& \overset{(e)}{\leq}
\frac{-\eta_2}{2} \mathbb{E}[ \Vert \nabla \overline{F}(\overline{\mathbf{w}}(\tau_s^{\mathsf{L}}\tau_s^{\mathsf{G}}k'-1)) \Vert^2 ] 
+ \eta_2
\bigg(
\frac{1}{U(s)} \sum_{u \in \mathcal{U}(s)} \sum_{j \in \mathcal{W}_u} \frac{\Delta_j(k' \tau_s^{\mathsf{G}})}{\overline{\Delta}_u(k' \tau_s^{\mathsf{G}})} 
(\sigma_{j}^{\mathsf{F}}(\tau_s^{\mathsf{L}}\tau_s^{\mathsf{G}} k' -1))^2
+ \mu_F^2 \Upsilon(k',\tau_s^{\mathsf{L}},\tau_s^{\mathsf{G}})
\bigg),
\end{aligned}
\end{equation}
where (a) comes from introducing $\nabla F_j(\mathbf{w}_j)$ and $\nabla F_j(\mathbf{w})$ terms, (b) is the definition of $\nabla \overline{F}$, (c) follows from linearity of expectation, (d) is due to
\begin{equation}
\label{eq:theorem1_eq7}
\begin{aligned}
& -\eta_2 \mathbb{E}[ \nabla \overline{F}(\overline{\mathbf{w}}(\tau_s^{\mathsf{L}}\tau_s^{\mathsf{G}}k'-1))^T(X+Y) ] 
\leq \frac{1}{2} \eta_2 \mathbb{E}[ \Vert \nabla \overline{F}(\overline{\mathbf{w}}(\tau_s^{\mathsf{L}}\tau_s^{\mathsf{G}}k'-1)) \Vert^2 ] + \frac{1}{2} \mathbb{E}[\Vert X+Y\Vert^2] \\
& \leq
\frac{1}{2} \eta_2 \mathbb{E}[ \Vert \nabla \overline{F}(\overline{\mathbf{w}}(\tau_s^{\mathsf{L}}\tau_s^{\mathsf{G}}k'-1)) \Vert^2 ] + \eta_2 \mathbb{E}[\Vert X \Vert^2] + \eta_2 \mathbb{E}[\Vert Y\Vert^2]
\end{aligned}
\end{equation}
(i.e., $(A+B)^2 \geq 0$ and Cauchy-Schwarz),
and (e) applies the results of~\eqref{eq:theorem1_eq4} and~\eqref{eq:theorem1_eq5}.
Next, we analyze~\eqref{eq:theorem1_eq2}(ii) as follows:
\begin{align} \label{eq:theorem1_eq9}
& 
\mathbb{E}\bigg[ \bigg\Vert 
\frac{1}{U(s)} \sum_{u \in \mathcal{U}(s)} \sum_{j \in \mathcal{W}_u} \frac{\Delta_j(k' \tau_s^{\mathsf{G}}) }{\overline{\Delta}_u(k' \tau_s^{\mathsf{G}})} 
\nabla \widetilde{F}_j(\mathbf{w}_j(\tau_s^{\mathsf{L}}\tau_s^{\mathsf{G}} k'-1))
\bigg\Vert^2 \bigg] \\
& \overset{(a)}{\leq}
\mathbb{E}\bigg[ \bigg\Vert 
X + Y + \frac{1}{U(s)} \sum_{u \in \mathcal{U}(s)} \sum_{j \in \mathcal{W}_u} \frac{\Delta_j(k' \tau_s^{\mathsf{G}}) }{\overline{\Delta}_u(k' \tau_s^{\mathsf{G}})} 
\nabla F_{j}(\overline{\mathbf{w}}(\tau_s^{\mathsf{L}}\tau_s^{\mathsf{G}} k'-1))
\bigg\Vert^2 \bigg] \nonumber \\
& \overset{(b)}{\leq}
3\mathbb{E}[ \Vert 
X \Vert^2]
+ 3\mathbb{E}[ \Vert
Y \Vert^2]
+ 3\mathbb{E} \bigg[ \bigg\Vert \frac{1}{U(s)} \sum_{u \in \mathcal{U}(s)} \sum_{j \in \mathcal{W}_u} \frac{\Delta_j(k' \tau_s^{\mathsf{G}}) }{\overline{\Delta}_u(k' \tau_s^{\mathsf{G}})} 
\nabla F_{j}(\overline{\mathbf{w}}(\tau_s^{\mathsf{L}}\tau_s^{\mathsf{G}} k'-1))
\bigg\Vert^2 \bigg] \nonumber \\
& \overset{(c)}{\leq}
3 \frac{1}{U(s)} \sum_{u \in \mathcal{U}(s)} \sum_{j \in \mathcal{W}_u} \frac{\Delta_j(k' \tau_s^{\mathsf{G}}) }{\overline{\Delta}_u(k' \tau_s^{\mathsf{G}})} 
(\sigma_{j}^{\mathsf{F}}(\tau_s^{\mathsf{L}}\tau_s^{\mathsf{G}} k' -1))^2 
+ 3 \mu_F^2 \Upsilon(k'\tau_s^{\mathsf{L}}\tau_s^{\mathsf{G}}) \nonumber \\
&
+ 3\mathbb{E} \bigg[ \bigg\Vert \frac{1}{U(s)} \sum_{u \in \mathcal{U}(s)} \sum_{j \in \mathcal{W}_u} \frac{\Delta_j(k' \tau_s^{\mathsf{G}}) }{\overline{\Delta}_u(k' \tau_s^{\mathsf{G}})} 
\nabla F_{j}(\overline{\mathbf{w}}(\tau_s^{\mathsf{L}}\tau_s^{\mathsf{G}} k'-1))
\bigg\Vert^2 \bigg] \nonumber \\
& \overset{(d)}{\leq}
3 \frac{1}{U(s)} \sum_{u \in \mathcal{U}(s)} \sum_{j \in \mathcal{W}_u} \frac{\Delta_j(k' \tau_s^{\mathsf{G}}) }{\overline{\Delta}_u(k' \tau_s^{\mathsf{G}})} 
(\sigma_{j}^{\mathsf{F}}(\tau_s^{\mathsf{L}}\tau_s^{\mathsf{G}} k' -1))^2 + 3 \mu_F^2 \Upsilon(k'\tau_s^{\mathsf{L}}\tau_s^{\mathsf{G}}) \nonumber \\
&
+ 3\mathbb{E} \bigg[ \bigg\Vert \frac{1}{U(s)} \sum_{u \in \mathcal{U}(s)} \sum_{j \in \mathcal{W}_u} \frac{\Delta_j(k' \tau_s^{\mathsf{G}}) }{\overline{\Delta}_u(k' \tau_s^{\mathsf{G}})} 
\bigg(\nabla F_{j}(\overline{\mathbf{w}}(\tau_s^{\mathsf{L}}\tau_s^{\mathsf{G}} k'-1)) - \nabla \overline{F}(\overline{\mathbf{w}}(\tau_s^{\mathsf{L}}\tau_s^{\mathsf{G}}k'-1)) +\nabla \overline{F}(\overline{\mathbf{w}}(\tau_s^{\mathsf{L}}\tau_s^{\mathsf{G}}k'-1)) \bigg)
\bigg\Vert^2 \bigg] \nonumber \\
& \overset{(e)}{\leq}
3 \frac{1}{U(s)} \sum_{u \in \mathcal{U}(s)} \sum_{j \in \mathcal{W}_u} \frac{\Delta_j(k' \tau_s^{\mathsf{G}}) }{\overline{\Delta}_u(k' \tau_s^{\mathsf{G}})} 
(\sigma_{j}^{\mathsf{F}}(\tau_s^{\mathsf{L}}\tau_s^{\mathsf{G}} k' -1))^2 + 3 \mu_F^2 \Upsilon(k'\tau_s^{\mathsf{L}}\tau_s^{\mathsf{G}}) \nonumber \\
&
+ 6\mathbb{E} \bigg[ \bigg\Vert \frac{1}{U(s)} \sum_{u \in \mathcal{U}(s)} \bigg( \sum_{j \in \mathcal{W}_u} \frac{\Delta_j(k' \tau_s^{\mathsf{G}}) }{\overline{\Delta}_u(k' \tau_s^{\mathsf{G}})} 
\nabla F_{j}(\overline{\mathbf{w}}(\tau_s^{\mathsf{L}}\tau_s^{\mathsf{G}} k'-1)) - \nabla F_u(\overline{\mathbf{w}}(\tau_s^{\mathsf{L}}\tau_s^{\mathsf{G}}k'-1)) \bigg) \bigg\Vert^2 \bigg]
+ 6 \mathbb{E} \bigg[ \bigg\Vert \nabla \overline{F}(\overline{\mathbf{w}}(\tau_s^{\mathsf{L}}\tau_s^{\mathsf{G}}k'-1))
\bigg\Vert^2 \bigg] \nonumber \\
& \overset{(f)}{\leq}
6\mathbb{E}[\Vert \nabla \overline{F}(\overline{\mathbf{w}}(\tau_s^{\mathsf{L}}\tau_s^{\mathsf{G}} k'-1)) \Vert^2 ] 
+ 3 \frac{1}{U(s)} \sum_{u \in \mathcal{U}(s)} \sum_{j \in \mathcal{W}_u} \frac{\Delta_j(k' \tau_s^{\mathsf{G}}) }{\overline{\Delta}_u(k' \tau_s^{\mathsf{G}})} 
(\sigma_{j}^{\mathsf{F}}(\tau_s^{\mathsf{L}}\tau_s^{\mathsf{G}} k' -1))^2 + 3 \mu_F^2 \Upsilon(k'\tau_s^{\mathsf{L}}\tau_s^{\mathsf{G}}) 
+ 6 \frac{1}{U(s)} \sum_{u \in \mathcal{U}(s)} (\gamma_F^u)^2 \nonumber, 
\end{align}
where, we (a) substitute the result from~\eqref{eq:theorem1_eq3}, (b) apply $(\sum_i^n a_i)^2 \leq n(\sum_i a_i^2)$, (c) combine the results of~\eqref{eq:theorem1_eq4} and~\eqref{eq:theorem1_eq5}, (d) introduce the global gradient $\nabla \overline{F}$, (e) recall that $\nabla \overline{F} = \frac{1}{U(s)} \sum_{u \in \mathcal{U}(s)} \nabla \overline{F}_u$ and use $(\sum_i^n a_i)^2 \leq n(\sum_i a_i^2)$, and (f) apply Jensen's inequality and Lemma~\ref{theory:lemma_gamma}.
Combining the results of~\eqref{eq:theorem1_eq6} and~\eqref{eq:theorem1_eq9} in~\eqref{eq:theorem1_eq2} yields:
\begin{equation} \label{eq:theorem1_eq10}
\begin{aligned}
&\mathbb{E}[\overline{F}(\overline{\mathbf{w}}(\tau_s^{\mathsf{L}}\tau_s^{\mathsf{G}} k'))] - \mathbb{E}[\overline{F}(\overline{\mathbf{w}}(\tau_s^{\mathsf{L}}\tau_s^{\mathsf{G}} k'-1))] 
\\
& \leq
- \frac{\eta_2 }{2}
\mathbb{E}[ \Vert \nabla \overline{F}(\overline{\mathbf{w}}(\tau_s^{\mathsf{L}}\tau_s^{\mathsf{G}}k'-1)) \Vert^2 ] 
+ \eta_2
\bigg(
\frac{1}{U(s)} \sum_{u \in \mathcal{U}(s)} \sum_{j \in \mathcal{W}_u} \frac{\Delta_j(k' \tau_s^{\mathsf{G}}) }{\overline{\Delta}_u(k' \tau_s^{\mathsf{G}})} 
(\sigma_{j}^{\mathsf{F}}(\tau_s^{\mathsf{L}}\tau_s^{\mathsf{G}} k' -1))^2
+ \mu_F^2 \Upsilon(k'\tau_s^{\mathsf{L}}\tau_s^{\mathsf{G}})
\bigg)
\\
& + \eta_2^2 \frac{\mu_F}{2} 
\bigg(6\mathbb{E}[\Vert \nabla \overline{F}(\overline{\mathbf{w}}(\tau_s^{\mathsf{L}}\tau_s^{\mathsf{G}} k'-1)) \Vert^2 ] 
+ 3 \frac{1}{U(s)} \sum_{u \in \mathcal{U}(s)} \sum_{j \in \mathcal{W}_u} \frac{\Delta_j(k' \tau_s^{\mathsf{G}}) }{\overline{\Delta}_u(k' \tau_s^{\mathsf{G}})} 
(\sigma_{j}^{\mathsf{F}}(\tau_s^{\mathsf{L}}\tau_s^{\mathsf{G}} k' -1))^2 + 3 \mu_F^2 \Upsilon(k'\tau_s^{\mathsf{L}}\tau_s^{\mathsf{G}}) 
+ 6 \frac{1}{U(s)} \sum_{u \in \mathcal{U}(s)} (\gamma_F^u)^2 \bigg).
\end{aligned}
\end{equation}

\noindent
With some algebra, we obtain:
\begin{equation} \label{eq:theorem1_eq11}
\begin{aligned}
& \mathbb{E}[\overline{F}(\overline{\mathbf{w}}(\tau_s^{\mathsf{L}}\tau_s^{\mathsf{G}} k'))] 
- \mathbb{E}[\overline{F}(\overline{\mathbf{w}}(\tau_s^{\mathsf{L}}\tau_s^{\mathsf{G}} k'-1))] 
\leq 
(6\eta_2^2 \frac{\mu_F}{2} - \frac{\eta_2}{2})\mathbb{E}[\Vert \nabla \overline{F}(\overline{\mathbf{w}}(\tau_s^{\mathsf{L}}\tau_s^{\mathsf{G}}k'-1)) \Vert^2] \\
&
+ (3\eta_2^2 \frac{\mu_F}{2} + \eta_2)
\bigg( 
\frac{1}{U(s)} \sum_{u \in \mathcal{U}(s)} \sum_{j \in \mathcal{W}_u} \frac{\Delta_j(k' \tau_s^{\mathsf{G}}) }{\overline{\Delta}_u(k' \tau_s^{\mathsf{G}})} 
(\sigma_{j}^{\mathsf{F}}(\tau_s^{\mathsf{L}}\tau_s^{\mathsf{G}} k' -1))^2 
+ \mu_F^2 \Upsilon(k'\tau_s^{\mathsf{L}}\tau_s^{\mathsf{G}}) 
\bigg)
+
3\eta_2^2 \mu_F \frac{1}{U(s)} \sum_{u \in \mathcal{U}(s)} (\gamma_F^u)^2 .
\end{aligned}
\end{equation}

\noindent
Since our goal from the beginning was to find an upper bound for the first-order stationary point, $\mathbb{E}[\Vert \nabla \overline{F}(\overline{\mathbf{w}}(t))\Vert^2]$, we now take the average of~\eqref{eq:theorem1_eq11} over all time $t$, which yields:
\begin{equation}
\label{eq:theorem1_eq12}
\begin{aligned}
& \frac{1}{\tau_s^{\mathsf{L}} \tau_s^{\mathsf{G}} K_s^{\mathsf{G}}} 
\sum_{k'=1}^{K_s^{\mathsf{G}}} \sum_{k = (k'-1) \tau_s^{\mathsf{G}} +1}^{\tau_s^{\mathsf{G}} k'} \sum_{t = (k-1) \tau_s^{\mathsf{L}} +1}^{\tau_s^{\mathsf{L}} k} 
\mathbb{E}[\overline{F}(\overline{\mathbf{w}}(t))] 
- \mathbb{E}[\overline{F}(\overline{\mathbf{w}}(t-1))] 
= \frac{1}{\tau_s^{\mathsf{L}} \tau_s^{\mathsf{G}} K_s^{\mathsf{G}}} (
\mathbb{E}[\overline{F}(\overline{\mathbf{w}}(\tau_s^{\mathsf{L}}\tau_s^{\mathsf{G}} K_s^{\mathsf{G}}))] 
- \mathbb{E}[\overline{F}(\overline{\mathbf{w}}(0))] )
\\
& 
\leq 
\frac{1}{\tau_s^{\mathsf{L}} \tau_s^{\mathsf{G}} K_s^{\mathsf{G}}} 
\sum_{k'=1}^{K_s^{\mathsf{G}}} \sum_{k = (k'-1)\tau_s^{\mathsf{G}} +1}^{\tau_s^{\mathsf{G}} k'} \sum_{t = (k-1) \tau_s^{\mathsf{L}} +1}^{\tau_s^{\mathsf{L}} k} 
\bigg[(6\eta_2^2 \frac{\mu_F}{2} - \frac{\eta_2}{2})\mathbb{E}[\Vert \nabla \overline{F}(\overline{\mathbf{w}}(t-1)) \Vert^2] \\
&
+ (3\eta_2^2 \frac{\mu_F}{2} + \eta_2)
\bigg( 
\frac{1}{U(s)} \sum_{u \in \mathcal{U}(s)} \sum_{j \in \mathcal{W}_u} \frac{\Delta_j(k' \tau_s^{\mathsf{G}}) }{\overline{\Delta}_u(k' \tau_s^{\mathsf{G}})} 
(\sigma_{j}^{\mathsf{F}}(t -1))^2 
+ \mu_F^2 \Upsilon(k'\tau_s^{\mathsf{L}}\tau_s^{\mathsf{G}}) 
\bigg)
+
3\eta_2^2 \mu_F \frac{1}{U(s)} \sum_{u \in \mathcal{U}(s)} (\gamma_F^u)^2 \bigg].
\end{aligned}
\end{equation}

\noindent 
We set $\eta_2 < \frac{1}{6\mu_F}$ and obtain:
\begin{align} \label{eq:theorem1_eq13}
&-\frac{6\eta_2^2 \frac{\mu_F}{2} - \frac{\eta_2}{2}}{\tau_s^{\mathsf{L}} \tau_s^{\mathsf{G}} K_s^{\mathsf{G}}}
\sum_{k'=1}^{K_s^{\mathsf{G}}} \sum_{k = \tau_s^{\mathsf{G}} (k'-1)+1}^{\tau_s^{\mathsf{G}} k'} \sum_{t = \tau_s^{\mathsf{L}} (k-1)+1}^{\tau_s^{\mathsf{L}} k} 
\mathbb{E}[\Vert \nabla \overline{F}(\overline{\mathbf{w}}(t-1)) \Vert^2] \\
& \leq
\frac{1}{\tau_s^{\mathsf{L}} \tau_s^{\mathsf{G}} K_s^{\mathsf{G}}} \bigg(
\mathbb{E}[\overline{F}(\overline{\mathbf{w}}(0))]  
- \mathbb{E}[\overline{F}(\overline{\mathbf{w}}(\tau_s^{\mathsf{L}}\tau_s^{\mathsf{G}} K_s^{\mathsf{G}}))] 
+ 
\sum_{k'=1}^{K_s^{\mathsf{G}}} \sum_{k = \tau_s^{\mathsf{G}} (k'-1)+1}^{\tau_s^{\mathsf{G}} k'} \sum_{t = \tau_s^{\mathsf{L}} (k-1)+1}^{\tau_s^{\mathsf{L}} k} \nonumber \\
&
\bigg[
(3\eta_2^2 \frac{\mu_F}{2} + \eta_2)
\bigg( 
\frac{1}{U(s)} \sum_{u \in \mathcal{U}(s)} \sum_{j \in \mathcal{W}_u} \frac{\Delta_j(k' \tau_s^{\mathsf{G}}) }{\overline{\Delta}_u(k' \tau_s^{\mathsf{G}})} 
(\sigma_{j}^{\mathsf{F}}(t -1))^2 
+ \mu_F^2 \Upsilon(k',\tau_s^{\mathsf{L}},\tau_s^{\mathsf{G}}) 
\bigg)
+
3\eta_2^2 \mu_F \frac{1}{U(s)} \sum_{u \in \mathcal{U}(s)} (\gamma_F^u)^2
\bigg] \bigg). \nonumber
\end{align}

Finally, noting that $\mathbb{E}[\overline{F}(\overline{\mathbf{w}}(0))] = \overline{F}(\overline{\mathbf{w}}(0))$ and that $-\mathbb{E}[\overline{F}(\overline{\mathbf{w}}(\tau_s^{\mathsf{L}} \tau_s^{\mathsf{G}} K_s^{\mathsf{G}}))] \leq -F^*$, we have the result:
\begin{equation} \label{eq:theorem1_eq14}
\begin{aligned}
&
\frac{1}{\tau_s^{\mathsf{L}} \tau_s^{\mathsf{G}} K_s^{\mathsf{G}}}
\sum_{k'=1}^{K_s^{\mathsf{G}}} \sum_{k = (k'-1) \tau_s^{\mathsf{G}} +1}^{\tau_s^{\mathsf{G}} k'} \sum_{t = (k-1)\tau_s^{\mathsf{L}} +1}^{\tau_s^{\mathsf{L}} k} 
\mathbb{E}[\Vert \nabla \overline{F}(\overline{\mathbf{w}}(t-1)) \Vert^2] 
\leq 
\frac{1}{\frac{\eta_2}{2} - 6\eta_2^2 \frac{\mu_F}{2}}
\frac{1}{\tau_s^{\mathsf{L}} \tau_s^{\mathsf{G}} K_s^{\mathsf{G}}}
\bigg[
\overline{F}(\overline{\mathbf{w}}(0)) - F^*
+ 
\sum_{k'=1}^{K_s^{\mathsf{G}}} \sum_{k = (k'-1)\tau_s^{\mathsf{G}} +1}^{\tau_s^{\mathsf{G}} k'} \\
&
\sum_{t = (k-1)\tau_s^{\mathsf{L}} +1}^{\tau_s^{\mathsf{L}} k} 
\bigg[
(3\eta_2^2 \frac{\mu_F}{2} + \eta_2)
\bigg( 
\frac{1}{U(s)} \sum_{u \in \mathcal{U}(s)} \sum_{j \in \mathcal{W}_u} \frac{\Delta_j(k' \tau_s^{\mathsf{G}}) }{\overline{\Delta}_u(k' \tau_s^{\mathsf{G}})} 
(\sigma_{j}^{\mathsf{F}}(t -1))^2 
+ \mu_F^2 \Upsilon(k'\tau_s^{\mathsf{L}}\tau_s^{\mathsf{G}}) 
\bigg)
+
3\eta_2^2 \mu_F \frac{1}{U(s)} \sum_{u \in \mathcal{U}(s)} (\gamma_F^u)^2
\bigg]
\bigg].
\end{aligned}
\end{equation}
\end{proof}

\newpage
\section{Proof of Lemma 3} \label{app:l3_proof}
Expanding the left hand side of~\eqref{eq:cluster_grad_bound} using the definitions of $\nabla \overline{F}$ and $\nabla \overline{F}^{\mathsf{D}}$, and upper bounding it using the Jensen's inequality yields $\frac{1}{U(s)} \sum_{u \in \mathcal{U}(s)} \mathbb{E}\big[\Vert
\nabla \overline{F}_u(\mathbf{w}(t)) - \nabla F^{\mathsf{D}}_{C(u,s)}(\mathbf{w}(t))
\Vert^2\big] $, where we substitute $C(u,s)$ for $c$ as the coupling of active UAV swarm to actively trained device cluster allows us to apply~\eqref{eq:local_meta_function} onto $\nabla \overline{F}_u$ and obtain: $\frac{1}{U(s)} \sum_{u \in \mathcal{U}(s)} \sum_{j \in \mathcal{W}_u} \frac{\Delta_j(k)}{\overline{\Delta}_u(k)}\mathbb{E}
\big[\Vert \nabla F_j(\mathbf{w}(t)) - \nabla F^{\mathsf{D}}_{C(u,s)}(\mathbf{w}(t))
\Vert^2\big] $ by applying of Jensen's inequality.
Similar techniques used in Lemma~\ref{theory:lemma_sigma} then yield the result.

\section{An Overview of Geometric Programming} \label{app:gp}
A prerequisite to geometric programming (GP) is the notion of \textit{monomials} and \textit{posynomials}, which we provide below. 
         \begin{definition}
         A {{monomial}} is a function $f: \mathbb{R}^n_{++}\rightarrow \mathbb{R}$: $f(\bm{y})=d y_1^{a_1} y_2^{a_2} \cdots y_n ^{a_n}$, with $d\geq 0$, $\bm{y}=[y_1,\cdots,y_n]$, and $a_j\in \mathbb{R}$, $\forall j$, where $\mathbb{R}^n_{++}$ denotes the strictly positive quadrant of $n$-dimensional Euclidean space. Also, a {posynomial} $g$ is a sum of monomials: $g(\bm{y})= \sum_{m=1}^{M} d_m y_1^{\alpha^{(1)}_m}  \cdots y_n ^{\alpha^{(n)}_m}$.
\end{definition}
A standard GP is a non-convex
problem formulated as minimizing a posynomial under posynomial 
inequality constraints and monomial equality constraints~\cite{chiang2007power}:
        \begin{equation}\label{eq:GPformat}
            \begin{aligned}
        &\min_{\bm{y}} g_0 (\bm{y})~~\\
          & \textrm{s.t.} ~~\\ &g_i(\bm{y})\leq 1, \; i=1,\cdots,I,\\& f_\ell(\bm{y})=1, \; \ell=1,\cdots,L,
            \end{aligned}
        \end{equation}
        where  $g_i(\bm{y})=\sum_{m=1}^{M_i} d_{i,m} y_1^{a^{(1)}_{i,m}} \cdots y_n ^{a^{(n)}_{i,m}}$, $\forall i$, and $f_\ell(\bm{y})= d_\ell y_1^{a^{(1)}_\ell}  \cdots y_n ^{a^{(n)}_\ell}$, $\forall \ell$. Since the log-sum-exp function $f(\bm{y}) = \log \sum_{j=1}^n e^{y_j}$ is convex, where $\log$ denotes the natural logarithm, with logarithmic change of variables and constants $z_i=\log(y_i)$, $b_{i,k}=\log(d_{i,k})$, $b_\ell=\log (d_\ell)$, and applying the $\log$ on the objective and constrains of~\eqref{eq:GPformat}, the GP in its standard format can be transformed to the following convex programming formulation:
          \begin{equation}~\label{GPtoConvex}
            \begin{aligned}
            &\min_{\bm{z}} \;\log \sum_{m=1}^{M_0} e^{\left(\bm{a}^{\top}_{0,m}\bm{z}+ b_{0,m}\right)}\\&\textrm{s.t.}~ \log \sum_{m=1}^{M_i} e^{\left(\bm{a}^{\top}_{i,m}\bm{z}+ b_{i,m}\right)}\leq 0,~ i=1,\cdots,I, \\&~~~~~~ \bm{a}_\ell^\top \bm{z}+b_\ell =0,\; \ell=1,\cdots,L,
            \end{aligned}
        \end{equation}
        where $\bm{z}=[z_1,\cdots,z_n]^\top$, $\bm{a}_{i,m}=\left[a_{i,m}^{(1)},\cdots, a_{i,m}^{(n)}\right]^\top$, $\forall i,m$, and $\bm{a}_{\ell}=\left[a_{\ell}^{(1)},\cdots, a_{\ell}^{(n)}\right]^\top$\hspace{-2mm}, $\forall \ell$.
        
        As can be seen, in $\bm{\mathcal{P}}$, there are multiple terms in the objective function (in the upper bound of convergence of the ML model) that are in the format of ratio between two posynomials, which are not posynomial. We thus aim to transform a ratio of two posynomials to a ratio between a posynomial (in the numerator) and a monomial (in the denominator). Given the fact that the ratio between a posynomial and a monomial is a posynomial, we then aim to transform the problem to the standard GP format. To carry out this transformation, we exploit arithmetic-geometric mean inequality which lower bounds a posynomial with a monomial. 
\begin{lemma}[\textbf{Arithmetic-geometric mean inequality}~\cite{chiang2007power}]\label{Lemma:ArethmaticGeometric}
         A posynomial $g(\bm{x})=\sum_{k=1}^{K} u_k(\bm{x})$, where $u_k(\bm{x})$ is a monomial, $\forall k$, can be lower-bounded via a monomial as follows:
         \begin{equation}\label{eq:approxPosMon}
             g(\bm{x})\geq \hat{g}(\bm{x})\triangleq \prod_{k=1}^{K}\left( \frac{u_k(\bm{x})}{a_k(\bm{y})}\right)^{a_k(\bm{y})},
         \end{equation}
         where $a_k(\bm{y})=u_k(\bm{y})/g(\bm{y})$, $\forall k$, and $\bm{y}>0$ is a fixed point.
\end{lemma}

\newpage
\section{Proof of Proposition~\ref{prop:optOpt}}\label{app:optOpt}
Let us first rewrite problem~$(\bm{\mathcal{P}})$ in its equivalent form as follows:
\begin{align}\label{eq:EquiP1}
   & (\bm{{\mathcal{P}}}): \min_{\boldsymbol{\rho},\boldsymbol{\varrho},\boldsymbol{\alpha},\boldsymbol{g},\Omega} ~~~\Omega~~~~~~~~~~~\\
&\textrm{s.t.}~~\\ &\eqref{eq:f1_con1}-\eqref{eq:f1_con2},\\&
 ~(1-\theta)(a)+\theta{(b)} \leq \Omega,~ \Omega \geq 0,
    \end{align}
    where $\Omega\in \mathbb{R}^+$ is an auxiliary variable used to move the objective function into the constraints, and $(a)$, and $(b)$ are those introduced in~\eqref{eq:f1_obj}.
The corresponding approximated problem can also be expressed as follows:
  \begin{align}\label{eq:ApproxP1}
   & (\bm{\widehat{\mathcal{P}}}_m): \min_{\boldsymbol{\rho},\boldsymbol{\varrho},\boldsymbol{\alpha},\boldsymbol{g},\Omega} ~~~\Omega~~~~~~~~~~~
\textrm{s.t.}~~ \eqref{eq:f1_con1}-\eqref{eq:f1_con2},
 ~(1-\theta) \widetilde{(a)}_m+\theta (b) \leq \Omega,~ \Omega \geq 0,
    \end{align}
where $\theta\widetilde{(a)}$ follows from the procedure outlined to obtain $\widetilde{\bm{\mathcal{P}}}_m$  in the main text, i.e., applying the posynomial condensation technique. It is easy to verify that the solution of $\widetilde{\bm{\mathcal{P}}}_m$ coincides with that of $\widehat{\bm{\mathcal{P}}}_m$.  Thus to prove the proposition, it is enough to prove that solving $\widehat{\bm{\mathcal{P}}}_m$  generates a sequence of improved
feasible solutions for problem $\bm{\widetilde{\mathcal{P}}}_m$ that converge to a point $\bm{x}^\star$
satisfying the Karush-Kuhn-Tucker (KKT) conditions of $\bm{\mathcal{P}}$. Note that $\widehat{\bm{\mathcal{P}}}_m$ can be solved using the procedure Algorithm~\ref{alg:cent}.


Following the justifications in \textbf{Observation 2} in Sec.~\ref{ssec:gp-solve},  the constraints of $\widehat{\bm{\mathcal{P}}}_m$   are separable with respect to each individual UAV swam. Thus, the performance of the distributed algorithm proposed to solve $\widehat{\bm{\mathcal{P}}}_m$  distributedly at each UAV swarm coincides with that of the centralized one for a fixed set of estimated parameters. 
Under the approximations described in~\eqref{eq:delta_u_approx} and~\eqref{eq:D_j_approx}, 
the algorithm in fact solves an~\textit{inner approximation} of problem $\bm{\mathcal{P}}$~\cite{GeneralInnerApp}. Hence, it is sufficient to prove the following three conditions for the sequence of generated solutions by the algorithm~\cite{GeneralInnerApp}:
\\

\begin{enumerate}
    \item
 \textit{The approximations used in problem $\bm{\widehat{\mathcal{P}}}_m$ should tighten the constraints of problem ${\bm{\mathcal{P}}}$:} Since the constraints~\eqref{eq:f1_con1}-\eqref{eq:f1_con2} are common to both Problems $(\bm{\mathcal{P}})$ and $\bm{\widehat{\mathcal{P}}}_m$, it is enough to show that $(1-\theta) (a) +\theta{(b)} \leq (1-\theta) \widetilde{(a)} + \theta (b)$ for the approximated constraint, assuming some solution $\bm{x}[m]$ ($\bm{x}$ is the solution vector defined in Algorithm~\ref{alg:cent}). Equivalently, it is sufficient to show that $(a) \leq  \widetilde{(a)}$ for $\bm{x}[m]$.

To show this, it is sufficient to show (i) $\Xi(s) \leq \widetilde{\Xi(s)}$ , and (ii) $\widehat{\Xi}(k)\leq \widetilde{\widehat{\Xi}(k)}$, $\forall k$, where $\Xi(s),\widehat{\Xi}(k)$ are the two terms in $(a)$ (see~\eqref{eq:f1_obj}) and $\widetilde{\Xi(s)}$ and $ \widetilde{\widehat{\Xi}(k)}$ are their respective approximations obtained via~\eqref{eq:delta_u_approx} and~\eqref{eq:D_j_approx}. 
Condition (i) holds since $\overline{\Delta}_u(k) \geq \widehat{\overline{\Delta}}_u(k)$, and condition (ii) also holds since $D_j(k) \geq  \widehat{D}_j(k)$ $\forall k$, as in equations~\eqref{eq:delta_u_approx} and~\eqref{eq:D_j_approx}.\footnote{Note that we  replace $\overline{\Delta}_u(k)$ and $D_j(k)$ with their approximated versions in~\eqref{eq:delta_u_approx} and~\eqref{eq:D_j_approx} only when they appear in the denominator of the terms in (a). Thus lower bounding these terms result in upper bounding (a) as desired.}
\\

    \item  \textit{Upon convergence, the value of each approximated constraint in problem $\bm{\widehat{\mathcal{P}}}_m$ should coincide with that of the corresponding original constraint in $\bm{{\mathcal{P}}}$:}  Since the constraints~\eqref{eq:f1_con1}-\eqref{eq:f1_con2} are common in problems $(\bm{\mathcal{P}})$ and $(\widetilde{\bm{\mathcal{P}}})$, we need to show $(1-\theta) (a) +\theta{(b)} = (1-\theta) \widetilde{(a)} + \theta (b)$ for the approximated constraint upon convergence. Note that when the algorithm converges we have 
        \begin{equation}\label{eq:equiDelt}
\widetilde{\delta}_{j,i,n}(k) =\widetilde{\delta}_{j,i,n}(k;m),
~~\delta_{j,h,i,n}(k) =\delta_{j,h,i,n}(k;m),~~n \in \{1,2,3\},
\end{equation}
where $\widetilde{\delta}_{j,i,n}(k;m)$ and $\delta_{j,h,i,n}(k;m)$  are the approximations used in~\eqref{eq:delta_u_approx} 
(see~\eqref{eq:little_delta_defs} for the definition of $\widetilde{\delta}_{j,i,n}(k)$ and $\delta_{j,h,i,n}(k)$). Also, upon convergence, we have:
\begin{equation}\label{eq:equiLamb}
    \lambda_{i,j}(k) =\lambda_{i,j}(k;m) ,~~ \widetilde{\lambda}_{h,i,j}(k) =\widetilde{\lambda}_{h,i,j}(k;m),
\end{equation}
where $\lambda_{i,j}(k;m),\widetilde{\lambda}_{h,i,j}(k;m)$ are the approximations used in~\eqref{eq:D_j_approx} 
(see~\eqref{eq:little_lambda_defs} for the definition of $\lambda_{i,j}(k),\widetilde{\lambda}_{h,i,j}(k)$).

    Considering the terms inside (a), we need to show that upon convergence (i) $\Xi(s) = \widetilde{\Xi(s)}$, 
    and (ii) $\widehat{\Xi}(k)= \widetilde{\widehat{\Xi}(k)}$. 
    To demonstrate that (i) holds, it is sufficient to show that $\overline{\Delta}_u(k) = \widehat{\overline{\Delta}}_u(k)$,  
    and $D_j(k) = \widehat{D}_j(k)$, upon convergence,. 
    In the following, we demonstrate that $\overline{\Delta}_u(k) = \widehat{\Delta}_u(k)$ upon convergence: 
    
    {\small
    \begin{align}
    &\widehat{\overline{\Delta}}_u(k) \Bigg |_{\eqref{eq:equiDelt},\eqref{eq:equiLamb}}
    = 
    \prod_{j\in \mathcal{W}_u} \Bigg[
    \prod_{i \in \mathcal{C}(u,s)} 
    \bigg[
    \bigg(\frac{ \widetilde{\delta}_{i,j,1}(k;m)
    \left[ \overline{\Delta}_u(k;m) \right] } 
    { \widetilde{\delta}_{i,j,1}(k;m)}\bigg)^{\frac{ \widetilde{\delta}_{i,j,1}(k;m) } {\overline{\Delta}_u(k;m) } }
    \nonumber \\
    & \times 
    \bigg(\frac{ \widetilde{\delta}_{i,j,2}(k;m)
    \left[ \overline{\Delta}_u(k;m) \right] } 
    { \widetilde{\delta}_{i,j,2}(k;m)}\bigg)^{\frac{ \widetilde{\delta}_{i,j,2}(k;m) } {\overline{\Delta}_u(k;m) } }
    \times
    \bigg(\frac{ \widetilde{\delta}_{i,j,3}(k;m)
    \left[ \overline{\Delta}_u(k;m) \right] } 
    { \widetilde{\delta}_{i,j,3}(k;m)}\bigg)^{\frac{ \widetilde{\delta}_{i,j,3}(k;m) } {\overline{\Delta}_u(k;m) } } \bigg] \nonumber \\
    & \times 
    \prod_{h \in \widehat{\mathcal{W}}_u} 
    \prod_{i \in \mathcal{C}(u,s)}
    \bigg[
    \bigg( \frac{ \delta_{j,h,i,1}(k;m)
    \left[ \overline{\Delta}_u(k;m) \right] }
    { \delta_{j,h,i,1}(k;m) } \bigg)^{ \frac{ \delta_{j,h,i,1}(k;m) } {\overline{\Delta}_u(k;m) } } \nonumber \\
    & \bigg( \frac{ \delta_{j,h,i,2}(k;m)
    \left[ \overline{\Delta}_u(k;m) \right] }
    { \delta_{j,h,i,2}(k;m) } \bigg)^{ \frac{ \delta_{j,h,i,2}(k;m) } {\overline{\Delta}_u(k;m) } } 
    \times     
    \bigg( \frac{ \delta_{j,h,i,3}(k;m)
    \left[ \overline{\Delta}_u(k;m) \right] }
    { \delta_{j,h,i,3}(k;m ) } \bigg)^{ \frac{ \delta_{j,h,i,3}(k;m) } {\overline{\Delta}_u(k;m) } } 
    \bigg]
    \Bigg] \nonumber \\
    &=
    \prod_{j\in \mathcal{W}_u} \Bigg[
    \prod_{i \in \mathcal{C}(u,s)} 
    \bigg[
    \bigg({
    \left[ \overline{\Delta}_u(k;m) \right] } 
    \bigg)^{\frac{ \widetilde{\delta}_{i,j,1}(k;m) } {\overline{\Delta}_u(k;m) } }
    \nonumber \\
    & \times 
    \bigg({
    \left[ \overline{\Delta}_u(k;m) \right] } 
    \bigg)^{\frac{ \widetilde{\delta}_{i,j,2}(k;m) } {\overline{\Delta}_u(k;m) } }
    \times
    \bigg({ 
    \left[ \overline{\Delta}_u(k;m) \right] } 
    \bigg)^{\frac{ \widetilde{\delta}_{i,j,3}(k;m) } {\overline{\Delta}_u(k;m) } } \bigg] \nonumber \\ 
    & \times 
    \prod_{h \in \widehat{\mathcal{W}}_u} 
    \prod_{i \in \mathcal{C}(u,s)}
    \bigg[
    \bigg( {
    \left[ \overline{\Delta}_u(k;m ) \right] }
    \bigg)^{ \frac{ \delta_{j,h,i,1}(k;m) } {\overline{\Delta}_u(k;m) } }  \bigg( {
    \left[ \overline{\Delta}_u(k;m ) \right] }
     \bigg)^{ \frac{ \delta_{j,h,i,2}(k;m) } {\overline{\Delta}_u(k;m) } } \\
    &
    \times     
    \bigg(
    {
    \left[ \overline{\Delta}_u(k;m ) \right] } \bigg)^{ \frac{ \delta_{j,h,i,3}(k;m ) } {\overline{\Delta}_u(k;m ) } } 
    \bigg]
    \Bigg]
    \nonumber \\
    &
    =  \overline{\Delta}_u(k)^{\left(\frac{ \sum_{j\in \mathcal{W}_u} 
    \sum_{i \in \mathcal{C}(u,s)} \widetilde{\delta}_{i,h,1}(k;m) + \widetilde{\delta}_{i,h,2}(k;m) + \widetilde{\delta}_{i,h,3}(k;m) } {\overline{\Delta}_u(k;m) } \right)}
    \nonumber \\
    &   \times  
    \overline{\Delta}_u(k)^{\left(\frac{ \sum_{j\in \mathcal{W}_u}  
    \sum_{h \in \widehat{\mathcal{W}}_u} 
    \sum_{i \in \mathcal{C}(u,s)} \delta_{j,h,i,1}(k;m) + \delta_{j,h,i,2}(k;m ) + \delta_{j,h,i,3}(k;m )  } {\overline{\Delta}_u(k;m ) } \right)}
    \nonumber \\
    & = \overline{\Delta}_u(k)^
    {\left(\frac{\overline{\Delta}_u(k;m) } 
    {\overline{\Delta}_u(k;m) } \right)} =\overline{\Delta}_u(k)\Bigg |_{\eqref{eq:equiDelt},\eqref{eq:equiLamb}}. \nonumber 
    \end{align}
    }
    Using a similar technique, it can be shown that $D_j(k) = \widehat{D}_j(k)$, upon convergence, and thus (i) holds. The proof for (ii) is similar, which is omitted for brevity.
           \\
           
 \item \textit{The KKT conditions of $\bm{{\mathcal{P}}}$ should be satisfied after the series of approximations converges in problem~$\bm{\widehat{\mathcal{P}}}_m$:} 
 Since the constraints~\eqref{eq:f1_con1}-\eqref{eq:f1_con2} are common in problems $(\bm{\mathcal{P}})$ and $(\widetilde{\bm{\mathcal{P}}})$, for the approximated constraint we should have 
 $\triangledown (1-\theta) ((a) + \theta{(b)} ) = \triangledown \left( (1-\theta) (\widetilde{a})+\theta{(b)} \right)$, upon convergence ($\triangledown$ denotes the gradient sign). 
 Note that $(a)$ involves the product between two ratios of posynomials 
 (as can be seen from~\eqref{eq:Xi}, one is $\frac{\Delta_j(k)}{\overline{\Delta}_u(k)}$, where both the numerator and denominator are posynomial with respect to the optimization variables according to~\eqref{eq:f1_con14}, 
 and the other one is $\sigma^{\mathsf{F}}_j$, which can be written as ratio of two posynomials according to~\eqref{eq:lemma1-1}), where we approximate the denominator of each ratio via a monomial in $\widetilde{(a)}$. 
 For compactness, let us define $(a) \triangleq \frac{A_1(\bm{x})A_2(\bm{x})}{B_1(\bm{x})B_2(\bm{x})}$, where $A_1(\bm{x})$ and $B_1(\bm{x})$ are posynomials corresponding to the numerator and denominator of $\frac{\Delta_j(k)}{\Delta_u(k)}$ (encompassing all the coefficients) and $A_2(\bm{x})$ and $B_2(\bm{x})$ are posynomials corresponding to the numerator and denominator of $\sigma^{\mathsf{F}}_j$ ($\bm{x}$ denotes the set of optimization variables). We provide the proof for the general case.
 In general, the posynomials in the denominators can be described as $B_1(\bm{x})=\sum_{k=1}^{K_1} u^{(1)}_k(\bm{x})$ and $B_2(\bm{x})=\sum_{k=1}^{K_2} u^{(2)}_k(\bm{x})$, where $u^{(1)}_k$-s and $u^{(2)}_k$-s are monomial functions.
 Accordingly, we can write $\widetilde{(a)}$ as $\widetilde{(a)}=\frac{A_1(\bm{x})A_2(\bm{x})}{\widetilde{B}_1(\bm{x})\widetilde{B}_2(\bm{x})}$, where $\widetilde{B}_1(\bm{x})=\prod_{k=1}^{K_1}\left( \frac{u^{(1)}_k(\bm{x})}{a_k^{(1)}(\bm{y})}\right)^{a^{(1)}_k(\bm{y})}$ and $\widetilde{B}_2(\bm{x})=\prod_{k=1}^{K_2}\left( \frac{u^{(2)}_k(\bm{x})}{a^{(2)}_k(\bm{y})}\right)^{a_k^{(2)}(\bm{y})}$ are the monomial approximation of ${B}_1(\bm{x})$ and ${B}_2(\bm{x})$ obtained according to~\eqref{eq:approxPosMon} (equivalent to the condensations carried out in~\eqref{eq:delta_u_approx} and~\eqref{eq:D_j_approx}).  In the following, we show that the desired result holds for partial derivative with respect to an arbitrary element $x_i$ considering $\bm{x}=[x_1,\cdots,x_i,\cdots,x_n]$ (note that upon convergence $\bm{x}=\bm{y}$ in~\eqref{eq:approxPosMon}, which in our problem translates to the equalities in~\eqref{eq:equiDelt},\eqref{eq:equiLamb}):
 
 {\small
          \begin{align}
            &\frac{\partial \left(  \frac{A_1(\bm{x})A_2(\bm{x})}{\widetilde{B}_1(\bm{x})\widetilde{B}_2(\bm{x})} \right)}{\partial x_i} \Bigg|_{\bm{x}=\bm{y}}=\frac{\frac{\partial A_1(\bm{x})A_2(\bm{x})}{\partial x_i}\widetilde{B}_1(\bm{y})\widetilde{B}_2(\bm{y})  -   \frac{\partial \widetilde{B}_1(\bm{x})\widetilde{B}_2(\bm{x})}{\partial x_i} A_1(\bm{y})A_2(\bm{y})}{\left(\widetilde{B}_1(\bm{y})\widetilde{B}_2(\bm{y})\right)^2}\Bigg|_{\bm{x}=\bm{y}}
           \nonumber \\&
           \overset{(a)}{=}\hspace{-0mm}\frac{\frac{\partial A_1(\bm{x})A_2(\bm{x})}{\partial x_i}\Bigg|_{\bm{x}=\bm{y}}       {B}_1(\bm{y}){B}_2(\bm{y})  -  \frac{\hspace{-1.8mm}\partial   \displaystyle\prod_{k=1}^{K_1}\left( \frac{u^{(1)}_k(\bm{x})}{a^{(1)}_k(\bm{y})}\right)^{\alpha^{(1)}_k(\bm{y})}}{\partial x_i}\Bigg|_{\bm{x}=\bm{y}}    {B}_2(\bm{y}) A_1(\bm{y})A_2(\bm{y})}{\left(   {B}_1(\bm{y}){B}_2(\bm{y})\right)^2}
            \nonumber \\&
           \hspace{45mm}+\frac{-\frac{\hspace{-1.8mm}\partial  \displaystyle\prod_{k=1}^{K_2}\left( \frac{u^{(2)}_k(\bm{x})}{a^{(2)}_k(\bm{y})}\right)^{a^{(2)}_k(\bm{y})}}{\partial x_i}\Bigg|_{\bm{x}=\bm{y}}    {B}_1(\bm{y}) A_1(\bm{y})A_2(\bm{y})}{\left(   {B}_1(\bm{y}){B}_2(\bm{y})\right)^2}
            \nonumber \\&
            =\frac{\frac{\partial A_1(\bm{x})A_2(\bm{x})}{\partial x_i}\bigg|_{\bm{x}=\bm{y}}      {B}_1(\bm{y}){B}_2(\bm{y})}{\left(   {B}_1(\bm{y}){B}_2(\bm{y})\right)^2}
            \nonumber \\&
            ~~~+\frac{ - {\displaystyle\sum_{n=1}^{K_1} \frac{\partial u^{(1)}_n(\bm{x})}{\partial x_i}\bigg|_{\bm{x}=\bm{y}} \left( \frac{u^{(1)}_n(\bm{y})}{\alpha^{(1)}_n(\bm{y})}\right)^{\alpha^{(1)}_n(\bm{y})-1} \left(\displaystyle\prod_{k=1, k\neq n}^{K_1}\left( \frac{u^{(1)}_k(\bm{y})}{\alpha^{(1)}_k(\bm{y})}\right)^{\alpha^{(1)}_k(\bm{y})}\right)}{B}_2(\bm{y}) A_1(\bm{y})A_2(\bm{y})}{\left(   {B}_1(\bm{y}){B}_2(\bm{y})\right)^2}
            \nonumber \\&
            ~~~+\frac{ - {\displaystyle\sum_{n=1}^{K_2} \frac{\partial u^{(2)}_n(\bm{x})}{\partial x_i}\Bigg|_{\bm{x}=\bm{y}} \left( \frac{u^{(2)}_n(\bm{y})}{a^{(2)}_n(\bm{y})}\right)^{a^{(2)}_n(\bm{y})-1} \left(\displaystyle\prod_{k=1, k\neq n}^{K_2}\left( \frac{u^{(2)}_k(\bm{y})}{a^{(2)}_k(\bm{y})}\right)^{a^{(2)}_k(\bm{y})}\right)}{B}_1(\bm{y}) A_1(\bm{y})A_2(\bm{y})}{\left(   {B}_1(\bm{y}){B}_2(\bm{y})\right)^2}
            \nonumber \\   &\overset{(b)}{=} \frac{\frac{\partial A_1(\bm{y})A_2(\bm{y})}{\partial x_i}\Bigg|_{\bm{x}=\bm{y}}    {B}_1(\bm{y}){B}_2(\bm{y}) }{\left(   {B}_1(\bm{y}){B}_2(\bm{y})\right)^2}\hspace{-8mm}
             \nonumber\\   &~~~+  \frac{ - {\displaystyle\sum_{n=1}^{K_1}\frac{\partial u^{(1)}_n(\bm{x})}{\partial x_i}\Bigg|_{\bm{x}=\bm{y}} {B}_1(\bm{y})^{\alpha^{(1)}_n(\bm{y})-1} \left({B}_1(\bm{y})^{\sum_{k=1, k\neq n}^{K_1}\alpha^{(1)}_k(\bm{y})}\right)} B_2(\bm{y})A_1(\bm{y})A_2(\bm{y})}{\left(   {B}_1(\bm{y}){B}_2(\bm{y})\right)^2}
            \nonumber \\   &~~~+  \frac{ - {\displaystyle\sum_{n=1}^{K_2}\frac{\partial u^{(2)}_n(\bm{x})}{\partial x_i}\Bigg|_{\bm{x}=\bm{y}} {B}_2(\bm{y})^{a^{(2)}_n(\bm{y})-1} \left({B}_2(\bm{y})^{\sum_{k=1, k\neq n}^{K_2}a^{(2)}_k(\bm{y})}\right)} B_1(\bm{y})A_1(\bm{y})A_2(\bm{y})}{\left(   {B}_1(\bm{y}){B}_2(\bm{y})\right)^2}
            \nonumber \\&
            \overset{(c)}{=} \frac{\frac{\partial A_1(\bm{y})A_2(\bm{y})}{\partial x_i}\bigg|_{\bm{x}=\bm{y}}   {B}_1(\bm{y}){B}_2(\bm{y}) }{\left(   {B}_1(\bm{y}){B}_2(\bm{y})\right)^2}\hspace{-8mm}
            \nonumber  \\&
             ~~~+\frac{ - \displaystyle\sum_{n=1}^{K}\frac{\partial u^{(1)}_n(\bm{x})}{\partial x_i}\bigg|_{\bm{x}=\bm{y}} B_2(\bm{y})A_1(\bm{y})A_2(\bm{y})-
             \displaystyle\sum_{n=1}^{K}\frac{\partial u^{(2)}_n(\bm{x})}{\partial x_i}\bigg|_{\bm{x}=\bm{y}} B_1(\bm{y})A_1(\bm{y})A_2(\bm{y})}{\left(   {B}_1(\bm{y}){B}_2(\bm{y})\right)^2}\hspace{-8mm}
            \nonumber  \\&
               = \frac{\frac{\partial A_1(\bm{x})A_2(\bm{x})}{\partial x_i}\bigg|_{\bm{x}=\bm{y}}   {B}_1(\bm{y}){B}_2(\bm{y})-\frac{\partial B_1(\bm{x})}{\partial x_i}\bigg|_{\bm{x}=\bm{y}} {A}_1(\bm{y}){A}_2(\bm{y}){B}_2(\bm{y})-\frac{\partial B_2(\bm{x})}{\partial x_i}\bigg|_{\bm{x}=\bm{y}}{A}_1(\bm{y}){A}_2(\bm{y}){B}_1(\bm{y}) }{\left(   {B}_1(\bm{y}){B}_2(\bm{y})\right)^2}\hspace{-8mm}
            \nonumber \\&=\frac{\partial \left(  \frac{A_1(\bm{x})A_2(\bm{x})}{   {B}_1(\bm{x}){B}_2(\bm{x})} \right)}{\partial x_i}\Bigg|_{\bm{x}=\bm{y}}.
                    \end{align}
                    }
          
         In $(a)$, we used the fact that $B_1(\bm{y})= \widetilde{B}_1(\bm{y})$ and $B_2(\bm{y})= \widetilde{B}_2(\bm{y})$ (this is the equality of monomial approximation with the original posynomial upon convergence that we showed in bullet point 2 above). In $(b)$, we used the fact that (see~\eqref{eq:approxPosMon}) $a^{(1)}_k(\bm{y})=u^{(1)}_k(\bm{y})/B_1(\bm{y})$ and $a^{(2)}_k(\bm{y})=u^{(2)}_k(\bm{y})/B_2(\bm{y})$, $\forall k$. Also,
         in $(c)$ we use the fact that $\sum_{k=1}^{K_{1}}\alpha^{(1)}_k(\bm{y})=1$ and $\sum_{k=1}^{K_{2}}a^{(2)}_k(\bm{y})=1$.  The proof for the rest of partial derivatives, and thus the gradient, is similar.
\end{enumerate}
Verification of the three aforementioned bullet points results in the conclusion of the proof.

\newpage
\section{Proof of Lemma~\ref{lemma:conceptDrift}} \label{app:lemConc}
Given the previous computed gradient for the local model at the device cluster, i.e.,  $\nabla F_c (\mathbf w_c(t^{\mathsf{V}}_c) | \widetilde{\mathcal{D}}_c(t^{\mathsf{V}}_c))$, the value of the local gradient given the outdated model $\mathbf w_c(t^{\mathsf{V}}_c)$ for the recent data distribution $\widetilde{\mathcal{D}}_c(t)$ at time $t$ can be expressed as follows:
\begin{align}
    \left \Vert \nabla F_c \left(\mathbf w_c(t^{\mathsf{V}}_c) \big| \widetilde{\mathcal{D}}_c(t) \right)\right\Vert^2 =&\Bigg \Vert \nabla F_c \left(\mathbf w_c(t^{\mathsf{V}}_c) \big| \widetilde{\mathcal{D}}_c(t) \right) 
    -\nabla F_c \left(\mathbf w_c(t^{\mathsf{V}}_c) \big| \widetilde{\mathcal{D}}_c(t-1) \right) 
    +\nabla F_c \left(\mathbf w_c(t^{\mathsf{V}}_c) \big| \widetilde{\mathcal{D}}_c(t-1) \right)
    \nonumber \\&
    -\nabla F_c \left(\mathbf w_c(t^{\mathsf{V}}_c) \big| \widetilde{\mathcal{D}}_c(t-2) \right)
    +\nabla F_c \left(\mathbf w_c(t^{\mathsf{V}}_c) \big| \widetilde{\mathcal{D}}_c(t-2) \right)-\cdots 
    \nonumber \\&
   -\nabla F_c (\mathbf w_c(t^{\mathsf{V}}_c) \big| \widetilde{\mathcal{D}}_c(t^{\mathsf{V}}_c)) +\nabla F_c (\mathbf w_c(t^{\mathsf{V}}_c) \big| \widetilde{\mathcal{D}}_c(t^{\mathsf{V}}_c))\Bigg\Vert^2 
    \nonumber \\& 
    \overset{(a)}{\leq} \Bigg(\Big \Vert \nabla F_c \left(\mathbf w(t^{\mathsf{V}}_c) \big| \widetilde{\mathcal{D}}_c(t) \right) 
    -\nabla F_c \left(\mathbf w_c(t^{\mathsf{V}}_c) \big| \widetilde{\mathcal{D}}_c(t-1) \right) \Big\Vert
    +\Big\Vert \nabla F_c \left(\mathbf w_c(t^{\mathsf{V}}_c) \big| \widetilde{\mathcal{D}}_c(t-1) \right)
   \nonumber  \\&
    -\nabla F_c \left(\mathbf w_c(t^{\mathsf{V}}_c) \big| \widetilde{\mathcal{D}}_c(t-2) \right)\Big\Vert
    +\Big\Vert\nabla F_c \left(\mathbf w_c(t^{\mathsf{V}}_c) \big| \widetilde{\mathcal{D}}_c(t-2) \right)-\nabla F_c \left(\mathbf w_c(t^{\mathsf{V}}_c) \big| \widetilde{\mathcal{D}}_c(t-3) \right)\Big\Vert +\cdots +
   \nonumber  \\&
  \Big\Vert\nabla F_c (\mathbf w_c(t^{\mathsf{V}}_c) \big| \widetilde{\mathcal{D}}_c(t^{\mathsf{V}}_c+1)) -\nabla F_c (\mathbf w_c(t^{\mathsf{V}}_c) \big| \widetilde{\mathcal{D}}_c(t^{\mathsf{V}}_c))\Big\Vert +\Big\Vert\nabla F_c (\mathbf w_c(t^{\mathsf{V}}_c) \big| \widetilde{\mathcal{D}}_c(t^{\mathsf{V}}_c))\Big\Vert \Bigg)^2 \nonumber \\
   &\overset{(b)}{\leq}
    (t-t^{\mathsf{V}}_c+1) \left\Vert\nabla F_c (\mathbf w_c(t^{\mathsf{V}}_c) \big| \widetilde{\mathcal{D}}_c(t^{\mathsf{V}}_c))\right\Vert^2+(t-t^{\mathsf{V}}_c+1)\sum_{t'=t^{\mathsf{V}}_c+1}^{t} \Lambda_c(t'),
\end{align}
where in $(a)$ we used triangle inequality, and (b) is the result of Cauchy–Schwarz inequality (i.e, for any $N$ numbers $ A_1,\cdots, A_N$, we have $\left(\sum_{i=1}^{N}  A_i\right)^2 \leq N \sum_{i=1}^{N} ( A_i)^2$).

\end{changemargin}
\newpage

\balance

\end{document}